\documentclass[10pt,twocolumn,letterpaper]{article}

\usepackage[pagenumbers]{cvpr} 

\usepackage{booktabs}  
\usepackage{graphicx}  
\usepackage{flafter} 
\usepackage{multirow}  
\usepackage{stfloats}  
\usepackage[most]{tcolorbox}
\definecolor{cvprblue}{rgb}{0.21,0.49,0.74}
\usepackage[pagebackref,breaklinks,colorlinks,allcolors=cvprblue]{hyperref}

\usepackage{xcolor}
\definecolor{vinaycolor}{RGB}{180,0,120}

\newif\ifshownotes
\shownotestrue

\newcommand{\pname}{EnLA}

\def\confName{CVPR}
\def\confYear{2027}

\title{EnLA: Enriching Linear Attention for Diffusion Transformers}

\author{Soumya Banerjee\textsuperscript{1}, Vinay Kumar Verma, Xiang Cheng\textsuperscript{2},\\
Ricardo Henao\textsuperscript{2} and Lawrence Carin\textsuperscript{1}\\[4pt]
\textsuperscript{1}BII, A*STAR\\
\textsuperscript{2}Department of Electrical and Computer Engineering, Duke University\\[4pt]
{\tt\small bsoumya@a-star.edu.sg, vinayugc@gmail.com, xiang.cheng@duke.edu}\\
{\tt\small ricardo.henao@duke.edu, Lawrence\_Carin@a-star.edu.sg}
}

\begin{document}
\maketitle

\begin{abstract}
Diffusion transformers generate high-quality images, but the quadratic cost of softmax attention makes long token sequences expensive.
Linear attention removes this cost by summing the key--value outer products of all tokens into one shared state per head, which every query then reads; this uniform aggregation, however, degrades generation quality.
We propose \pname{}, which enriches the keys and values \emph{before} they are aggregated.
A lightweight spatial mixer network (SMN), built from pointwise and depthwise convolutions, mixes features across attention heads and across neighboring spatial positions, and outputs signed residual corrections to the kernel-transformed keys and to the values.
Queries and the key-based normalization are left unchanged, so the operator remains linear in the number of tokens.
On ImageNet-1K class-conditional generation, \pname{} improves FID over vanilla linear attention, LiT, and softmax attention at every DiT scale from S/2 to XL/2 (\eg, $18.40$ \vs $22.03$ and $19.68$ for linear and softmax attention on DiT-XL/2 without guidance).
\pname{} also improves FID when finetuning a pretrained SANA text-to-image model on three datasets, and improves ImageNet-1K classification accuracy on DeiT backbones.
An analysis of the trained model shows that the key corrections account for most of the change in the attention readout.
\end{abstract}

\begin{figure}[!t]
    \centering
    \captionsetup{skip=2pt}
    \includegraphics[width=\linewidth]{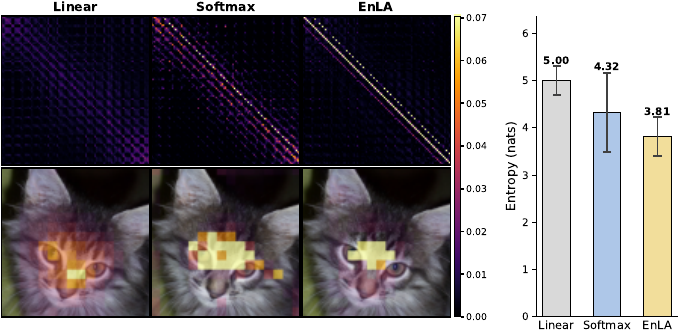}
    \caption{Effective token-mixing operators on DiT-XL/2~\cite{Peebles_2023_ICCV} at $256\times256$ ($N{=}256$), block 15, $t{=}500$. \textbf{Left top:} row-normalized absolute magnitudes, averaged over the 16 heads, shared color scale. \textbf{Left bottom:} the row for query token 151 over the input image. \textbf{Right:} entropy of each query's mixing weights, averaged over queries, heads and blocks; lower is more concentrated.}
    \vspace{-7pt}
    
    \label{fig:attention_magnitude}
\end{figure}

\begin{figure*}[t]
    \centering
    \captionsetup{skip=2pt}
    \scalebox{1}[1.04]{\includegraphics[width=\textwidth]{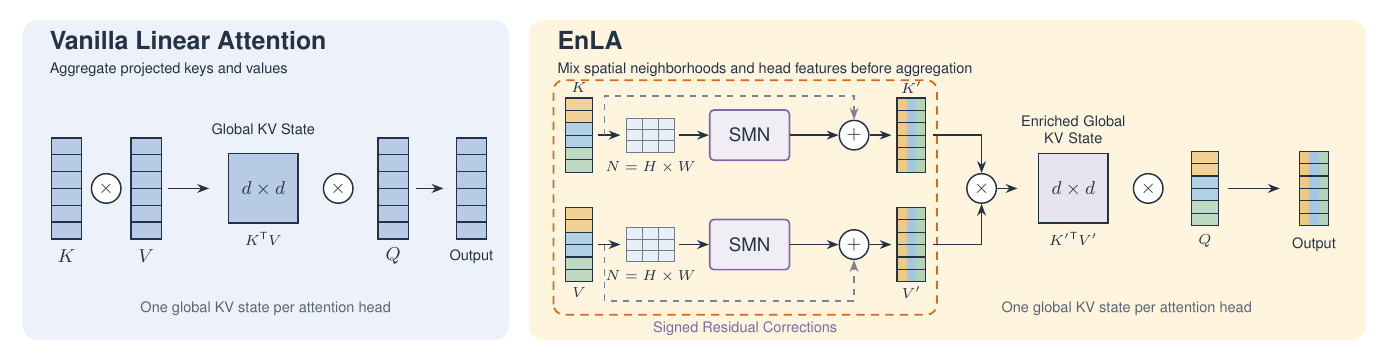}}
    \caption{Vanilla linear attention (left) and \pname{} (right).
    In \pname{}, a spatial mixer network (SMN) mixes spatial neighborhoods and features across attention heads to produce signed residual updates to the keys and values before the per-head KV state is formed.}
    \vspace{-7pt}
    
    \label{fig:linear_attention_comparison}
\end{figure*}

\vspace{-5mm}
\section{Introduction}
\label{sec:introduction}

Diffusion transformers (DiTs)~\cite{Peebles_2023_ICCV} have emerged as a simple and scalable architecture for image generation, adopting vision transformer backbones~\cite{dosovitskiy2021vit} in place of the U-Net architectures commonly used in diffusion models~\cite{NEURIPS2020_4c5bcfec,Rombach_2022_CVPR}.
By processing images as sequences of latent patches, they scale through model depth, width, and token count.
This scalability is constrained by the quadratic cost of softmax attention~\cite{NIPS2017_3f5ee243}: higher resolutions produce longer token sequences, sharply increasing attention computation and memory, and the cost is paid again at every denoising step.

Linear attention~\cite{pmlr-v119-katharopoulos20a} avoids the $N\times N$ query--key similarity matrix of softmax attention.
Replacing the softmax similarity with a decomposable kernel lets the computation be reordered from $(QK^\top)V$ to $Q(K^\top V)$, where $Q$, $K$, and $V$ denote queries, keys, and values, respectively: keys and values are first aggregated into a single $d\times d$ state, which each query then reads.
For $N$ tokens of dimension $d$, this lowers the cost from $\mathcal{O}(N^2d)$ to $\mathcal{O}(Nd^2)$ while retaining a global receptive field, which has motivated its use in diffusion models such as SANA~\cite{ICLR2025_34345e24} and LiT~\cite{Wang_2025_ICCV}.

However, reordering the attention computation has a price.
The state is a sum of token-wise key--value outer products with equal coefficients~\cite{cao2025saga}, and although retrieval remains query-dependent, every query reads the same precomputed state~\cite{zhang2026mhla}.
The associations of different tokens are superimposed and can interfere~\cite{schlag2021linear}, and a single global state keeps no separate summary of individual spatial neighborhoods~\cite{Wang_2025_ICCV}, so global access does not by itself give effective local modeling.
Softmax attention, in contrast, determines query-dependent weights before aggregating values.
The effect on generation quality is substantial: on ImageNet-1K~\cite{deng2009imagenet,russakovsky2015imagenet}, replacing softmax attention with vanilla linear attention in DiT-S/2 increases FID from $68.55$ to $84.65$.
This raises our central question:

\vspace{-1mm}
\begin{tcolorbox}[
    colback=yellow!10,
    colframe=yellow!40!black,
    boxrule=0.5pt,
    arc=2pt,
    left=6pt,
    right=6pt,
    top=3pt,
    bottom=3pt,
    width=\columnwidth
]
\centering
\textbf{\emph{How can linear attention retain its
computational scaling while improving the quality
of image generation in diffusion transformers?}}
\end{tcolorbox}

We approach this question by changing what enters the shared state, not how it is read.
We propose \pname{}, in which a \emph{spatial mixer network} (SMN) enriches the keys and values before aggregation (Figure~\ref{fig:linear_attention_comparison}).
SMN restores the two-dimensional token grid and applies a grouped pointwise convolution, a depthwise convolution, and a second grouped pointwise convolution, with nonlinearities in between.
The pointwise layers mix information across heads through complementary feature groupings; the depthwise layer gathers each token's spatial neighborhood.
The output is added as a residual correction to the kernel-transformed keys and to the values.
The corrections are signed, so the enriched keys are not restricted to non-negative coordinates, while the queries and the normalization by the original keys are left untouched.
\pname{} still forms one state per head, but each token's contribution to that state now depends on its spatial context and on the other heads, and the cost remains linear in $N$.
The effect is visible in a trained DiT-XL/2. Each token's output is a weighted combination of all $N$ tokens, and under vanilla linear attention those weights are spread almost uniformly, so every token contributes to every other. With \pname{}, the weights are less uniform and concentrate on fewer tokens, resembling softmax attention (Figure~\ref{fig:attention_magnitude}).

Convolutions have been combined with attention before: CvT~\cite{wu2021cvt} uses convolutional query/key/value projections in softmax attention for recognition, and EfficientViT~\cite{Cai_2023_ICCV}, focused linear attention~\cite{Han_2023_ICCV}, and LiT~\cite{Wang_2025_ICCV} add depthwise convolutions to linear attention.
\pname{} differs in applying a nonlinear, cross-head mixer to both the post-kernel keys and the values, as signed residual corrections, under the unmodified key normalization. The proposed correction can be integrated into existing linear-attention models while retaining their underlying attention formulation.

We evaluate \pname{} on ImageNet-1K class-conditional generation across four DiT scales, on text-to-image finetuning of a pretrained SANA model, and on ImageNet-1K classification.
Our contributions are:
\begin{itemize}
    \item We introduce \pname{}, which enriches linear attention with a spatial mixer network that injects spatial-neighborhood and cross-head information into the keys and values before they are aggregated into the shared state.
    \item We formulate the enrichment as signed residual corrections to the kernel-transformed keys and to the values, retaining the original queries and key normalization, and we analyze how the key and value corrections change the attention readout in a trained DiT-XL/2.
    \item \pname{} improves FID over vanilla linear attention and softmax attention at every DiT scale from S/2 to XL/2, with and without classifier-free guidance, and over SANA and MHLA when finetuning a pretrained SANA model. It also improves ImageNet-1K classification accuracy on DeiT-Tiny and DeiT-Small.
\end{itemize}

\section{Related Work}
\label{sec:related_work}

\textbf{Diffusion transformers.} Diffusion models learn to reverse a gradual noising process~\cite{pmlr-v37-sohl-dickstein15,NEURIPS2020_4c5bcfec,song2021scoresde}, with later work reducing the number of sampling steps~\cite{song2021ddim,pmlr-v139-nichol21a,NEURIPS2022_260a14ac} and improving conditioning~\cite{NEURIPS2021_49ad23d1,ho2022classifierfreediffusionguidance}.
Latent diffusion lowers the cost by denoising in the latent space of a pretrained autoencoder~\cite{Rombach_2022_CVPR}.
DiT~\cite{Peebles_2023_ICCV} replaces the convolutional U-Net~\cite{10.1007/978-3-319-24574-4_28} denoiser with a vision transformer over latent patches and improves with depth, width, and token count; transformer backbones are now standard in image generation~\cite{Bao_2023_CVPR,ma2024sit,ICLR2024_fe989bb0,10.1007/978-3-031-73411-3_5,pmlr-v235-esser24a,li2024hunyuanditpowerfulmultiresolutiondiffusion,ICLR2025_5c46ae13}.
Because the cost of softmax attention grows quadratically with the token count, high-resolution generation motivates more efficient attention.

\textbf{Linear attention.} Linear attention replaces the softmax similarity with a kernel that factorizes over queries and keys, so that keys and values can be aggregated once and read by all queries in time linear in the number of tokens~\cite{pmlr-v119-katharopoulos20a}; Performers obtain such a factorization from positive random features that approximate the softmax kernel~\cite{choromanski2022rethinkingattentionperformers}.
The resulting gap to softmax attention has been attributed to several causes, each with a corresponding remedy: focused linear attention~\cite{Han_2023_ICCV} sharpens the kernel and restores feature diversity; InLine attention~\cite{han2024inline} addresses non-injectivity and weak local modeling; MLLA~\cite{han2024mlla} incorporates Mamba~\cite{gu2023mamba} inspired design choices; RALA~\cite{fan2025rala} raises the rank of the aggregated state and of the output features; MALA~\cite{fan2025mala} accounts for query magnitude; and SAGA~\cite{cao2025saga} gates the token contributions that enter the state.
Closest to our setting, MHLA~\cite{zhang2026mhla} observes that all queries read one global summary and restores query-conditioned diversity by partitioning the tokens into groups, forming a local key--value summary per group, and letting each query group read a learned mixture of these summaries.
MHLA therefore changes \emph{how many} states are formed and how they are read.
\pname{} is complementary: it keeps a single state per attention head and changes \emph{what} each token contributes to it.
We compare with MHLA on text-to-image finetuning and on classification (Sections~\ref{sec:t2i} and~\ref{sec:classification}).

\textbf{Convolution inside attention.} Several works introduce convolutional locality into the attention operator itself.
CvT~\cite{wu2021cvt} replaces the linear query, key, and value projections of softmax attention with depthwise-separable convolutions; it targets image recognition and, to our knowledge, convolutional projections of this kind have not been studied in diffusion transformers.
For linear attention, EfficientViT~\cite{Cai_2023_ICCV} aggregates the query, key, and value tokens with small-kernel depthwise convolutions to obtain multi-scale tokens before ReLU linear attention, and focused linear attention~\cite{Han_2023_ICCV} adds a depthwise convolution of the values to the attention output.
\pname{} differs from these designs in four respects: the mixer acts on the \emph{kernel-transformed} keys and on the values, so that both matching and content are enriched; it is nonlinear and couples attention heads through two grouped pointwise layers with complementary feature groupings; its output is a \emph{signed} residual that is not passed through the kernel map again; and the normalization is still computed from the original keys.

\textbf{Linear attention for diffusion.} SANA~\cite{ICLR2025_34345e24,pmlr-v267-xie25b,Chen_2025_ICCV} combines ReLU linear attention with a deep-compression autoencoder and feed-forward blocks that contain a depthwise convolution, enabling efficient high-resolution text-to-image generation.
LiT~\cite{Wang_2025_ICCV} studies how to convert a DiT to linear attention and recommends a simplified linear attention with a depthwise convolution, few heads, weight inheritance from a pretrained DiT~\cite{Peebles_2023_ICCV}, and distillation~\cite{hinton2015distilling}.
DiG~\cite{zhu2024dig} builds diffusion backbones on gated linear attention, while EDiT~\cite{Becker_2025_ICCV} combines linear attention with KV compression and uses distillation~\cite{hinton2015distilling} from a pretrained PixArt-$\Sigma$ model~\cite{10.1007/978-3-031-73411-3_5}.
In these models locality enters through the feed-forward block or through a convolution applied around the attention operator.
\pname{} instead places a nonlinear spatial and cross-head mixer on the keys and values \emph{before} the shared state is formed, and can be added to an existing linear-attention model such as SANA or DiG.

\section{Proposed Approach}
\label{sec:proposed_approach}

\begin{figure*}[t]
    \centering
    \captionsetup{skip=2pt}
    \makebox[\textwidth][c]{\includegraphics[width=1.03\textwidth]{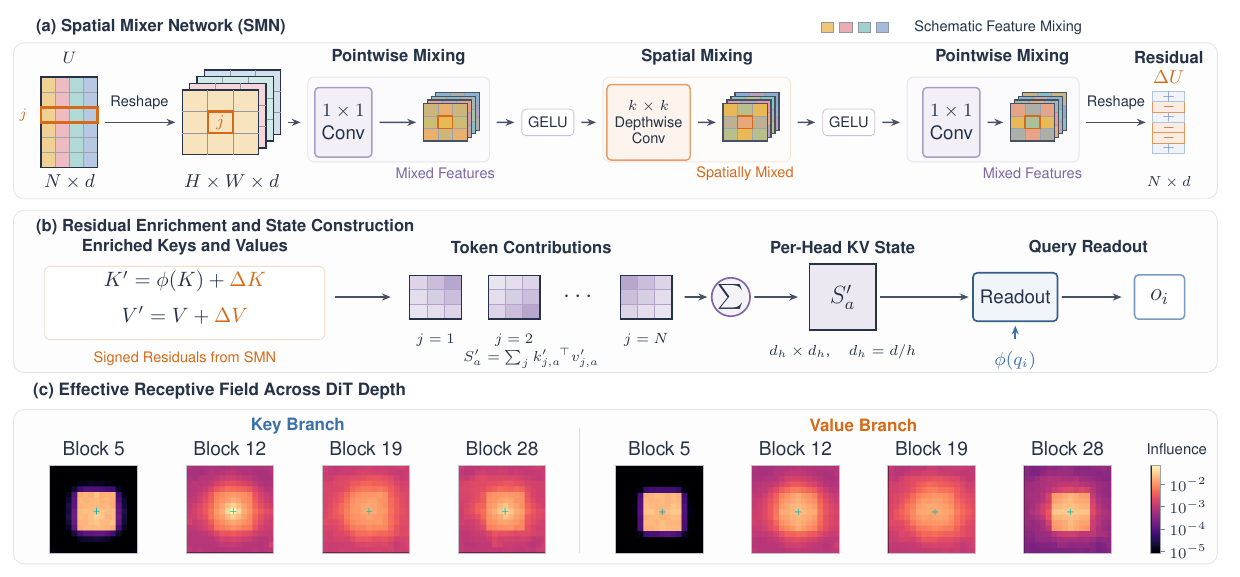}}
    \caption{Overview of \pname{}.
    (a) The spatial mixer network (SMN), which generates signed residual corrections.
    (b) Residual enrichment of keys and values before per-head KV aggregation and query-dependent retrieval.
    (c) Effective receptive fields of SMN's depthwise outputs across DiT-XL/2 blocks: key (left) and value (right), averaged over 32 images at $t=500$.
    Maps show normalized squared-gradient influence on the noisy latent input on a shared logarithmic scale; cyan crosses mark the query position.}
    \vspace{-5pt}
    
    \label{fig:proposed_method_overview}
\end{figure*}

\subsection{Preliminaries}
\label{sec:preliminaries}

Given a noisy latent $z_t$ at diffusion timestep $t$, a patch embedding layer produces an $H\times W$ grid of embeddings, which is flattened into a token sequence $X\in\mathbb{R}^{N\times d}$, where $N=HW$ and $d$ is the embedding dimension.
Queries, keys, and values are obtained through learned projections:
\begin{equation}
    Q=XW_Q,\qquad K=XW_K,\qquad V=XW_V,
    \label{eq:qkv_projections}
\end{equation}
where $W_Q,W_K,W_V\in\mathbb{R}^{d\times d}$, and $q_i$, $k_i$, and $v_i$ denote the $i$-th rows of $Q$, $K$, and $V$.
With $h$ attention heads, the $d$ channels are split into $h$ groups of $d_h=d/h$ channels and attention is applied to each group independently.
For clarity we write the attention equations for a single head and omit the head index and the batch dimension.

\textbf{Softmax Attention.} Softmax attention computes query-dependent similarities between all token pairs and uses their row-normalized weights to aggregate values, $O=\operatorname{softmax}\!\left(QK^\top/\sqrt{d_h}\right)V$~\cite{NIPS2017_3f5ee243}.
Forming the $N\times N$ similarities and aggregating the values costs $\mathcal{O}(N^2d)$ over all heads, which becomes expensive as image resolution grows.

\textbf{Linear Attention.} Linear attention replaces the exponential similarity with a factorized kernel $\phi(q_i)\phi(k_j)^\top$, where $\phi$ is a nonnegative feature map applied to each row~\cite{pmlr-v119-katharopoulos20a}.
Associativity of matrix multiplication then allows keys and values to be aggregated before the queries are applied:
\begin{equation}
\small
    o_i
    =\frac{\sum_{j=1}^{N}\phi(q_i)\phi(k_j)^\top v_j}
          {\sum_{m=1}^{N}\phi(q_i)\phi(k_m)^\top}
    =\frac{\phi(q_i)\left(\sum_{j=1}^{N}\phi(k_j)^\top v_j\right)}
          {\phi(q_i)\left(\sum_{m=1}^{N}\phi(k_m)^\top\right)}.
    \label{eq:linear_attention}
\end{equation}
The keys and values form a single state $S=\phi(K)^\top V\in\mathbb{R}^{d_h\times d_h}$, and $s=\sum_{m=1}^{N}\phi(k_m)^\top\in\mathbb{R}^{d_h\times1}$ supplies the normalization.
Both are computed once and shared by all queries, giving $o_i=\phi(q_i)S/(\phi(q_i)s)$.
No $N\times N$ matrix is formed, and the cost is $\mathcal{O}(Nd_h^2)$ per head, \ie, $\mathcal{O}(Nd^2/h)$ in total.
Thus, linear attention replaces an $N\times N$ similarity matrix with a $d_h\times d_h$ state per head, reducing attention computation when $N>d_h$.
This efficiency comes with an expressiveness trade-off: softmax attention can form sharp, query-specific distributions that emphasize some tokens and suppress others, whereas linear attention constrains every query to read one state built by summing token-wise key--value outer products with equal coefficients~\cite{cao2025saga,zhang2026mhla}.
\pname{} keeps the shared-state computation and instead enriches the keys and values that form these outer products.

\subsection{Spatial Mixer Network}
\label{sec:spatial_mixer_network}

Two kinds of information are not explicitly incorporated when a token's key and value enter the state of vanilla multi-head linear attention~\cite{pmlr-v119-katharopoulos20a}.
First, each head builds its state from its own channel group; heads are combined only after retrieval, so they cannot coordinate what they write.
Second, a token's contribution depends on that token alone, so local image structure, which helps separate signal from noise, is not represented in the state.
The spatial mixer network (SMN) supplies both before aggregation.

Let $\bar K=\phi(K)$ denote the kernel-transformed keys, where $\phi(x)=\operatorname{ELU}(x)+1$ is applied element-wise.
SMN takes $U\in\mathbb{R}^{N\times d}$, which is either $\bar K$ or $V$ at full feature width (all heads), restores its spatial arrangement $\mathcal{R}(U)\in\mathbb{R}^{H\times W\times d}$, and computes (Figure~\ref{fig:proposed_method_overview}(a))
\begin{equation}
    \begin{aligned}
        F_1&=\sigma\!\left(\operatorname{PW}_1(\mathcal{R}(U))\right),\\
        F_2&=\sigma\!\left(\operatorname{DW}_{k\times k}(F_1)\right),\\
        f_\theta(U)&=\mathcal{R}^{-1}\!\left(\operatorname{PW}_2(F_2)\right),
    \end{aligned}
    \label{eq:spatial_mixer}
\end{equation}
where $\sigma$ is GELU, $\operatorname{PW}_1$ and $\operatorname{PW}_2$ are $1\times1$ grouped pointwise convolutions, $\operatorname{DW}_{k\times k}$ is a depthwise convolution with padding that preserves the spatial size, $F_1,F_2\in\mathbb{R}^{H\times W\times d}$ with hidden width $d$, and $\theta$ collects the convolution parameters.
$\operatorname{PW}_1$ mixes features within channel groups at each position; the depthwise convolution then gathers the spatial neighborhood of these mixed features, and $\operatorname{PW}_2$ combines the result into the correction.
The two pointwise layers use complementary feature groupings to exchange information across groups and heads.
The key and value projections are already full-width linear maps, so what SMN adds is essentially \emph{nonlinear} cross-head interactions after the kernel map for keys and directly on projected values.
$\operatorname{PW}_2$ has no output activation, so the correction can be positive or negative.
Although each SMN is local, its input at block $\ell$ is produced by the preceding blocks, so the region of the input that influences a correction can widen through intermediate layers (Figure~\ref{fig:proposed_method_overview}(c)).

\begin{table*}[!t]
\centering
\caption{ImageNet-1K class-conditional generation \emph{without classifier-free guidance}. Best result per block in bold. $^\dagger$As reported in~\cite{Wang_2025_ICCV}, does not report sFID.}
\vspace{-5pt}
\label{tab:c2i-nocfg}
\footnotesize
\setlength{\tabcolsep}{5pt}
\renewcommand{\arraystretch}{1.05}
\begin{tabular*}{\textwidth}{@{\extracolsep{\fill}}lllccccc@{}}
\toprule
Scale & Resolution & Method & IS $\uparrow$ & FID $\downarrow$ & sFID $\downarrow$ & Prec. $\uparrow$ & Rec. $\uparrow$ \\
\midrule
\multirow{7}{*}{DiT-S/2} & \multirow{4}{*}{$256{\times}256$} & Softmax & 19.93 & 68.55 & 12.22 & 0.372 & 0.560 \\
 && Linear  & 16.22 & 84.65 & 19.11 & 0.301 & 0.454 \\
 && LiT$^\dagger$~\cite{Wang_2025_ICCV} & 22.08 & 63.21 & -- & 0.386 & \textbf{0.583} \\
 && \pname{} & \textbf{22.93} & \textbf{61.15} & \textbf{11.43} & \textbf{0.396} & 0.582 \\
\cmidrule{2-8}
 & \multirow{3}{*}{$512{\times}512$} & Softmax & 17.31 & 83.49 & \textbf{14.31} & 0.376 & \textbf{0.527} \\
 && Linear  & 11.25 & 126.41 & 31.19 & 0.222 & 0.288 \\
 && \pname{} & \textbf{19.57} & \textbf{75.90} & 14.40 & \textbf{0.406} & 0.513 \\
\midrule
\multirow{2}{*}{DiG-S/2} & \multirow{2}{*}{$256{\times}256$} & DiG~\cite{zhu2024dig} & 22.81 & 62.06 & 11.77 & 0.390 & 0.560 \\
&& \pname{} & \textbf{23.96} & \textbf{57.82} & \textbf{11.20} & \textbf{0.423} & \textbf{0.586} \\
\midrule
\multirow{4}{*}{DiT-B/2} & \multirow{4}{*}{$256{\times}256$} & Softmax & 34.18 & 42.37 & 8.08 & 0.496 & \textbf{0.632} \\
 && Linear  & 28.86 & 50.75 & 10.69 & 0.447 & 0.597 \\
 && LiT$^\dagger$~\cite{Wang_2025_ICCV} & 35.79 & 40.86 & -- & 0.497 & 0.629 \\
 && \pname{} & \textbf{38.38} & \textbf{38.36} & \textbf{8.06} & \textbf{0.506} & 0.630 \\
\midrule
\multirow{4}{*}{DiT-L/2} & \multirow{4}{*}{$256{\times}256$} & Softmax & 60.22 & 23.10 & 6.36 & \textbf{0.611} & 0.634 \\
 && Linear  & 53.93 & 27.18 & 7.49 & 0.577 & 0.635 \\
 && LiT$^\dagger$~\cite{Wang_2025_ICCV} & 59.02 & 24.04 & -- & 0.597 & \textbf{0.636} \\
 && \pname{} & \textbf{63.04} & \textbf{21.86} & \textbf{6.09} & 0.608 & 0.633 \\
\midrule
\multirow{4}{*}{DiT-XL/2} & \multirow{4}{*}{$256{\times}256$} & Softmax & 67.76 & 19.68 & 6.21 & 0.635 & 0.622 \\
 && Linear  & 63.66 & 22.03 & 6.86 & 0.609 & 0.635 \\
 && LiT$^\dagger$~\cite{Wang_2025_ICCV} & 65.39 & 20.66 & -- & 0.624 & \textbf{0.636} \\
 && \pname{} & \textbf{71.47} & \textbf{18.40} & \textbf{5.85} & \textbf{0.637} & 0.634 \\
\bottomrule
\end{tabular*}
\vspace{-5pt}
\end{table*}

\subsection{Enriched Linear Attention}
\label{sec:enriched_linear_attention}

\pname{} applies two independently parameterized mixers to the kernel-transformed keys and to the values, and adds their outputs as residuals (Figure~\ref{fig:proposed_method_overview}(b)):
\begin{equation}
    \Delta K=f_{\theta_K}(\bar K),\qquad
    \Delta V=f_{\theta_V}(V),
    \label{eq:spam_corrections}
\end{equation}
\begin{equation}
    K'=\bar K+\Delta K,\qquad V'=V+\Delta V.
    \label{eq:enriched_kv}
\end{equation}
The key correction changes how a token matches queries; the value correction changes the content the token contributes.
No positivity map is applied after the addition, so $K'$ can have negative coordinates.

\textbf{State Construction and Readout.} After enrichment, $K'$ and $V'$ are split into heads.
Each head aggregates its enriched keys and values into one state, which the queries read under the normalization of the \emph{original} keys:
\begin{equation}
    S'=K'^\top V',\qquad
    o_i=\frac{\phi(q_i)S'}
        {\phi(q_i)\left(\sum_{m=1}^{N}\phi(k_m)^\top\right)}.
    \label{eq:enriched_attention}
\end{equation}
The denominator is that of vanilla linear attention and remains positive, whereas a normalization computed from the signed $K'$ could vanish or change sign.
With $S=\bar K^\top V$ the vanilla state, expanding $S'$ gives
\begin{equation}
    \begin{aligned}
        S'&=S+\Delta S,\\
        \Delta S&=\bar K^\top\Delta V
        +\Delta K^\top V+\Delta K^\top\Delta V,
    \end{aligned}
    \label{eq:state_residual}
\end{equation}
where the three terms are the value correction, the key correction, and their interaction.
With $s=\sum_{m=1}^{N}\phi(k_m)^\top$, the output decomposes as
\begin{equation}
    o_i=\underbrace{\frac{\phi(q_i)S}{\phi(q_i)s}}_{\text{vanilla linear attention}}
    +\underbrace{\frac{\phi(q_i)\Delta S}{\phi(q_i)s}}_{\text{learned correction}}.
    \label{eq:output_residual}
\end{equation}

\textbf{Connection to Linear Attention.} The first term has the form of vanilla linear attention and the second is a learned contribution under the same denominator; when both corrections vanish, \pname{} reduces to vanilla linear attention exactly.
The effective weight of token $j$ for query $i$, $\phi(q_i)k_j'^{\top}/(\phi(q_i)s)$, can be negative and the weights need not sum to one, so a token can be subtracted from as well as added to the readout.
Equation~\eqref{eq:output_residual} is an algebraic identity and does not imply that the second term is small: in a trained model the two terms are of comparable magnitude (Section~\ref{sec:ablations}).

\textbf{Complexity Analysis.} Aggregation and readout in Equation~\eqref{eq:enriched_attention} cost $\mathcal{O}(Nd^2/h)$, as in vanilla linear attention.
Each SMN adds $\mathcal{O}(Nd^2/G)$ for the two grouped pointwise convolutions and $\mathcal{O}(Nk^2d)$ for the depthwise convolution, where $G$ is the number of channel groups.
As resolution increases, $N$ grows while $d$, $h$, and $d_h=d/h$ remain fixed.
For fixed widths, head count, group count, and kernel size, \pname{} is therefore linear in $N$ and avoids the $\mathcal{O}(N^2d)$ pairwise computation of softmax attention; measured costs are given in Section~\ref{sec:ablations}.

\section{Results and Discussion}
\label{sec:results_and_discussions}

\subsection{Class-Conditional Image Generation}
\label{sec:c2i}

\begin{figure*}[!t]
    \centering
    \captionsetup{skip=2pt}
    \includegraphics[width=\textwidth,height=.95\textheight,keepaspectratio]{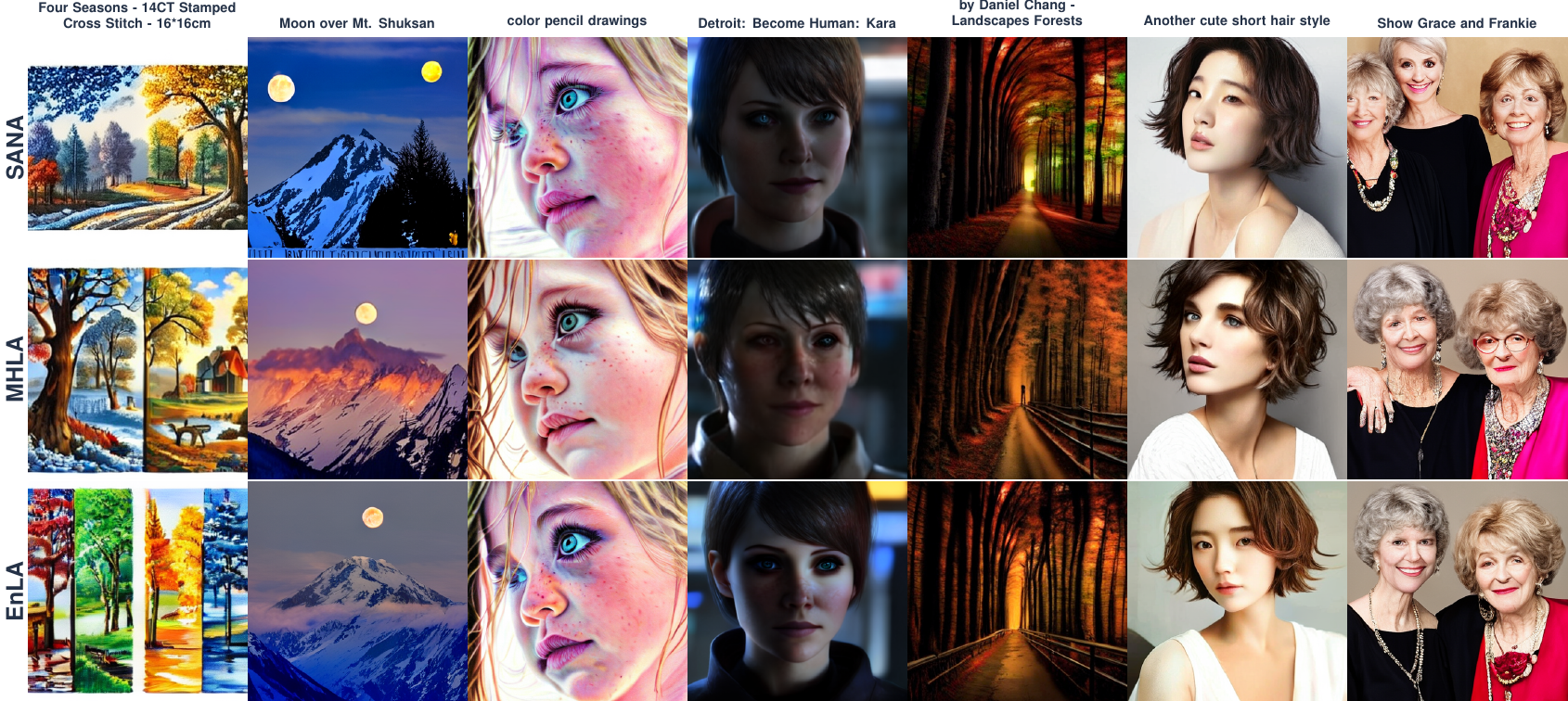}
    \caption{Text-to-image samples from SANA, MHLA, and \pname{} (top to bottom) after finetuning. Each column corresponds to the prompt shown above it; all three models use the same seed, classifier-free guidance scale, and number of sampling steps for each prompt.}
    \vspace{-5pt}
    \label{fig:t2i_qualitative}
\end{figure*}

We evaluate class-conditional generation on ImageNet-1K~\cite{deng2009imagenet,russakovsky2015imagenet} with DiT-S/2, B/2, L/2, and XL/2 at $256\times256$, and with DiT-S/2 at $512\times512$.
Softmax attention, vanilla linear attention, and \pname{} use the same DiT backbone~\cite{Peebles_2023_ICCV}, including depth, width, head count, and conditioning; only the attention module differs.
Following the training and sampling recipe of DiT~\cite{Peebles_2023_ICCV}, all models are trained from scratch for 400K steps with a global batch size of 256 on four NVIDIA H200 GPUs, using AdamW~\cite{loshchilov2017decoupled} with a constant learning rate of $10^{-4}$, no weight decay, and no warmup.
We generate 50,000 samples with 250 sampling steps and report FID~\cite{heusel2017fid}, Inception Score (IS)~\cite{salimans2016is}, sFID~\cite{nash2021sfid}, and precision and recall~\cite{kynkaanniemi2019precisionrecall}.
Table~\ref{tab:c2i-nocfg} reports results without classifier-free guidance.

\textbf{Comparison with vanilla linear and softmax attention.} \pname{} obtains the best FID and IS at all four DiT scales.
On DiT-S/2, FID decreases from $84.65$ (vanilla linear) and $68.55$ (softmax) to $\mathbf{61.15}$; on DiT-XL/2, from $22.03$ and $19.68$ to $\mathbf{18.40}$.
Precision and recall are within $0.003$ of the best method at every scale.
\pname{} thus closes the gap between linear and softmax attention under identical training and obtains a lower FID than the softmax DiT.
The same ordering holds with classifier-free guidance (scale $1.5$; Table~\ref{tab:c2i-cfg}), where \pname{} reaches FID $\mathbf{5.41}$ on DiT-XL/2 against $6.00$ for softmax and $6.75$ for vanilla linear attention.

\textbf{Comparison with linear-attention diffusion models.} Compared with the results reported for LiT~\cite{Wang_2025_ICCV}, \pname{} has lower FID and higher IS and precision at every scale, with recall within $0.003$.
On DiG-S/2~\cite{zhu2024dig}, adding SMN lowers FID from $62.06$ to $\mathbf{57.82}$ and improves all other metrics.

\textbf{Higher resolution.} At $512\times512$ the token count is four times larger.
\pname{} with DiT-S/2 achieves FID $\mathbf{75.90}$, compared with $126.41$ for vanilla linear attention and $83.49$ for softmax attention; sFID and recall are slightly behind softmax ($14.40$ \vs $14.31$ and $0.513$ \vs $0.527$).

\textbf{Attention patterns.} Figure~\ref{fig:attention_magnitude} (left) compares the effective token-mixing operators of the three attention types in one block (block 15) of DiT-XL/2.
Vanilla linear attention spreads weight diffusely over all tokens, whereas \pname{}, like softmax attention, concentrates weight on spatially neighboring tokens, which appear as bands around the diagonal. We provide attention map overlays for all blocks in Appendix~\ref{sec:supp_attention_patterns}.

\begin{table}[t]
\centering
\caption{ImageNet-1K class-conditional generation \emph{with classifier-free guidance}
(scale $1.5$), for the models of Table~\ref{tab:c2i-nocfg}.
}
\vspace{-5pt}
\label{tab:c2i-cfg}
\footnotesize
\setlength{\tabcolsep}{4.3pt}
\renewcommand{\arraystretch}{1.05}
\begin{tabular}{@{}lllcccc@{}}
\toprule
Scale & Resolution & Method & IS $\uparrow$ & FID $\downarrow$ & Prec. $\uparrow$ & Rec. $\uparrow$ \\
\midrule
\multirow{6}{*}{DiT-S/2} & \multirow{3}{*}{$256{\times}256$} & Softmax & 32.98 & 45.57 & 0.472 & 0.543 \\
 && Linear  & 25.33 & 62.36 & 0.381 & 0.470 \\
 && \pname{} & \textbf{41.14} & \textbf{37.71} & \textbf{0.520} & \textbf{0.547} \\
\cmidrule{2-7}
 & \multirow{3}{*}{$512{\times}512$} & Softmax & 26.58 & 59.67 & 0.489 & 0.521 \\
 && Linear  & 15.05 & 105.51 & 0.277 & 0.323 \\
 && \pname{} & \textbf{30.43} & \textbf{52.17} & \textbf{0.544} & \textbf{0.537} \\
\midrule
\multirow{3}{*}{DiT-B/2} & \multirow{3}{*}{$256{\times}256$} & Softmax & 70.38 & 20.52 & 0.652 & \textbf{0.555} \\
 && Linear  & 60.40 & 26.73 & 0.588 & 0.553 \\
 && \pname{} & \textbf{84.63} & \textbf{16.71} & \textbf{0.677} & 0.541 \\
\midrule
\multirow{3}{*}{DiT-L/2} & \multirow{3}{*}{$256{\times}256$} & Softmax & 134.47 & 7.64 & 0.779 & \textbf{0.528} \\
 && Linear  & 125.88 & 9.23 & 0.749 & 0.527 \\
 && \pname{} & \textbf{141.75} & \textbf{6.98} & \textbf{0.781} & 0.524 \\
\midrule
\multirow{3}{*}{DiT-XL/2} & \multirow{3}{*}{$256{\times}256$} & Softmax & 154.76 & 6.00 & \textbf{0.805} & 0.525 \\
 && Linear  & 149.19 & 6.75 & 0.783 & \textbf{0.534} \\
 && \pname{} & \textbf{163.56} & \textbf{5.41} & 0.801 & 0.524 \\
\bottomrule
\end{tabular}
\vspace{-5pt}
\end{table}

\subsection{Text-to-Image Generation}
\label{sec:t2i}

We next test whether the enrichment transfers to a pretrained text-to-image model that already uses linear attention.
We add SMN to the self-attention layers of Sana-600M~\cite{ICLR2025_34345e24}, leaving cross-attention unchanged, and compare with the original SANA architecture and with MHLA~\cite{zhang2026mhla}, all starting from the same public $512\times512$ checkpoint.

All three variants are finetuned for 40K steps with a batch size of 256 on four NVIDIA H200 GPUs, using AdamW~\cite{loshchilov2017decoupled} with no weight decay and a learning rate warmed up over 1K steps and then held at $10^{-4}$.
The SMN branches are randomly initialized and trained jointly with the inherited weights; both baselines are finetuned under the same protocol using their official implementations.
For each of LAION-Aes6.5~\cite{schuhmann2022laionb}, MS-COCO~\cite{lin2014microsoft}, and MultiGen-20M~\cite{qin2023unicontrol}, we use an $80/20$ train/test split and evaluate on the held-out $20\%$, whose images serve as the FID and LPIPS reference.
Samples are generated at $512\times512$ with a guidance scale of $4.5$ and evaluated with FID, CLIP score~\cite{radford2021clip}, and LPIPS~\cite{zhang2018lpips}.

Table~\ref{tab:t2i-sana} shows that \pname{} has the lowest FID on LAION-Aes6.5 and MultiGen-20M and matches SANA on COCO-2017, with equal LPIPS at the reported precision.
Relative to SANA, FID decreases from $8.02$ to $\mathbf{7.84}$ on LAION-Aes6.5 and from $8.76$ to $\mathbf{8.67}$ on MultiGen-20M; on COCO-2017 the difference ($19.58$ \vs $19.51$) is within the sampling spread.
MHLA improves FID over SANA on LAION-Aes6.5 but not on the other two datasets.
CLIP scores match or remain close to SANA's and exceed MHLA's on all three datasets.
The FID gains are modest, and indicate that enrichment can also benefit a pretrained linear-attention generator.

Figure~\ref{fig:t2i_qualitative} provides a qualitative comparison under matched sampling settings.
\pname{} depicts four distinct seasonal panels in the first column and avoids the duplicated moon and text-like artifacts produced by SANA in the second.
It also produces clearer facial detail in the Kara example and, in the final column, avoids both MHLA's malformed hand and the additional person generated by SANA.

\begin{table}[t]
\footnotesize
\centering
\setlength{\tabcolsep}{4pt}
\caption{Text-to-image finetuning of Sana-600M at $512{\times}512$ with guidance scale $4.5$. Results are averaged over two samplings. Best in bold, second best in blue.}
\vspace{-5pt}
\label{tab:t2i-sana}
\resizebox{\linewidth}{!}{%
\begin{tabular}{l|lccc}
\toprule
Dataset & Method & FID $\downarrow$ & CLIP $\uparrow$ & LPIPS $\downarrow$ \\
\midrule
\multirow{3}{*}{LAION-Aes6.5} & SANA & $8.02\,\pm\,0.03$ & $\mathbf{26.15\,\pm\,0.00}$ & $0.68\,\pm\,0.00$ \\
 & MHLA & \textcolor{blue}{$7.94\,\pm\,0.01$} & $25.93\,\pm\,0.02$ & $0.68\,\pm\,0.00$ \\
 & \pname{} & $\mathbf{7.84\,\pm\,0.03}$ & \textcolor{blue}{$26.05\,\pm\,0.01$} & $0.68\,\pm\,0.00$ \\
\midrule
\multirow{3}{*}{COCO-2017} & SANA & \textcolor{blue}{$19.58\,\pm\,0.13$} & \textcolor{blue}{$26.51\,\pm\,0.04$} & $0.74\,\pm\,0.00$ \\
 & MHLA & $20.57\,\pm\,0.12$ & $26.36\,\pm\,0.01$ & $0.74\,\pm\,0.00$ \\
 & \pname{} & $\mathbf{19.51\,\pm\,0.19}$ & $\mathbf{26.51\,\pm\,0.01}$ & $0.74\,\pm\,0.00$ \\
\midrule
\multirow{3}{*}{MultiGen-20M} & SANA & \textcolor{blue}{$8.76\,\pm\,0.07$} & $\mathbf{25.47\,\pm\,0.03}$ & $0.72\,\pm\,0.00$ \\
 & MHLA & $9.24\,\pm\,0.08$ & $25.36\,\pm\,0.01$ & $0.72\,\pm\,0.00$ \\
 & \pname{} & $\mathbf{8.67\,\pm\,0.06}$ & \textcolor{blue}{$25.45\,\pm\,0.01$} & $0.72\,\pm\,0.00$ \\
\bottomrule
\end{tabular}%
}
\vspace{-5pt}
\end{table}

\subsection{Image Classification}
\label{sec:classification}

SMN does not depend on the denoising objective, so we also evaluate \pname{} for recognition on ImageNet-1K.
We replace the attention in DeiT-Tiny and DeiT-Small~\cite{touvron2021deit} with \pname{}, keeping everything else fixed.
Following the training recipe used in prior work~\cite{touvron2021deit,fan2025mala,zhang2026mhla,Han_2023_ICCV,han2024inline}, we train from scratch at $224\times224$ for 300 epochs with a global batch size of 1024 and a peak learning rate of $10^{-3}$.

Table~\ref{tab:classification} shows that vanilla linear attention loses $2.4$ and $2.2$ points of top-1 accuracy relative to softmax attention on DeiT-Tiny and DeiT-Small, mirroring the gap observed for generation.
\pname{} closes this gap and exceeds softmax attention, reaching $\mathbf{76.4\%}$ and $\mathbf{81.0\%}$ with fewer FLOPs than softmax attention.
It is $0.6$ points above the best prior linear attention on DeiT-Tiny and on par with it on DeiT-Small ($+0.1$).
Results for a hierarchical backbone are given in Appendix~\ref{sec:supp_classification}.

\textbf{Implementation Details.} SMN mixes features with a two-stage block-diagonal factorization of a full-width pointwise map: two $1\times1$ grouped pointwise transformations under complementary feature groupings, with an intervening depthwise convolution and GELU~\cite{hendrycks2016gaussian} activations between the stages. We use $G=16$ groups throughout, with a $5\times5$ depthwise kernel for classification and $7\times7$ for generation.

\begin{table}[t]
\centering
\caption{ImageNet-1K classification with different attention mechanisms on DeiT. Results marked with
$^\dagger$ are reproduced under the same training setup as {\pname}.}
\vspace{-5pt}

\label{tab:classification}
\label{tab:deit-tiny}
\label{tab:deit-small}
\footnotesize
\setlength{\tabcolsep}{3pt}
\renewcommand{\arraystretch}{1.05}
\begin{tabular*}{\linewidth}{@{\extracolsep{\fill}}clccc@{}}
\toprule
 & Attention type & Params & FLOPs & Top-1 Acc. $\uparrow$ \\
\midrule
\multirow{7}{*}{\rotatebox{90}{DeiT-Tiny}} & Softmax Attention~\cite{touvron2021deit} & 5.7M & 1.2G & 72.2 \\
 & Linear Attention & 5.7M & 1.1G & 69.8 \\
 & Focused LA~\cite{Han_2023_ICCV} & 6.1M & 1.1G & 74.1 \\
 & InLine~\cite{han2024inline} & 6.5M & 1.1G & 74.5 \\
 & MALA~\cite{fan2025mala} & 6.3M & 1.1G & 75.1 \\
 & MHLA~\cite{zhang2026mhla} & 5.7M & 1.1G & 75.8 \\
 & \textbf{\pname{}} & 5.9M & 1.1G & \textbf{76.4} \\
\midrule
\multirow{6}{*}{\rotatebox{90}{DeiT-Small}} & Softmax Attention~\cite{touvron2021deit} & 22M & 4.6G & 79.8 \\
 & Linear Attention & 22M & 4.2G & 77.6 \\
 & RALA~\cite{fan2025rala} & 24M & 4.6G & 80.4 \\
 & MALA~\cite{fan2025mala} & 24M & 4.6G & 80.3 \\
 & MHLA$^{\dagger}$~\cite{zhang2026mhla} & 22M & 4.2G & 80.9 \\
 & \textbf{\pname{}} & 22M & 4.4G & \textbf{81.0} \\
\bottomrule
\end{tabular*}
\vspace{-5pt}
\end{table}

\subsection{Analysis and Ablations}
\label{sec:ablations}
\label{sec:retrieval_analysis}

\begin{table}[t]
    \centering
    \caption{Component ablations on DiT-S/2 at $256\times256$.}
\vspace{-5pt}
    \label{tab:ablation_baselines}
    \footnotesize
    \setlength{\tabcolsep}{7pt}
    \renewcommand{\arraystretch}{0.95}
        \begin{tabular}{@{}lcccc@{}}
            \toprule
            Method & FID $\downarrow$ & sFID $\downarrow$ & Prec. $\uparrow$ & Rec. $\uparrow$ \\
            \midrule
            \pname{} - no value enrichment & 61.52 & 11.79 & 0.394 & 0.572 \\
            \pname{} - no key enrichment & 62.14 & 12.21 & 0.389 & 0.578 \\
            \pname{} - no spatial mixing & 61.32 & 11.83 & 0.397 & 0.576 \\
            \pname{} - no pointwise mixing & 61.42 & 11.86 & 0.396 & 0.575 \\
            Softmax + SMN & 65.12 & 13.37 & 0.337 & 0.505 \\
            \midrule
            \pname{} & 61.15 & 11.43 & 0.396 & 0.582 \\
            \bottomrule
        \end{tabular}
\vspace{-5pt}
\end{table}

\begin{figure*}[!t]
    \centering
    \captionsetup{skip=2pt}
    \begin{minipage}{\textwidth}
    \newlength{\ablationplotheight}
    \settoheight{\ablationplotheight}{\includegraphics[width=.37\textwidth,trim=0 3pt 0 0,clip]{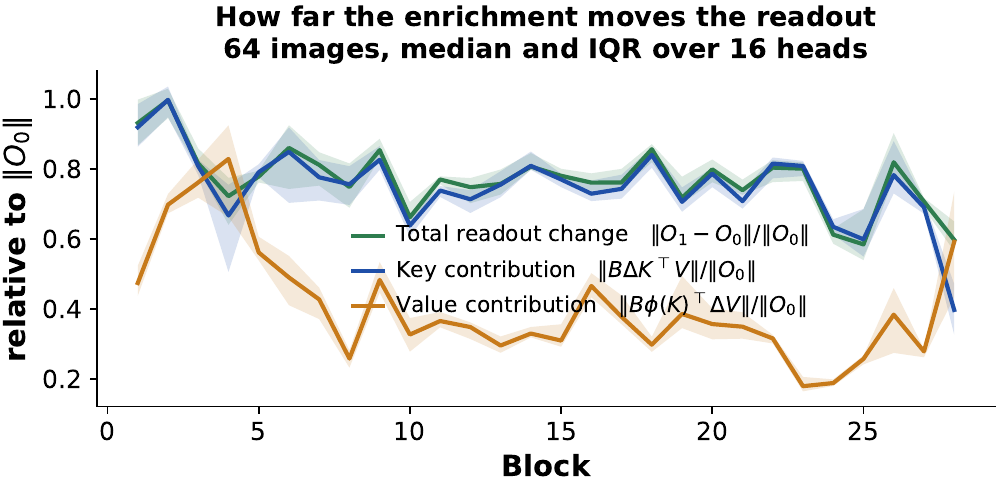}}
    \setlength{\ablationplotheight}{0.9595\ablationplotheight}
    \begin{minipage}[b]{0.329\textwidth}
        \centering\vspace{0pt}
        \includegraphics[width=\linewidth,height=\ablationplotheight,keepaspectratio,trim=0 3pt 0 0,clip]{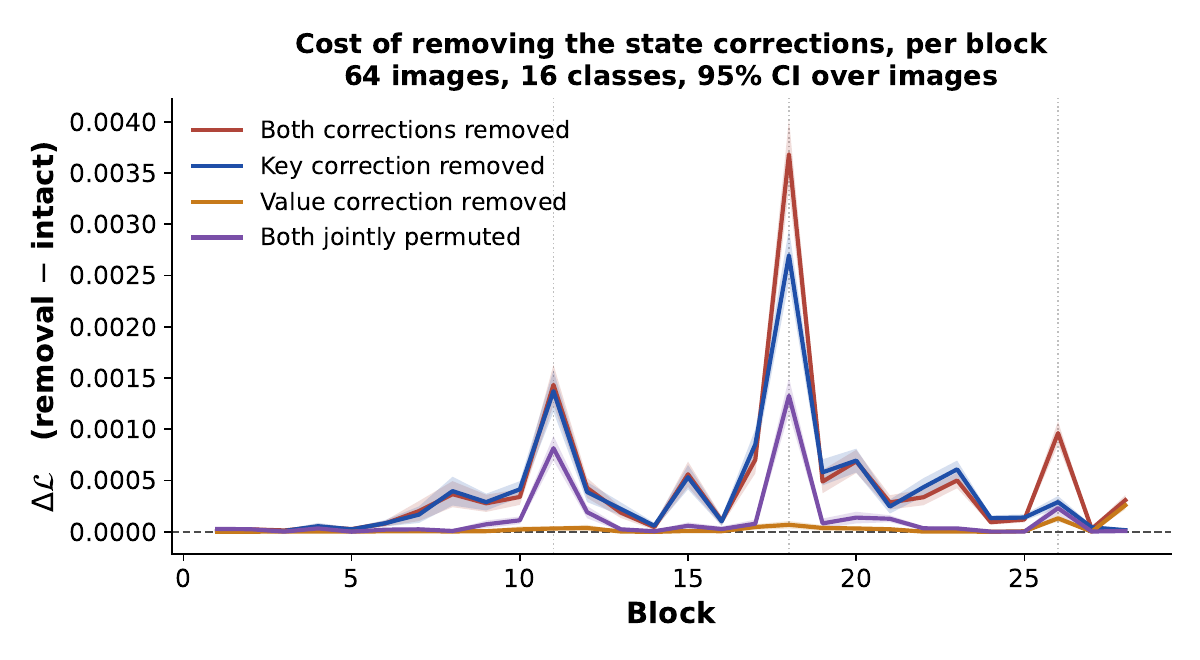}
        \par\vspace{2pt}{\footnotesize\textbf{(a)} Denoising interventions}
    \end{minipage}\hspace{2pt}%
    \begin{minipage}[b]{0.329\textwidth}
        \centering\vspace{0pt}
        \includegraphics[width=\linewidth,height=\ablationplotheight,keepaspectratio,trim=0 3pt 0 0,clip]{figures/readout_depth.pdf}
        \par\vspace{2pt}{\footnotesize\textbf{(b)} Readout contributions}
    \end{minipage}\hspace{2pt}%
    \begin{minipage}[b]{0.329\textwidth}
        \centering\vspace{0pt}
        \includegraphics[width=\linewidth,height=\ablationplotheight,keepaspectratio]{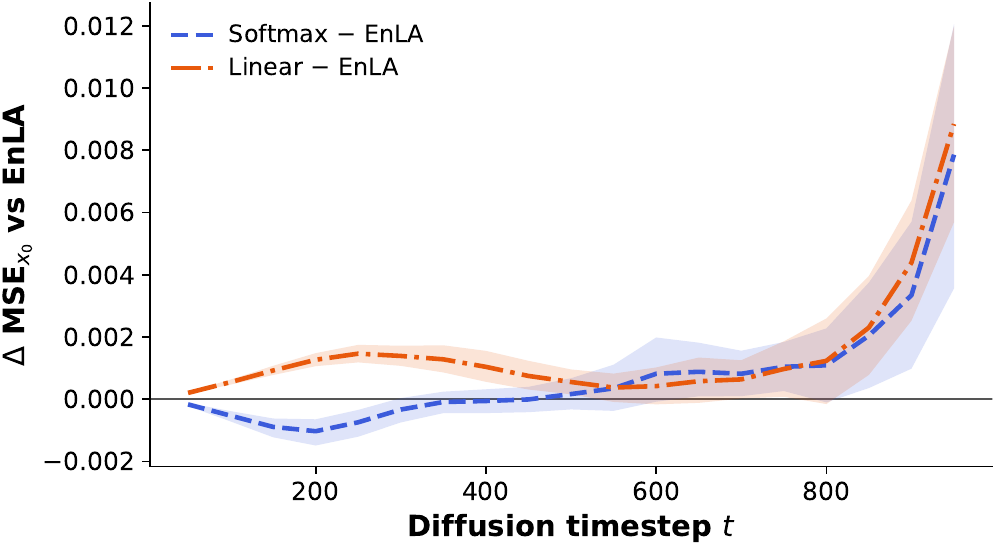}
        \par\vspace{2pt}{\footnotesize\textbf{(c)} Denoising Error Across Noise Levels}
    \end{minipage}
    \end{minipage}
    \caption{Analysis of \pname{} on DiT-XL/2.
    (a) Increase in denoising error when the key correction, the value correction, or both are removed at a single block, with 95\% confidence intervals.
    (b) Change in the attention readout caused by the corrections, relative to the vanilla readout $\|O_0\|$; median and interquartile range across heads.
    (c) Clean-latent MSE difference, $\Delta\mathrm{MSE}_{x_0}$ (baseline minus \pname{}); positive values favor \pname{}. Results average 100 shared noise draws per image across 64 images from 16 classes; bands show 95\% paired-bootstrap confidence intervals.}
    \vspace{-5pt}
    \label{fig:kv_ablation}
\end{figure*}

\textbf{How the Enrichment Affects Retrieval.} Let $B$ be the matrix of normalized queries with rows $B_{i,:}=\phi(q_i)/(\phi(q_i)s)$, where $s=\sum_{m=1}^{N}\phi(k_m)^\top$.
The vanilla and enriched readouts, $O_0=B\bar K^\top V$ and $O_1=BK'^\top V'$, differ by
\begin{equation}
    O_1-O_0=B\Delta K^\top V+BK'^\top\Delta V.
    \label{eq:readout_residual}
\end{equation}
We analyze the DiT-XL/2 model of Table~\ref{tab:c2i-nocfg} ($256\times256$) on 64 ImageNet images from 16 classes at $t\in\{900,700,500,300,100\}$.
Figure~\ref{fig:kv_ablation}(b) plots $\|O_1-O_0\|/\|O_0\|$ together with the key-only contribution $\|B\Delta K^\top V\|/\|O_0\|$ and the value-only contribution $\|B\bar K^\top\Delta V\|/\|O_0\|$; each uses the original counterpart, so the interaction term $B\Delta K^\top\Delta V$ belongs to neither.
Two observations follow.
First, the corrections are not a small perturbation: the median relative readout change across blocks is $0.77$, with per-block medians ranging from $0.58$ to $1.00$, so the learned term in Equation~\eqref{eq:output_residual} is comparable in magnitude to the vanilla term.
Second, the key-only contribution exceeds the value-only contribution in $26$ of the $28$ blocks, with medians across blocks of $0.76$ and $0.36$, respectively; the value-only contribution is larger only in blocks $3$ and $27$.

Figure~\ref{fig:kv_ablation}(a) measures the change in noise-prediction MSE, averaged over timesteps within each image, when the corrections are removed from the state construction of a single block while queries and normalization are left intact.
Removing the key correction increases the error at most blocks, most strongly at blocks 11 and 18, whereas removing the value correction alone has little effect.
The trained model therefore relies mainly on the key corrections.
These interventions are applied to a model trained with both corrections, so they measure reliance and not whether a model trained without a component would perform worse; that question is addressed by the component ablation below.

\begin{figure}[t]
    \centering
    \makebox[\linewidth][c]{\includegraphics[width=1.05\linewidth]{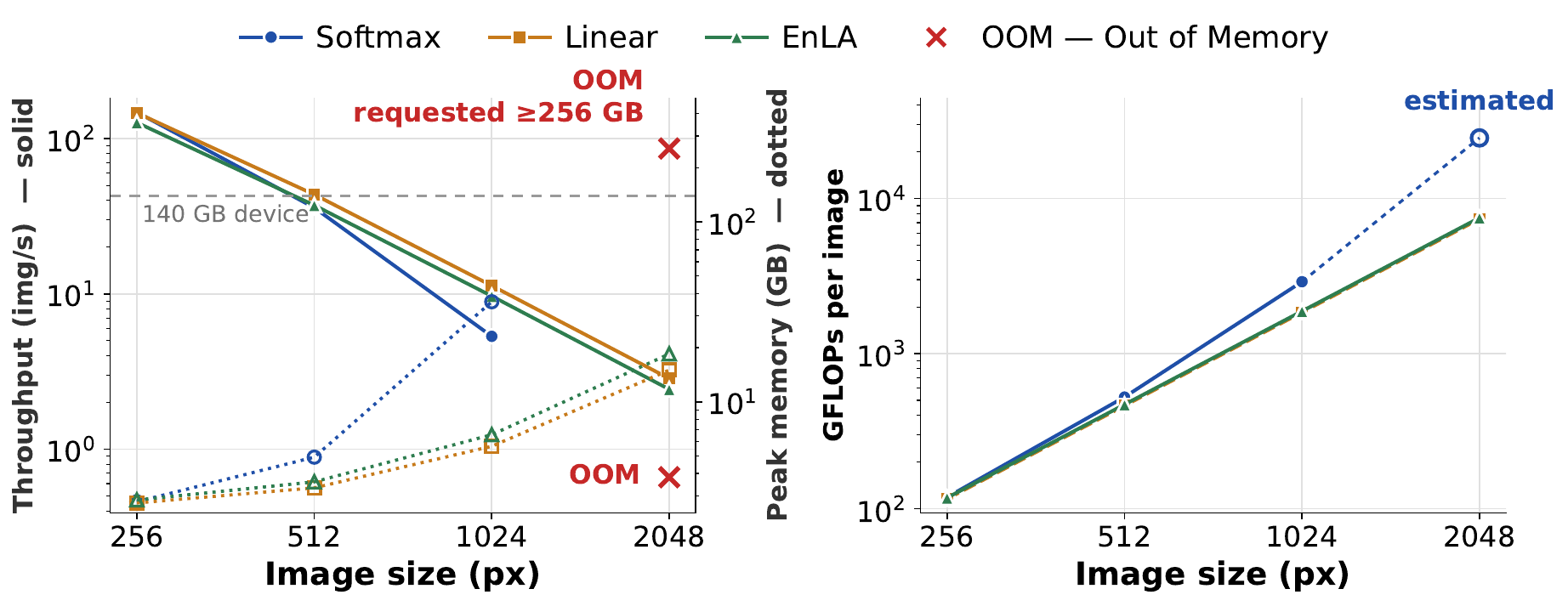}}
    \vspace{-15pt}
    \caption{Inference cost across image resolutions. \textbf{Left:} throughput (solid) and peak GPU memory (dotted); crosses indicate OOM. \textbf{Right:} GFLOPs per image; the hollow marker denotes the estimated softmax cost at $2048\times2048$.}
    \vspace{-9pt}
    \label{fig:computational_cost}
\end{figure}

\textbf{Denoising Across Noise Levels.} Figure~\ref{fig:kv_ablation}(c) compares clean-latent reconstruction errors under identical latents, labels, and noise using checkpoints trained for 400K steps.
We estimate the clean latent from each model's predicted noise and report MSE differences relative to \pname{}.
\pname{} achieves lower error than both softmax and vanilla linear attention at high noise levels, where little image structure remains visible.
This highlights its ability to recover the underlying clean latent from heavily corrupted inputs at noise levels corresponding to early reverse sampling.
At low noise levels \pname{} improves over vanilla linear attention but softmax attention has slightly lower error; differences narrow at intermediate timesteps.

\textbf{Computational Cost.} Figure~\ref{fig:computational_cost} reports DiT-XL/2 inference (batch 16, FP32, H200). At $256\times256$ and $512\times512$ softmax and vanilla linear attention have comparable throughput. At $1024\times1024$, \pname{} reaches 9.68 img/s at 6.56\,GB versus 5.33 img/s at 36.00\,GB for softmax ($1.82\times$ throughput, $81.8\%$ less peak memory); at $2048\times2048$ it runs at 2.43 img/s and 18.45\,GB while softmax runs out of memory.
Both use unfused implementations, so softmax stores the $N\times N$ attention matrix; fused kernels (\eg, FlashAttention~\cite{dao2022flashattention}) avoid this and would lower its memory and runtime, but not its quadratic computation.
See Appendix~\ref{sec:supp_computational_cost} for full results, including parameter counts.


\textbf{Component Analysis.} Table~\ref{tab:ablation_baselines} examines the contributions of key and value enrichment and the two mixing operations.
Removing key enrichment increases FID from $61.15$ to $62.14$, while removing value enrichment raises it to $61.52$, indicating a larger contribution from key enrichment.
Removing spatial or pointwise mixing also increases FID, to $61.32$ and $61.42$, respectively.
The full model achieves the lowest FID and sFID among these variants, although precision and recall do not improve uniformly.
Additionally, we examine whether key--value enrichment also benefits softmax attention. Applying SMN reduces FID from $68.55$ to $65.12$.

\textbf{Limitations.} \pname{} retains a single shared state per head and thus the representational constraints of shared-state aggregation.

\section{Conclusions}
\label{sec:conclusions}

We presented \pname{}, which enriches linear attention by modifying the keys and values before they are aggregated into the shared state.
A spatial mixer network adds signed residual corrections that carry spatial-neighborhood and cross-head information, while the queries, the key normalization, and the linear scaling in token count are unchanged.
On ImageNet-1K class-conditional generation, \pname{} improves FID over vanilla linear attention and softmax attention at every DiT scale from S/2 to XL/2. It also improves a pretrained SANA model when added to its linear attention, and improves classification accuracy on DeiT backbones. An analysis of the trained DiT-XL/2 shows that the corrections change the attention readout substantially and that the key corrections matter most. Component ablations further support the contributions of both enrichment branches and the spatial and pointwise mixing operations.

\section*{LLM Usage Statement}
\label{sec:llm_usage_statement}

We use large language models (LLMs) to refine our writing for clarity, readability, grammatical correctness, and logical flow, and to assist with code debugging, including Slurm-based HPC job launching. All research ideas, experimental designs, and experimental choices are the authors' own, and the authors take full responsibility for the contents of this paper.


{
    \small
    \let\publisherbackslash\backslash
    \def\backslash{\ensuremath{\publisherbackslash}}
    \bibliographystyle{ieeenat_fullname}
    \bibliography{main1}
}

\begingroup
\appendix
\setcounter{figure}{0}
\setcounter{table}{0}
\setcounter{equation}{0}
\renewcommand{\theHsection}{supp.\arabic{section}}
\renewcommand{\theHfigure}{supp.\arabic{figure}}
\renewcommand{\theHtable}{supp.\arabic{table}}
\renewcommand{\theHequation}{supp.\arabic{equation}}
\clearpage
\twocolumn[{\centering\Large\bfseries Appendix\par\vspace{12pt}}]

\begin{table}[!b]
\centering
\caption{Additional text-to-image generation results at 512px on MidJourney-v6 and 1024px on DALL-E 3 1M.}
\vspace{-5pt}
\label{tab:supp-t2i-additional}
\small
\setlength{\tabcolsep}{2pt}
\renewcommand{\arraystretch}{2.5}
\begin{tabular}{@{}c@{\hspace{14pt}}lccc@{}}
\toprule
 & Method & FID $\downarrow$ & CLIP $\uparrow$ & LPIPS $\downarrow$ \\
\midrule
\multirow{3}{*}{\rotatebox{90}{\shortstack{MidJourney-v6\\[4pt]512px}}} & \shortstack[l]{SANA\\\cite{ICLR2025_34345e24}} & $10.20\!\pm\!0.05$ & $30.39\!\pm\!0.00$ & $0.66\!\pm\!0.00$ \\
 & \shortstack[l]{MHLA\\\cite{zhang2026mhla}} & $10.24\!\pm\!0.04$ & $30.39\!\pm\!0.02$ & $0.66\!\pm\!0.00$ \\
 & EnLA & $\mathbf{9.80\!\pm\!0.02}$ & $\mathbf{30.41\!\pm\!0.01}$ & $0.66\!\pm\!0.00$ \\
\midrule
\noalign{\vskip 6pt}
\multirow{2}{*}{\rotatebox{90}{\shortstack{DALL-E3 1M\\[4pt]1024px}}} & \shortstack[l]{SANA\\\cite{ICLR2025_34345e24}} & $11.31\!\pm\!0.02$ & $32.85\!\pm\!0.01$ & $0.72\!\pm\!0.00$ \\
 & EnLA & $\mathbf{11.00\!\pm\!0.01}$ & $\mathbf{32.88\!\pm\!0.00}$ & $0.72\!\pm\!0.00$ \\[6pt]
\bottomrule
\end{tabular}
\vspace{-5pt}
\end{table}

\section{Attention Patterns Across Blocks}
\label{sec:supp_attention_patterns}
We extend the visualization on the left of Figure~1 in the main paper to all 28 blocks of DiT-XL/2~\cite{Peebles_2023_ICCV}. Figures~\ref{fig:supp_attention_blocks_1}--\ref{fig:supp_attention_blocks_5} show the attention-magnitude matrices and query-token overlays for linear attention, softmax attention, and EnLA.

\section{Additional Text-to-Image Samples}
\label{sec:supp_t2i_qualitative}
Figures~\ref{fig:supp_t2i_qualitative_1}--\ref{fig:supp_t2i_qualitative_6} present additional generation results from SANA~\cite{ICLR2025_34345e24}, MHLA~\cite{zhang2026mhla}, and EnLA, complementing the qualitative comparison in Figure~4 of the main paper.
Figures~\ref{fig:supp-dalle-grid}--\ref{fig:supp-dalle-remaining-2} and Figures~\ref{fig:supp-midjourney-1}--\ref{fig:supp-midjourney-3} provide further EnLA text-to-image samples at $1024\times1024$ and $512\times512$ resolutions, respectively, covering diverse subjects and visual styles.
Table~\ref{tab:supp-t2i-additional} reports additional quantitative results on MidJourney-v6~\cite{brivangl_midjourney_v6_llava_2024} at 512px and DALL-E3 1M~\cite{Egan_Dalle3_1_Million_2024} at 1024px, with all models finetuned for 60K steps. On MidJourney-v6, EnLA reduces FID from 10.20 for SANA and 10.24 for MHLA to $\mathbf{9.80}$, while improving the CLIP score to $\mathbf{30.41}$. On DALL-E 3 1M, EnLA improves FID from 11.31 to $\mathbf{11.00}$ and the CLIP score from 32.85 to $\mathbf{32.88}$ over SANA. LPIPS remains unchanged at the reported precision on both datasets.


\begin{table}[!b]
\centering
\caption{ImageNet-1K generation without classifier-free guidance.}
\vspace{-5pt}
\label{tab:supp-gating-fid}
\small
\setlength{\tabcolsep}{5pt}
\renewcommand{\arraystretch}{1.05}
\begin{tabular*}{\columnwidth}{@{\extracolsep{\fill}}lllc@{}}
\toprule
Scale & Resolution & Method & FID $\downarrow$ \\
\midrule
\multirow{4}{*}{DiT-S/2} & \multirow{2}{*}{$256{\times}256$} & MHLA~\cite{zhang2026mhla} & 59.80 \\
 & & EnLA & \textbf{58.43} \\
\cmidrule{2-4}
 & \multirow{2}{*}{$512{\times}512$} & MHLA~\cite{zhang2026mhla} & 78.63 \\
 & & EnLA & \textbf{73.26} \\
\midrule
\multirow{2}{*}{DiT-B/2} & \multirow{2}{*}{$256{\times}256$} & MHLA~\cite{zhang2026mhla} & 37.47 \\
 & & EnLA & \textbf{35.89} \\
\midrule
\multirow{2}{*}{DiT-L/2} & \multirow{2}{*}{$256{\times}256$} & MHLA~\cite{zhang2026mhla} & 21.37 \\
 & & EnLA & \textbf{20.92} \\
\midrule
\multirow{2}{*}{DiT-XL/2} & \multirow{2}{*}{$256{\times}256$} & MHLA~\cite{zhang2026mhla} & 19.17 \\
 & & EnLA & \textbf{17.77} \\
\bottomrule
\end{tabular*}
\vspace{-5pt}
\end{table}

\section{Additional Class-Conditional Generation Results}
\label{sec:supp-nocfg}
\label{sec:supp-cfg}
Table~\ref{tab:supp-gating-fid} compares EnLA with MHLA~\cite{zhang2026mhla} on ImageNet-1K~\cite{deng2009imagenet} generation, with EnLA achieving lower FID~\cite{heusel2017fid} across all reported scales and resolutions.
Tables~\ref{tab:supp-c2i-nocfg} and~\ref{tab:supp-c2i-cfg} provide the complete generation metrics without classifier-free guidance and with CFG $=1.5$, respectively, including results with query enrichment and output gating.
Figures~\ref{fig:supp-imagenet-1}--\ref{fig:supp-imagenet-4} show class-conditional generation samples at 256px using EnLA with the DiT-XL/2 backbone.

\section{Computational Cost}
\label{sec:supp_computational_cost}
Figure~\ref{fig:supp-flash-throughput} compares EnLA with FlashAttention~\cite{dao2022flashattention} using DiT-XL/2 at batch size 8 on a single NVIDIA H200. At $2048\times2048$, EnLA achieves 7.26 images/s against 4.89 images/s for FlashAttention; at $4096\times4096$, the corresponding throughputs are 1.78 and 0.39 images/s. These correspond to $\mathbf{1.49\times}$ and $\mathbf{4.56\times}$ speedups, computed from median latencies. FlashAttention uses a fused low-precision kernel, while EnLA uses an unfused implementation with fp32 attention computation to stabilize the denominator reduction over tokens. The comparison therefore reflects both the attention operators and their implementation and precision differences. Despite using higher-precision arithmetic without kernel fusion, EnLA achieves higher throughput at both resolutions and scales more favorably: quadrupling the token count from 2048px to 4096px increases its measured latency by $4.09\times$, compared with $12.54\times$ for FlashAttention. This is consistent with linear versus quadratic attention scaling and shows that EnLA's throughput advantage grows at higher resolutions.

Figure~\ref{fig:supp_infer_full} reports throughput, peak memory, and GFLOPs across image resolutions, including ratios to vanilla linear attention.
Table~\ref{tab:supp_params_flops} complements these measurements with parameter counts and GFLOPs across DiT~\cite{Peebles_2023_ICCV} scales and resolutions.

\section{Additional Ablations}
\label{sec:supp_ablations}

\paragraph{Placement of SMN.}
We examine whether applying SMN elsewhere in the attention branch can replace KV enrichment. Two variants add residual SMN corrections to the linear-attention output before or after its output projection:
\begin{equation}
\begin{aligned}
Y_{\mathrm{pos1}} &= W_O\!\left(A+\mathrm{SMN}(A)\right),\\
Y_{\mathrm{pos2}} &= W_O(A)+\mathrm{SMN}\!\left(W_O(A)\right),
\end{aligned}
\end{equation}
where $A$ is the vanilla linear-attention output and $W_O$ is its output projection.
Table~\ref{tab:supp-smn-placement} compares the placements on DiT-S/2 at 256px. These variants obtain FIDs of 62.49 and 61.98 without guidance, and 38.78 and 38.10 with guidance (CFG $=1.5$), respectively. EnLA achieves lower FIDs of 61.15 and 37.71, supporting enrichment of keys and values before aggregation.

\paragraph{Token-Dependent Corrections.}
At blocks 11, 18, and 26, we replace the key and value corrections with their image-dependent token means or permute them across positions to test whether their token assignment matters.
Both interventions increase denoising error, with permutation causing larger increases at all three blocks (Figure~\ref{fig:supp_token_controls}).
These results support the benefit of token-specific corrections applied at their corresponding positions.

\begin{figure*}[!b]
\centering
\includegraphics[width=\textwidth]{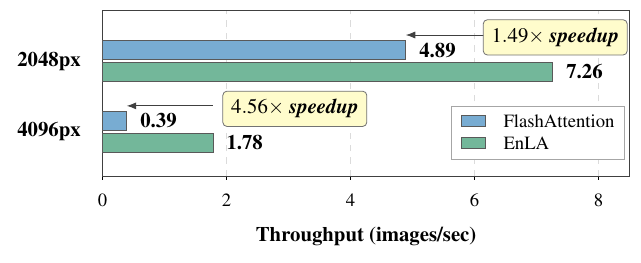}
\caption{Inference throughput of DiT-XL/2 at batch size 8 on a single NVIDIA H200. FlashAttention~\cite{dao2022flashattention} uses a fused low-precision softmax-attention kernel, whereas EnLA uses an unfused implementation with its attention computation in fp32, including the denominator reduction over tokens for numerical stability. Thus, the comparison includes differences in both attention operators and their kernel implementations and precision. Despite using higher-precision arithmetic without kernel fusion, EnLA achieves $1.49\times$ and $4.56\times$ speedups at $2048\times2048$ and $4096\times4096$, respectively, computed from median latencies. Doubling image width and height quadruples the token count; measured latency increases by $4.09\times$ for EnLA and $12.54\times$ for FlashAttention, consistent with their linear and quadratic attention scaling, respectively.}
\label{fig:supp-flash-throughput}
\end{figure*}

\newpage
\section{Additional Image Classification Results}
\label{sec:supp_classification}
Table~\ref{tab:sota-in1k} provides additional ImageNet-1K~\cite{deng2009imagenet} classification results using a non-DeiT~\cite{touvron2021deit} vision transformer (ViT)~\cite{dosovitskiy2021vit} architectures. EnLA achieves 82.5\% and 84.3\% top-1 accuracy, outperforming the listed models in both size groups.

\vspace{8pt}
\noindent\begin{minipage}[t]{\columnwidth}
\centering
\captionof{table}{SMN placement ablation on DiT-S/2 at 256px. Pos1 and pos2 apply SMN before and after the attention output projection, respectively. Best values within each guidance setting are bold.}
\vspace{-5pt}
\label{tab:supp-smn-placement}
\normalsize
\setlength{\tabcolsep}{7pt}
\renewcommand{\arraystretch}{1.05}
\begin{tabular}{@{}clccc@{}}
\toprule
Guidance & Placement & IS $\uparrow$ & FID $\downarrow$ & sFID $\downarrow$ \\
\midrule
\multirow{3}{*}{Without CFG} & Pos1 & 22.58 & 62.49 & 12.06 \\
 & Pos2 & 22.73 & 61.98 & 11.86 \\
 & EnLA & \textbf{22.93} & \textbf{61.15} & \textbf{11.43} \\
\midrule
\multirow{3}{*}{CFG $=1.5$} & Pos1 & 40.30 & 38.78 & 8.98 \\
 & Pos2 & 40.64 & 38.10 & 8.83 \\
 & EnLA & \textbf{41.14} & \textbf{37.71} & \textbf{8.56} \\
\bottomrule
\end{tabular}
\vspace{-5pt}
\end{minipage}

\clearpage
\begin{figure*}[p]
\centering
\begin{minipage}[t]{.49\textwidth}
\centering
{\small\textbf{Block 1}}\par\vspace{2pt}
\includegraphics[width=\linewidth]{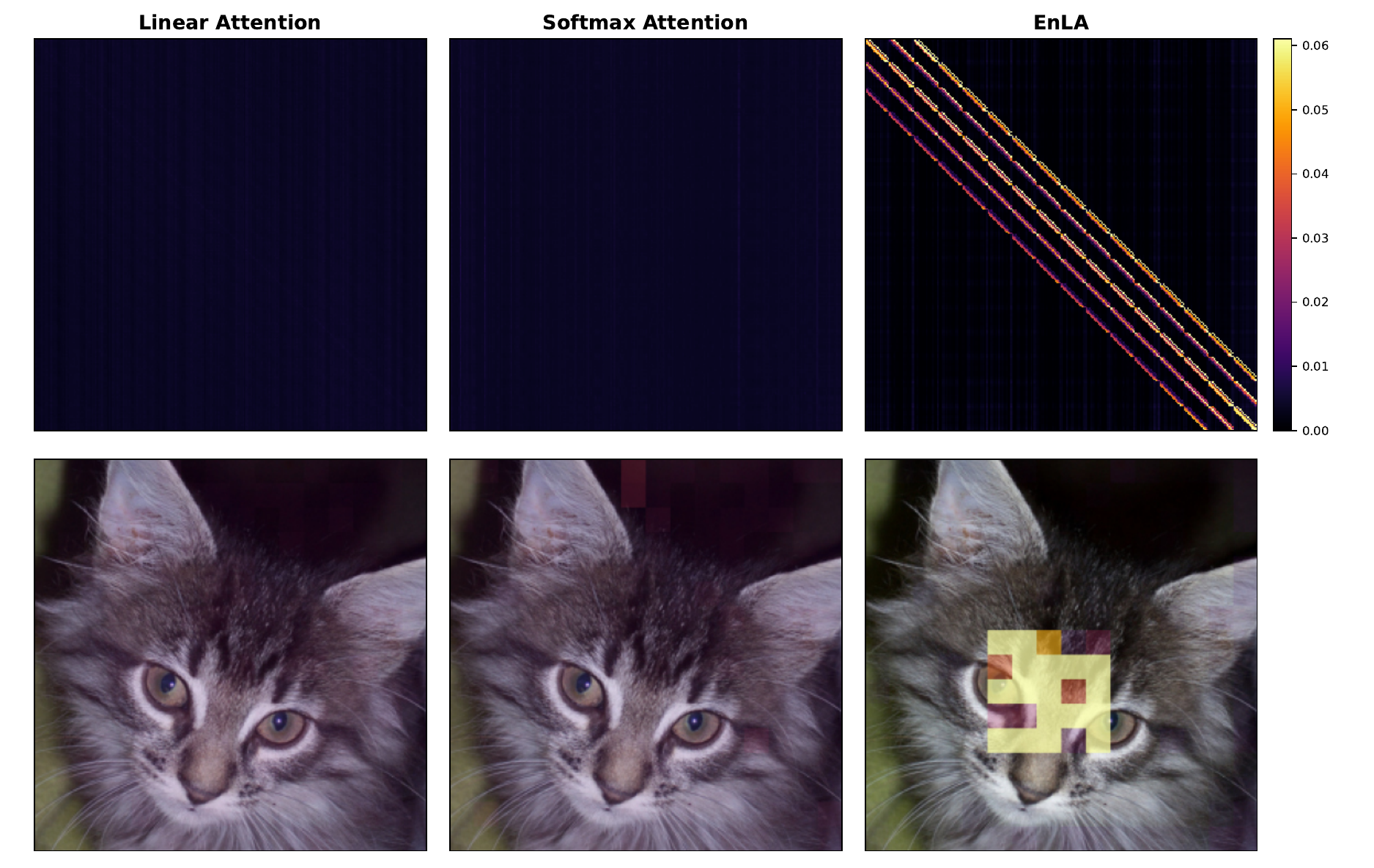}
\end{minipage}
\hfill
\begin{minipage}[t]{.49\textwidth}
\centering
{\small\textbf{Block 2}}\par\vspace{2pt}
\includegraphics[width=\linewidth]{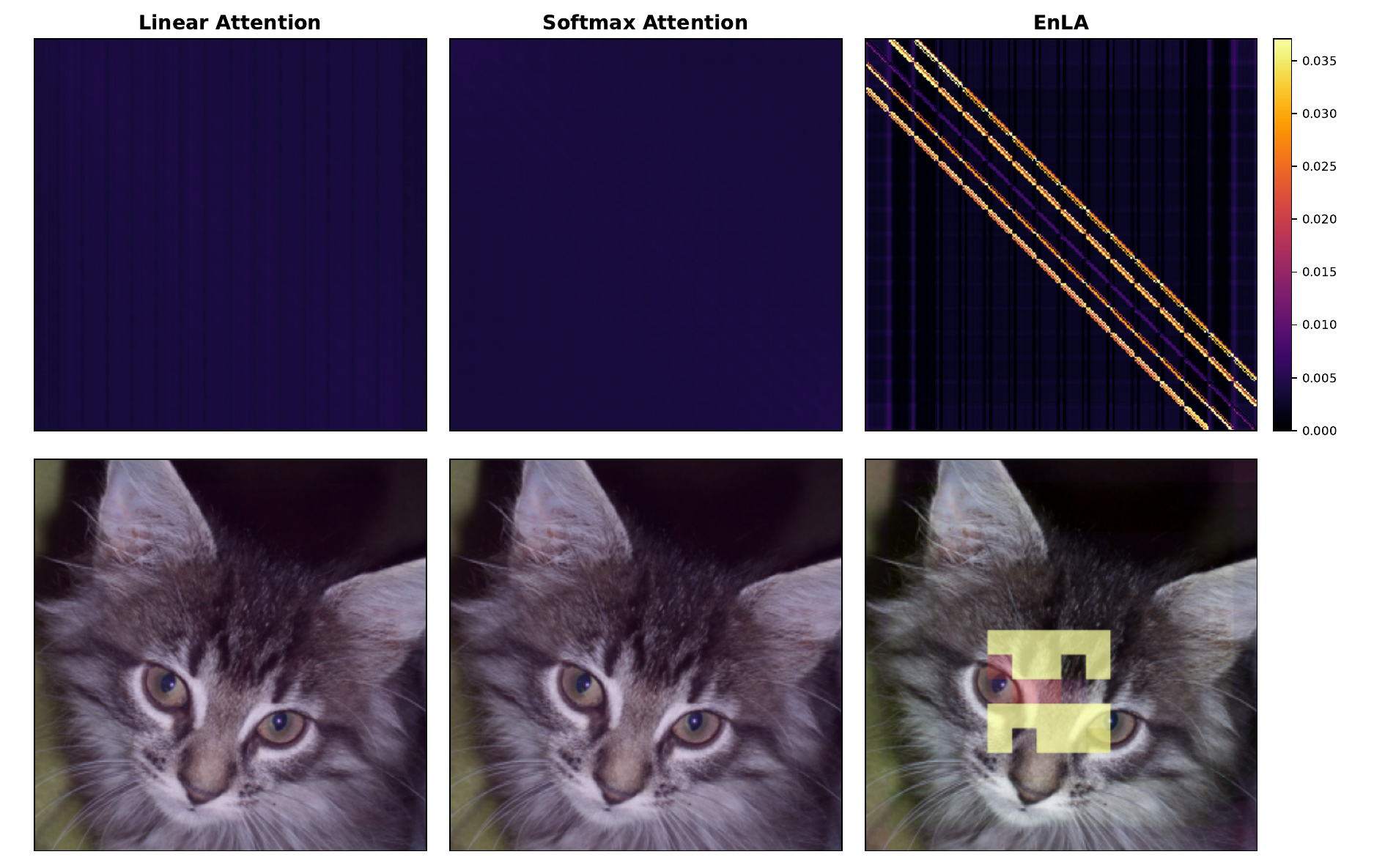}
\end{minipage}
\par\vspace{6pt}
\begin{minipage}[t]{.49\textwidth}
\centering
{\small\textbf{Block 3}}\par\vspace{2pt}
\includegraphics[width=\linewidth]{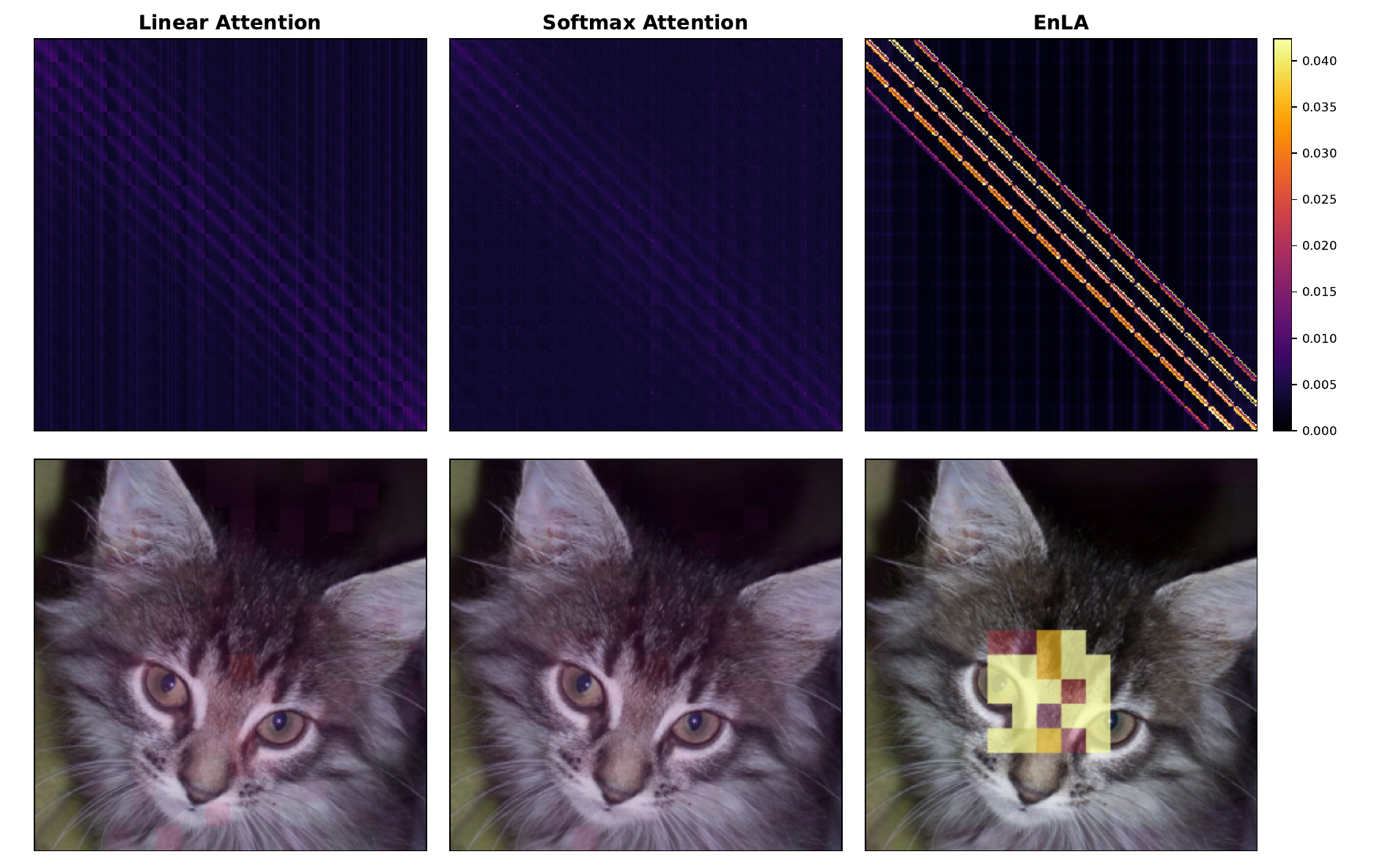}
\end{minipage}
\hfill
\begin{minipage}[t]{.49\textwidth}
\centering
{\small\textbf{Block 4}}\par\vspace{2pt}
\includegraphics[width=\linewidth]{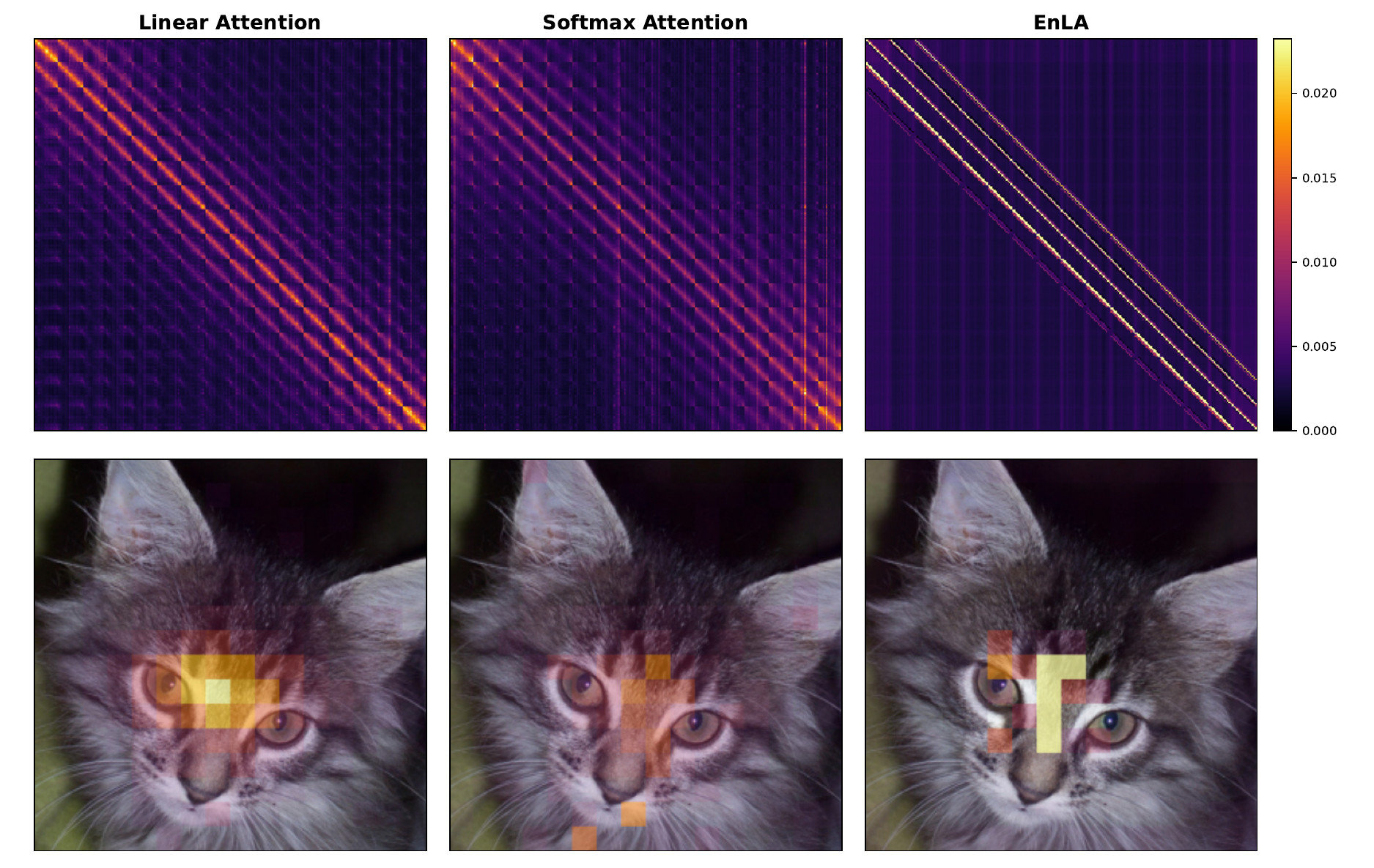}
\end{minipage}
\par\vspace{6pt}
\begin{minipage}[t]{.49\textwidth}
\centering
{\small\textbf{Block 5}}\par\vspace{2pt}
\includegraphics[width=\linewidth]{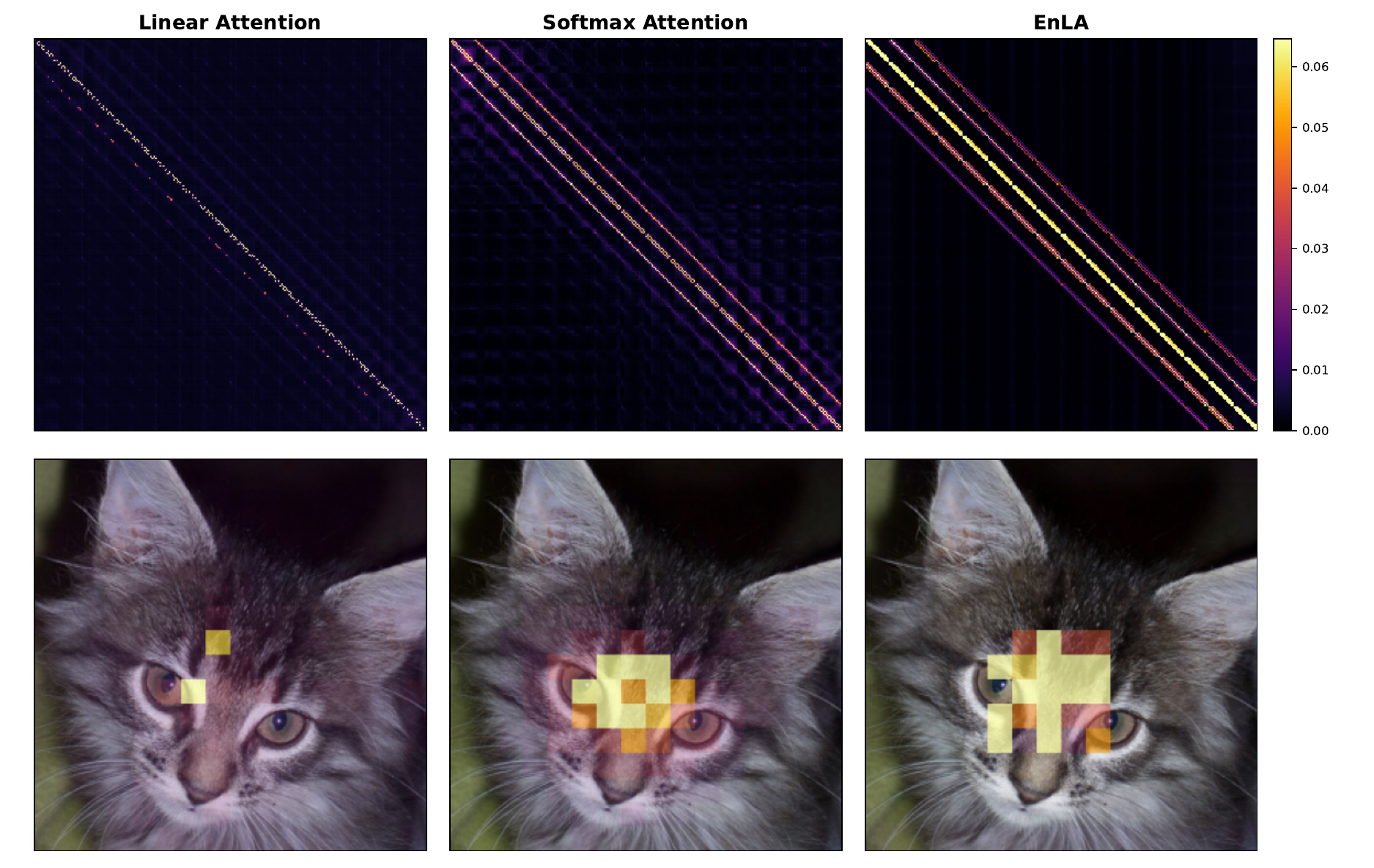}
\end{minipage}
\hfill
\begin{minipage}[t]{.49\textwidth}
\centering
{\small\textbf{Block 6}}\par\vspace{2pt}
\includegraphics[width=\linewidth]{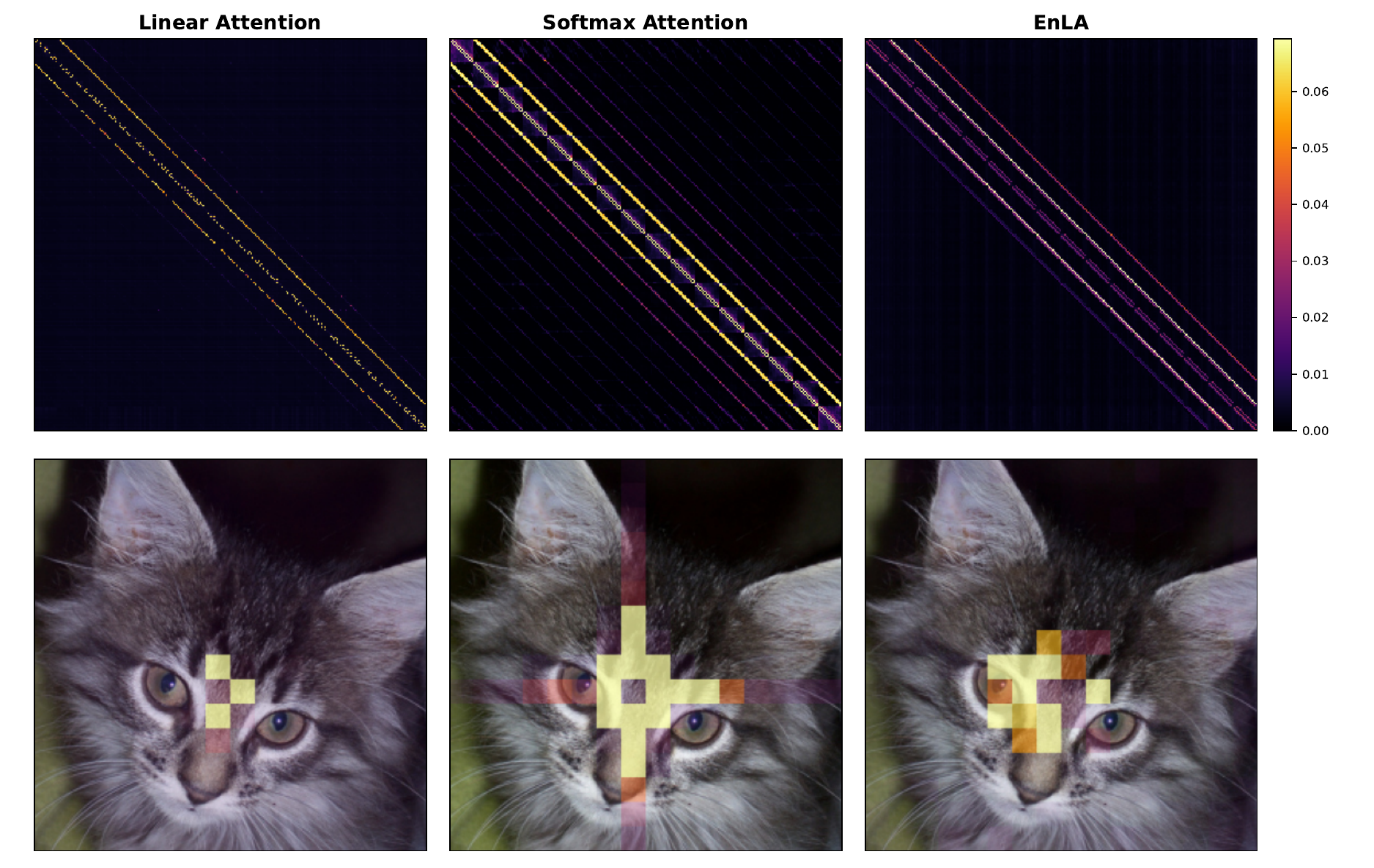}
\end{minipage}
\par\vspace{6pt}
\caption{Block-wise extension of the left part of the main paper's Figure~1: effective token-mixing operators in DiT-XL/2 at $256\times256$ resolution and $t=500$, blocks 1--6. Each panel compares linear attention, softmax attention, and EnLA from left to right. Top: row-normalized absolute magnitudes averaged over heads, with a shared color scale within each block. Bottom: the selected query's spatial magnitude map overlaid on the input image.}
\label{fig:supp_attention_blocks_1}
\end{figure*}
\begin{figure*}[p]
\centering
\begin{minipage}[t]{.49\textwidth}
\centering
{\small\textbf{Block 7}}\par\vspace{2pt}
\includegraphics[width=\linewidth]{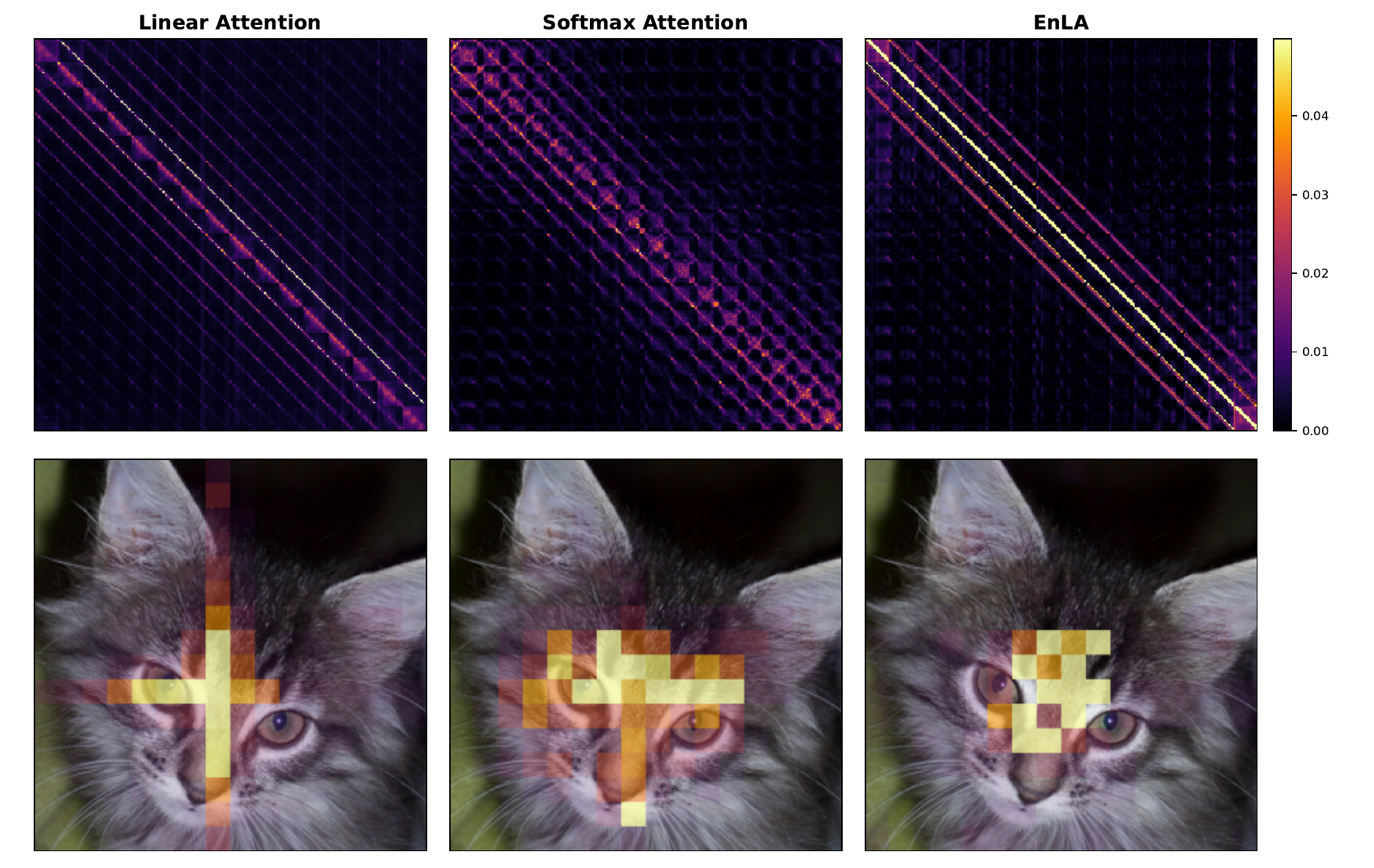}
\end{minipage}
\hfill
\begin{minipage}[t]{.49\textwidth}
\centering
{\small\textbf{Block 8}}\par\vspace{2pt}
\includegraphics[width=\linewidth]{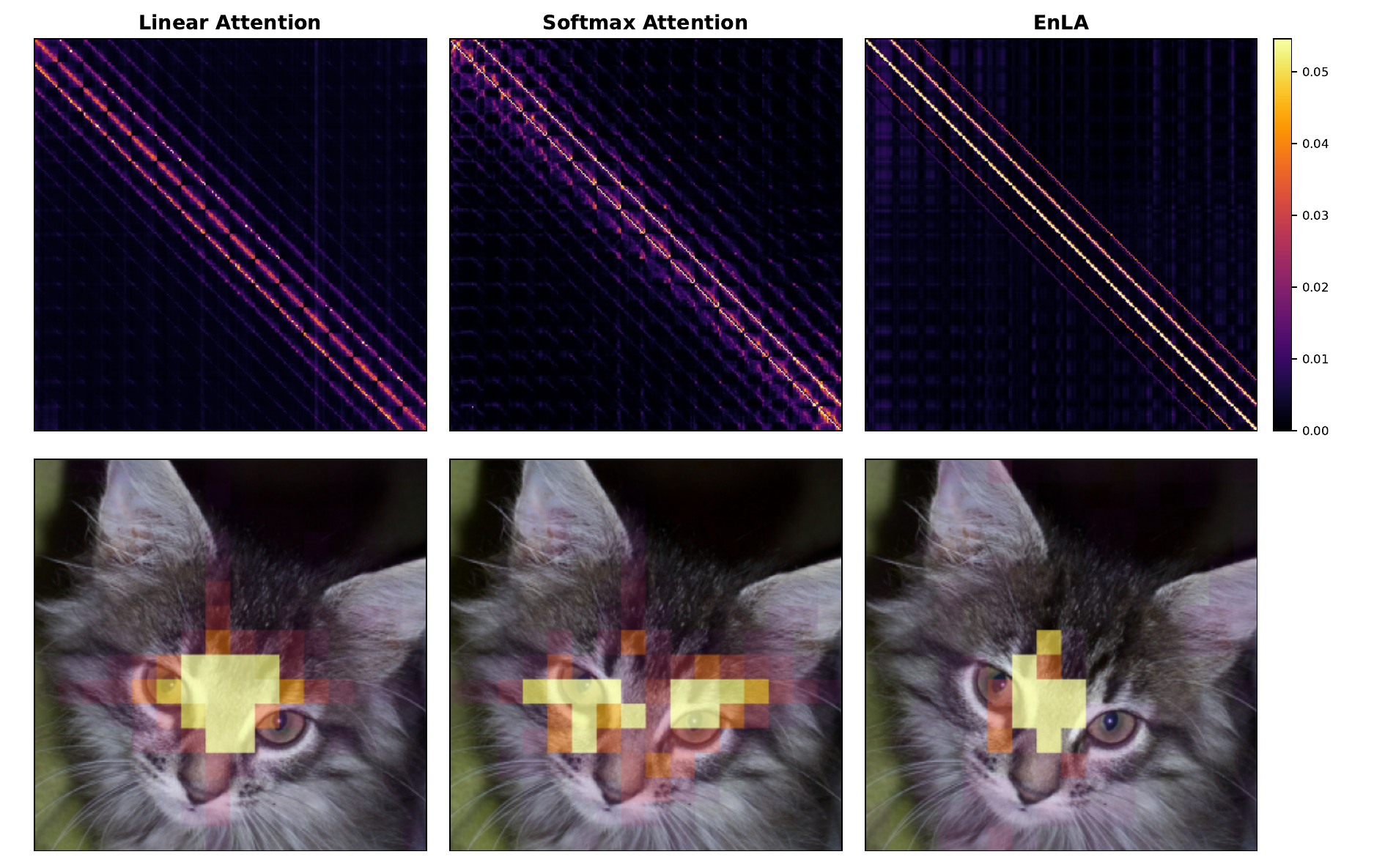}
\end{minipage}
\par\vspace{6pt}
\begin{minipage}[t]{.49\textwidth}
\centering
{\small\textbf{Block 9}}\par\vspace{2pt}
\includegraphics[width=\linewidth]{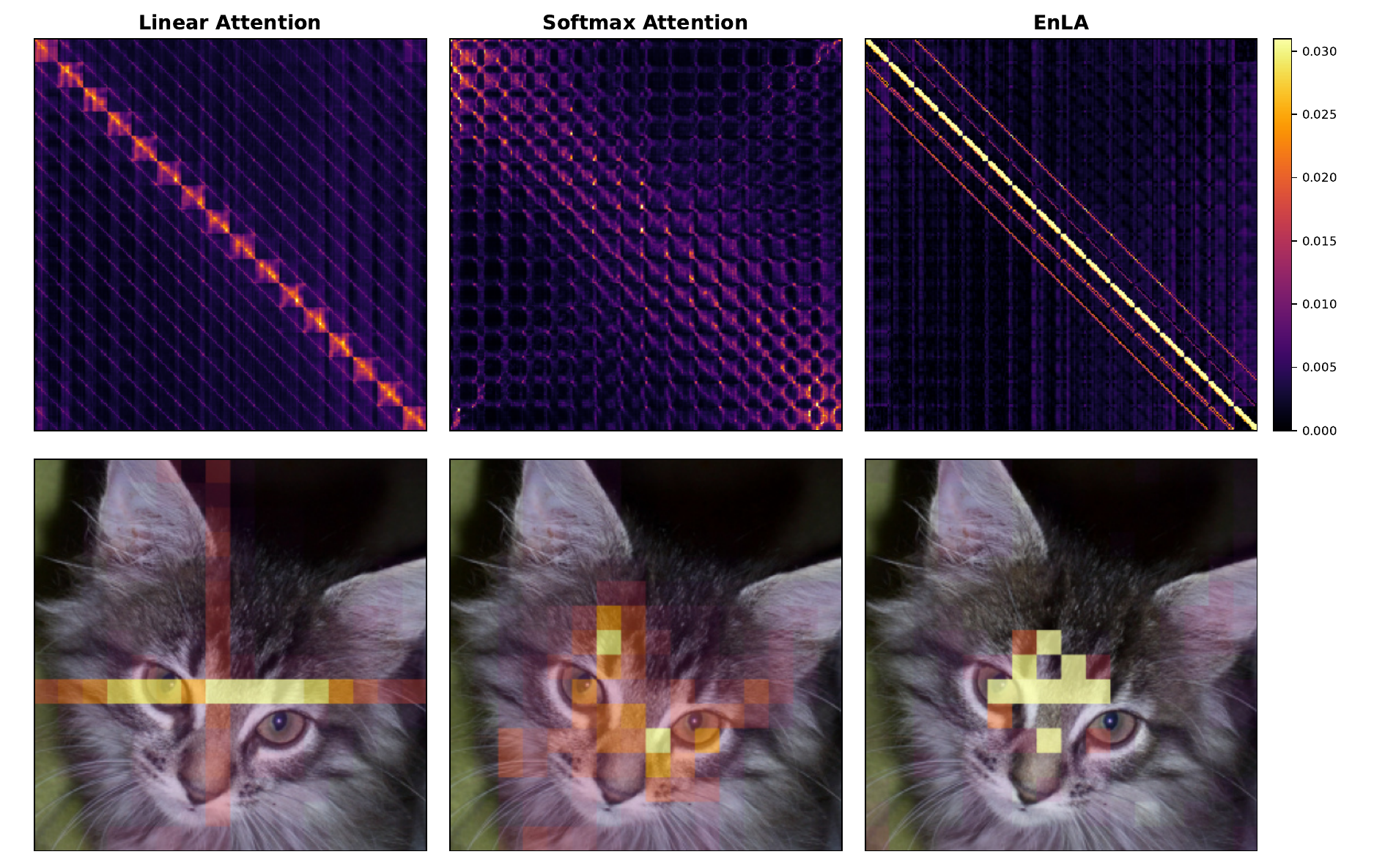}
\end{minipage}
\hfill
\begin{minipage}[t]{.49\textwidth}
\centering
{\small\textbf{Block 10}}\par\vspace{2pt}
\includegraphics[width=\linewidth]{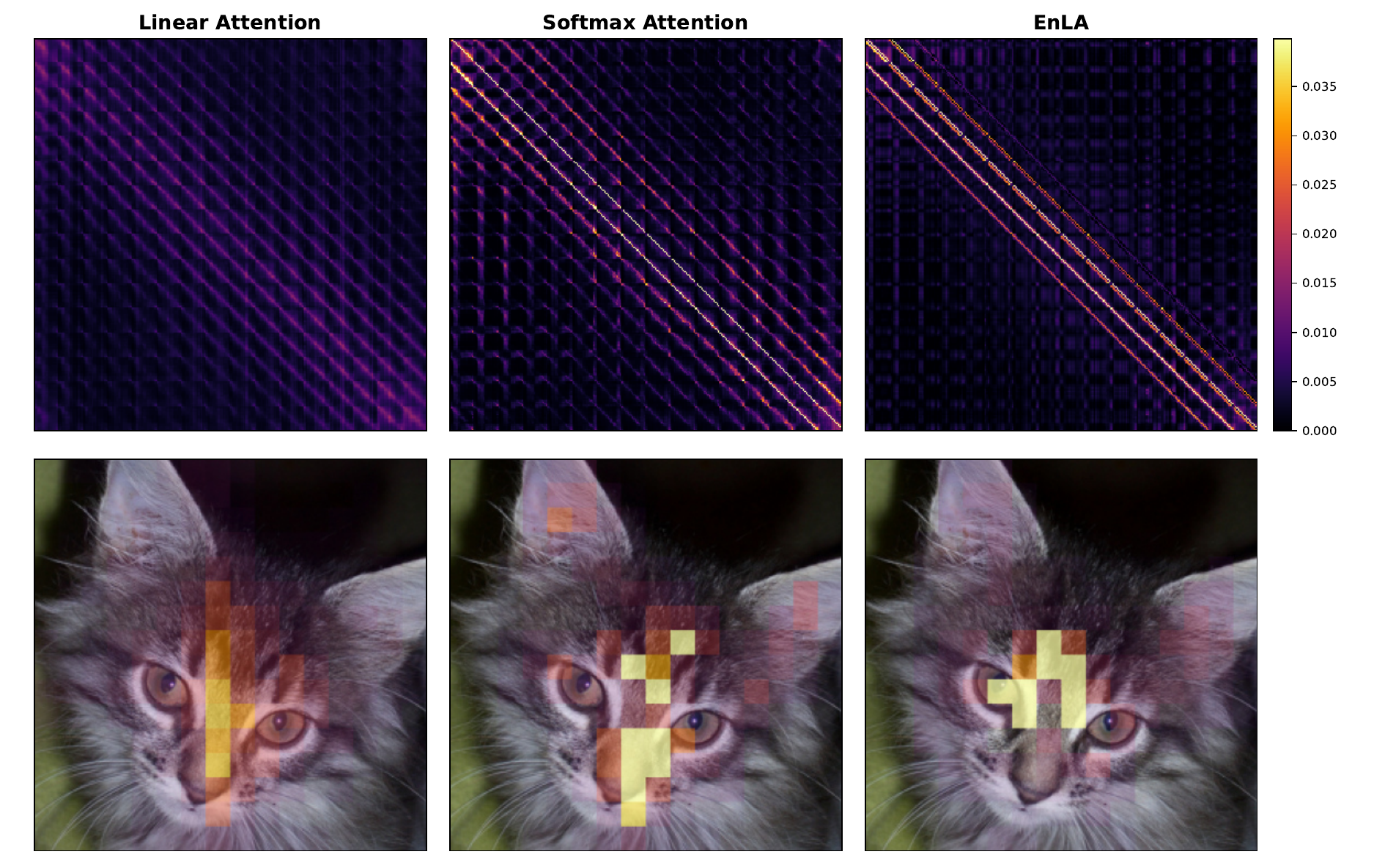}
\end{minipage}
\par\vspace{6pt}
\begin{minipage}[t]{.49\textwidth}
\centering
{\small\textbf{Block 11}}\par\vspace{2pt}
\includegraphics[width=\linewidth]{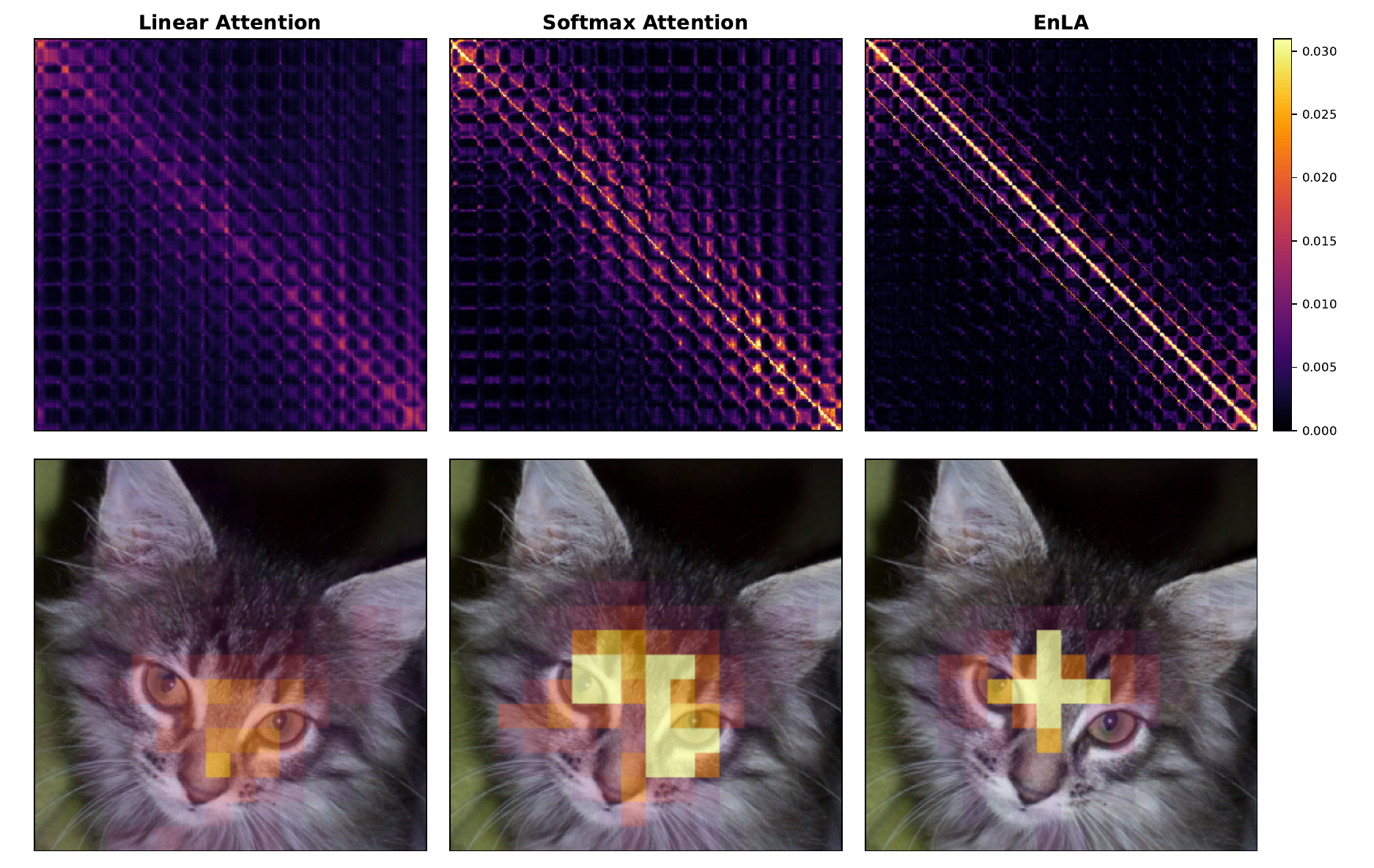}
\end{minipage}
\hfill
\begin{minipage}[t]{.49\textwidth}
\centering
{\small\textbf{Block 12}}\par\vspace{2pt}
\includegraphics[width=\linewidth]{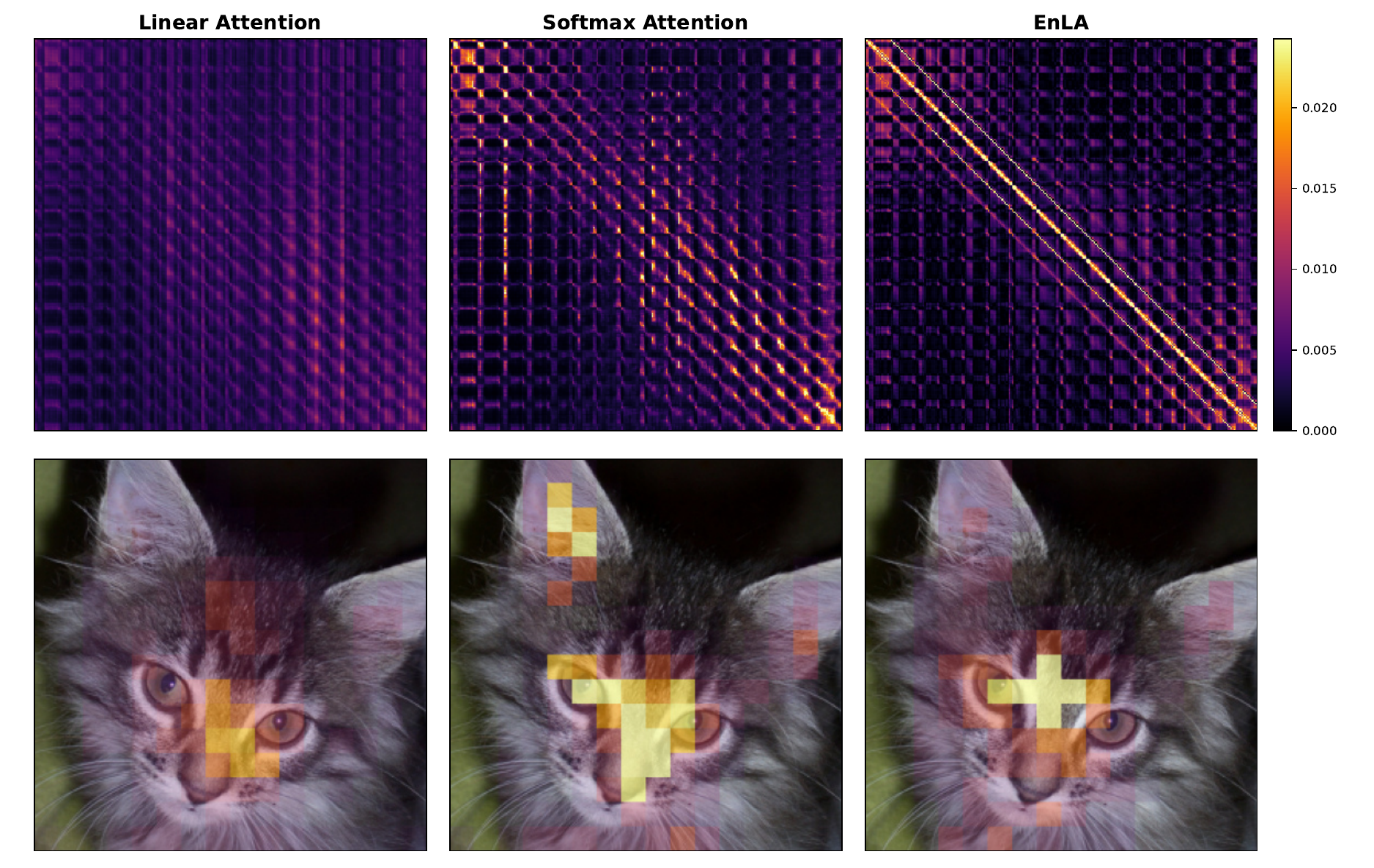}
\end{minipage}
\par\vspace{6pt}
\caption{Block-wise extension of the left part of the main paper's Figure~1: effective token-mixing operators in DiT-XL/2 at $256\times256$ resolution and $t=500$, blocks 7--12. Each panel compares linear attention, softmax attention, and EnLA from left to right. Top: row-normalized absolute magnitudes averaged over heads, with a shared color scale within each block. Bottom: the selected query's spatial magnitude map overlaid on the input image.}
\label{fig:supp_attention_blocks_2}
\end{figure*}
\begin{figure*}[p]
\centering
\begin{minipage}[t]{.49\textwidth}
\centering
{\small\textbf{Block 13}}\par\vspace{2pt}
\includegraphics[width=\linewidth]{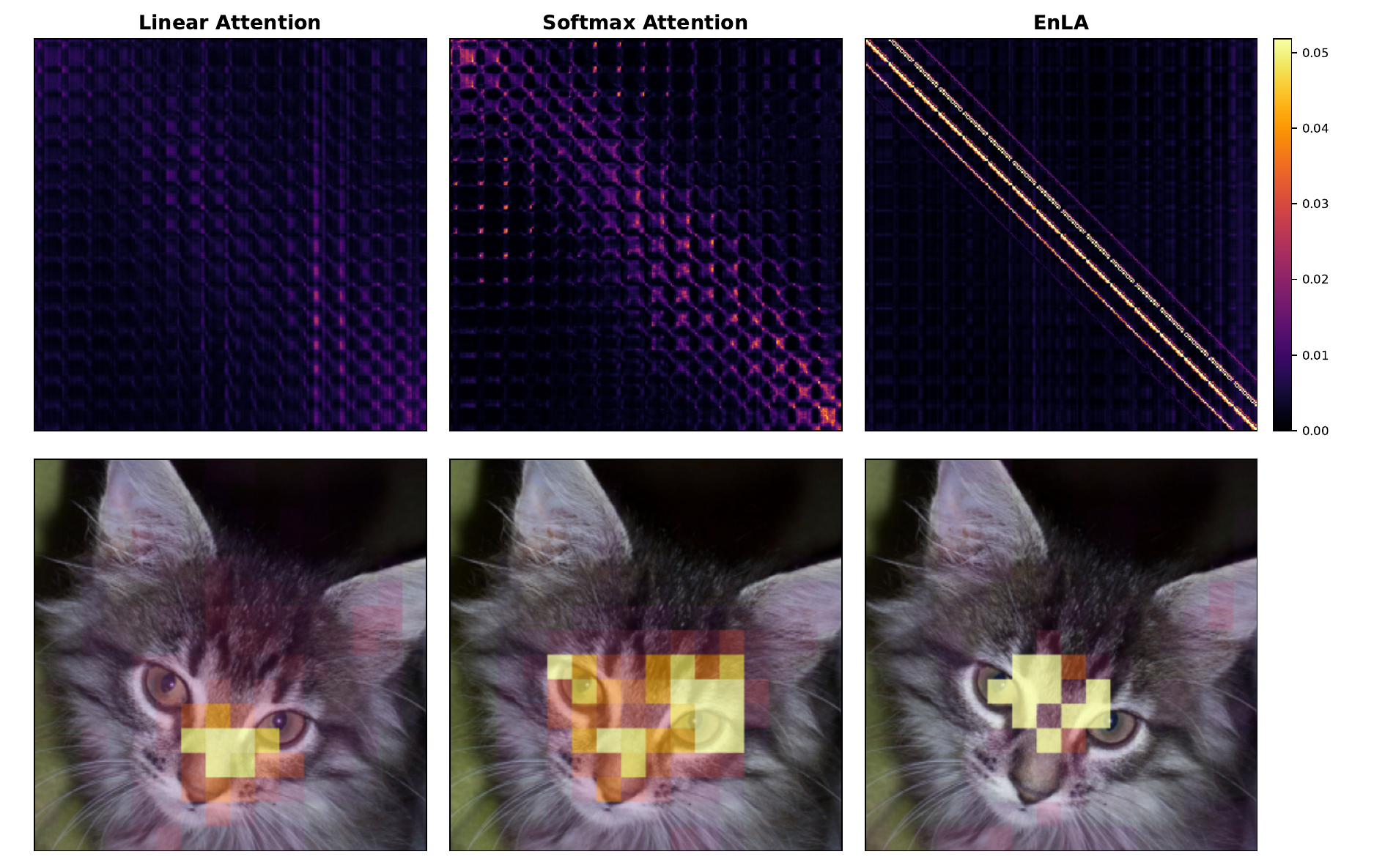}
\end{minipage}
\hfill
\begin{minipage}[t]{.49\textwidth}
\centering
{\small\textbf{Block 14}}\par\vspace{2pt}
\includegraphics[width=\linewidth]{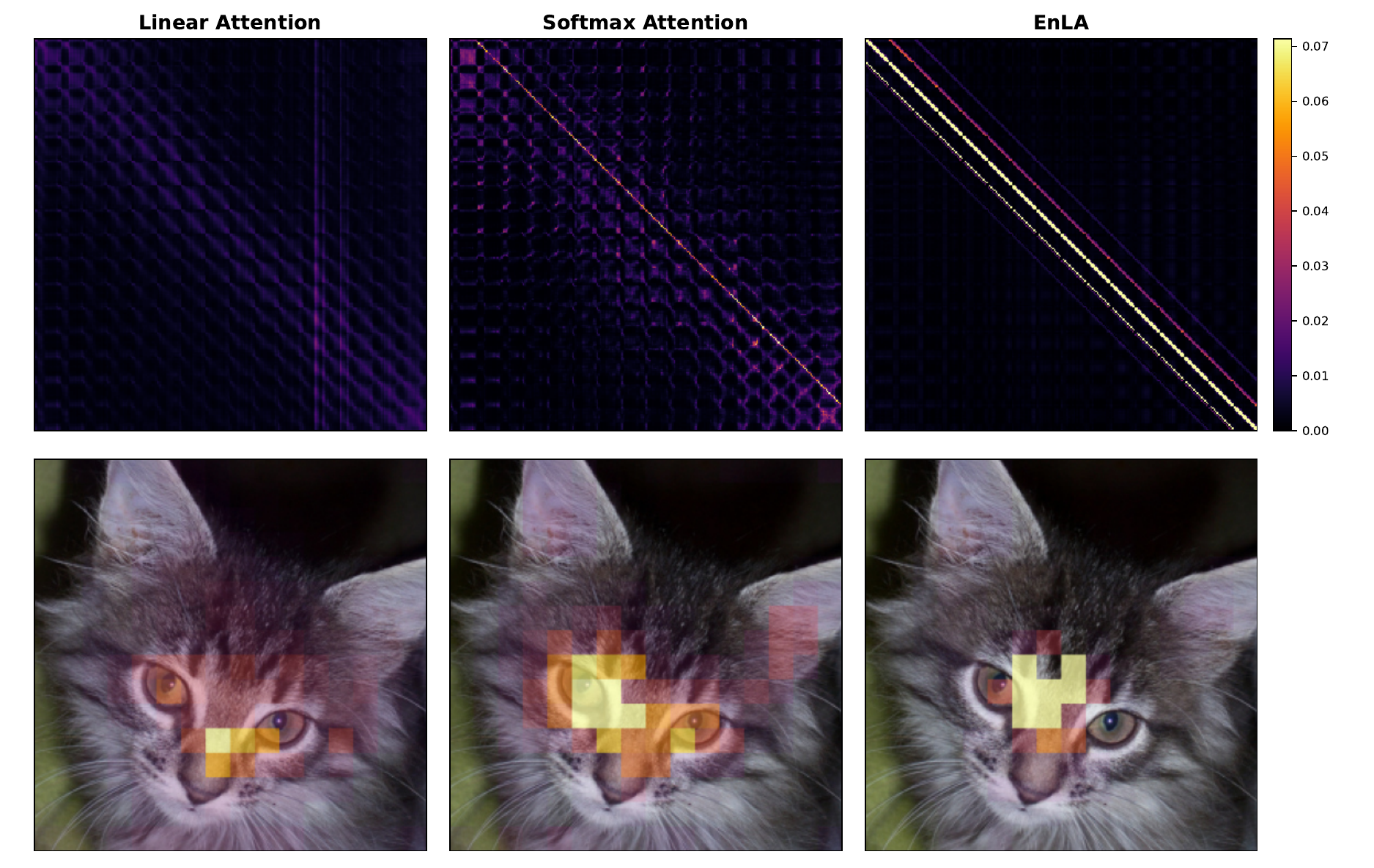}
\end{minipage}
\par\vspace{6pt}
\begin{minipage}[t]{.49\textwidth}
\centering
{\small\textbf{Block 15}}\par\vspace{2pt}
\includegraphics[width=\linewidth]{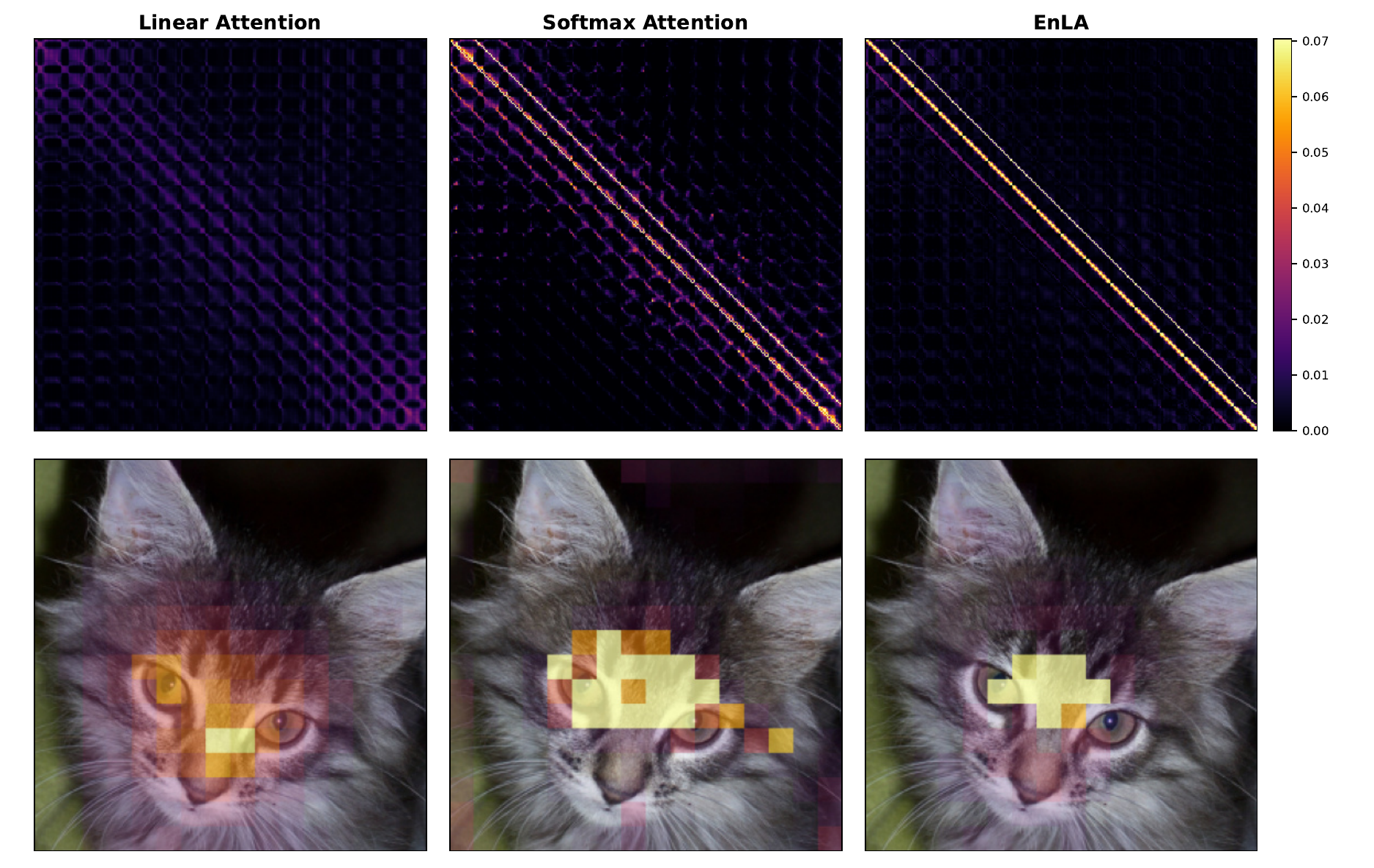}
\end{minipage}
\hfill
\begin{minipage}[t]{.49\textwidth}
\centering
{\small\textbf{Block 16}}\par\vspace{2pt}
\includegraphics[width=\linewidth]{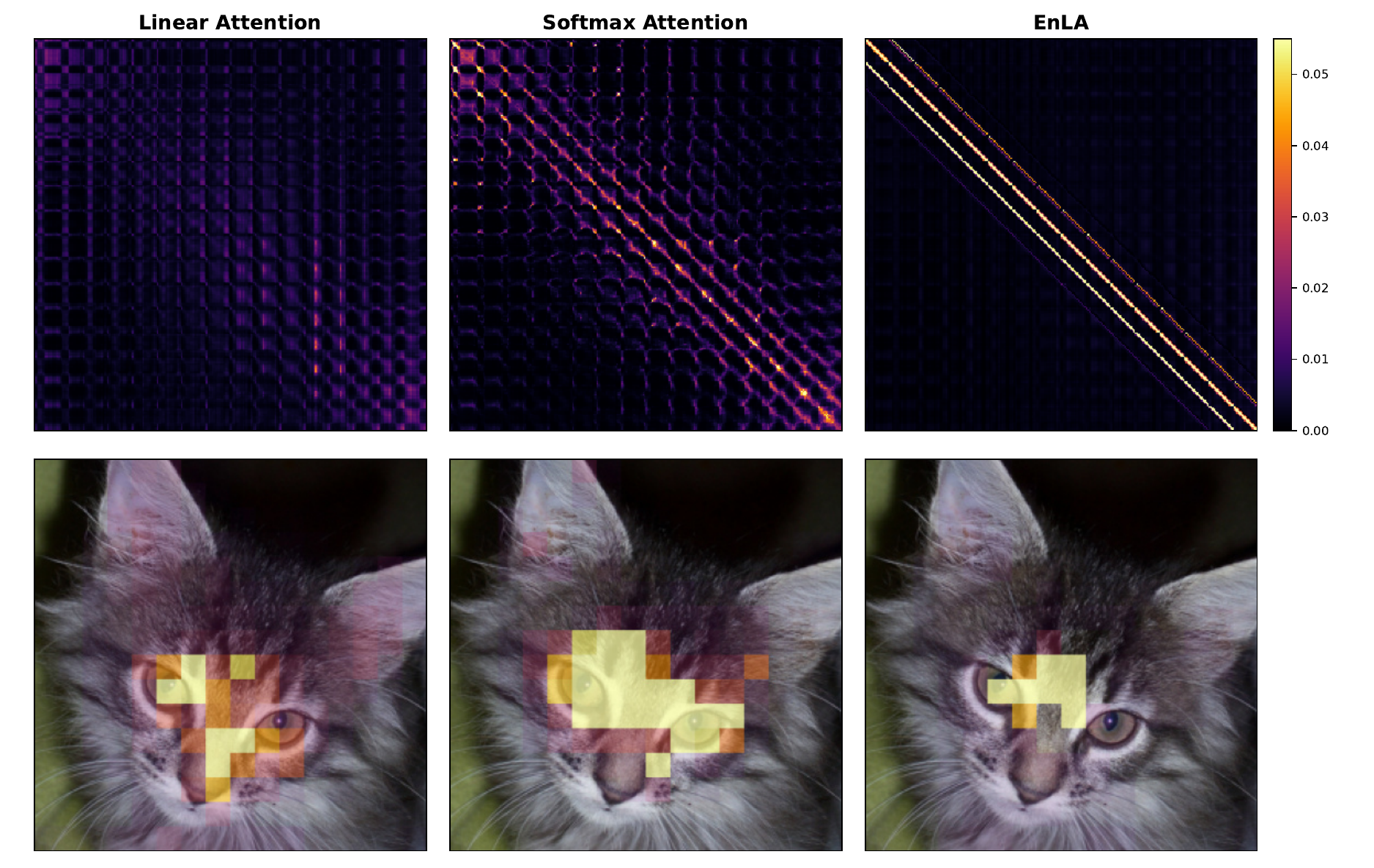}
\end{minipage}
\par\vspace{6pt}
\begin{minipage}[t]{.49\textwidth}
\centering
{\small\textbf{Block 17}}\par\vspace{2pt}
\includegraphics[width=\linewidth]{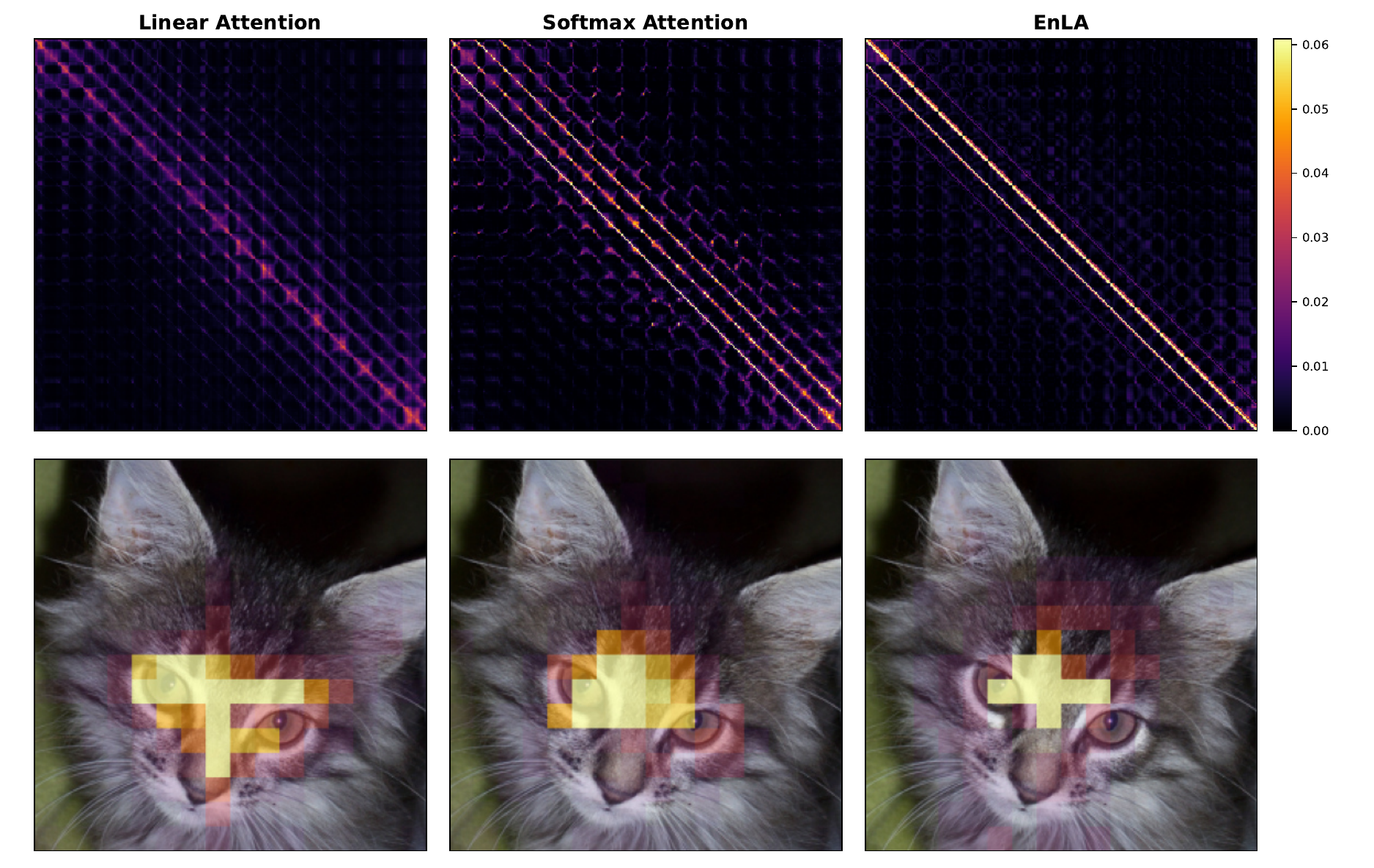}
\end{minipage}
\hfill
\begin{minipage}[t]{.49\textwidth}
\centering
{\small\textbf{Block 18}}\par\vspace{2pt}
\includegraphics[width=\linewidth]{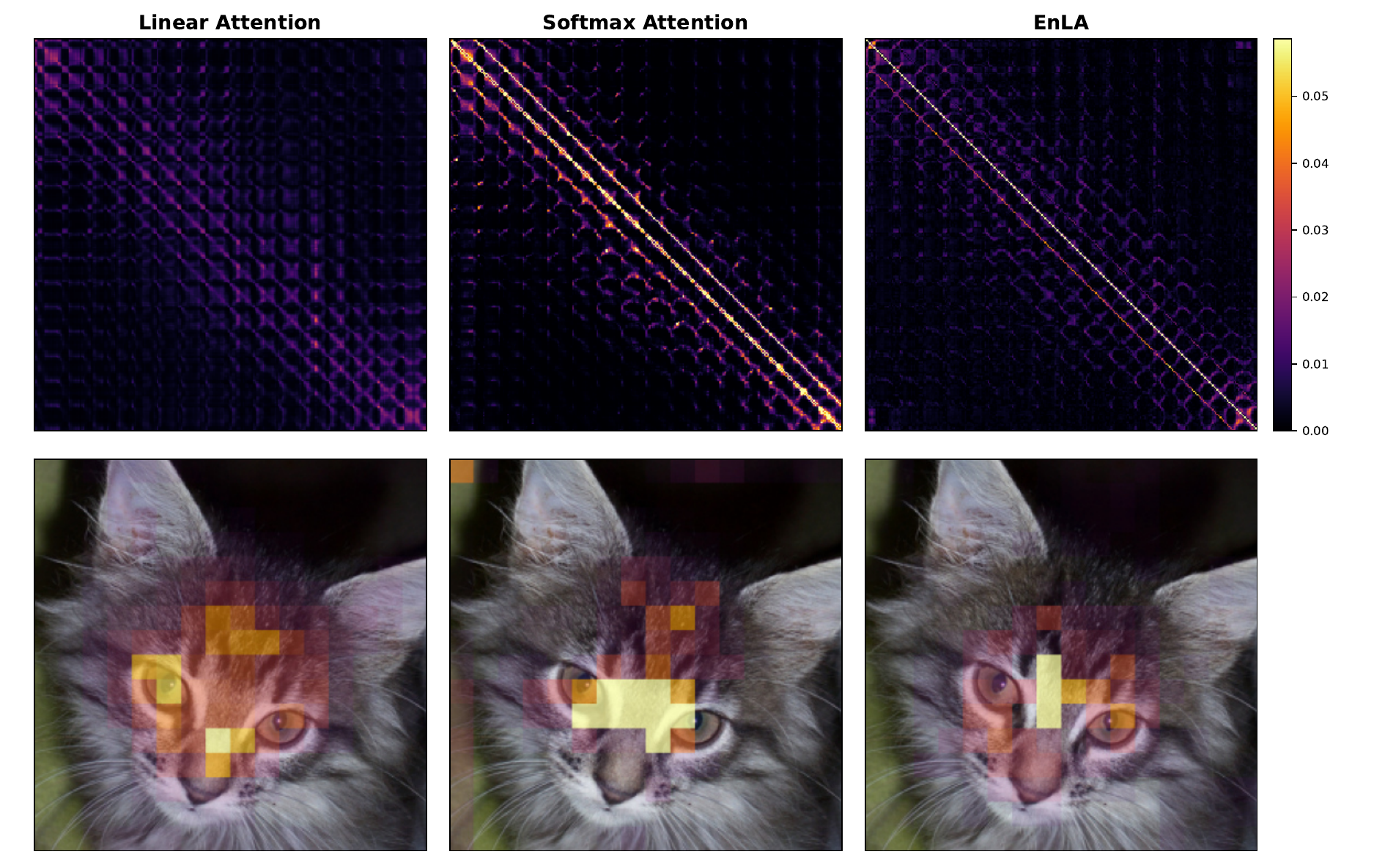}
\end{minipage}
\par\vspace{6pt}
\caption{Block-wise extension of the left part of the main paper's Figure~1: effective token-mixing operators in DiT-XL/2 at $256\times256$ resolution and $t=500$, blocks 13--18. Each panel compares linear attention, softmax attention, and EnLA from left to right. Top: row-normalized absolute magnitudes averaged over heads, with a shared color scale within each block. Bottom: the selected query's spatial magnitude map overlaid on the input image.}
\label{fig:supp_attention_blocks_3}
\end{figure*}
\begin{figure*}[p]
\centering
\begin{minipage}[t]{.49\textwidth}
\centering
{\small\textbf{Block 19}}\par\vspace{2pt}
\includegraphics[width=\linewidth]{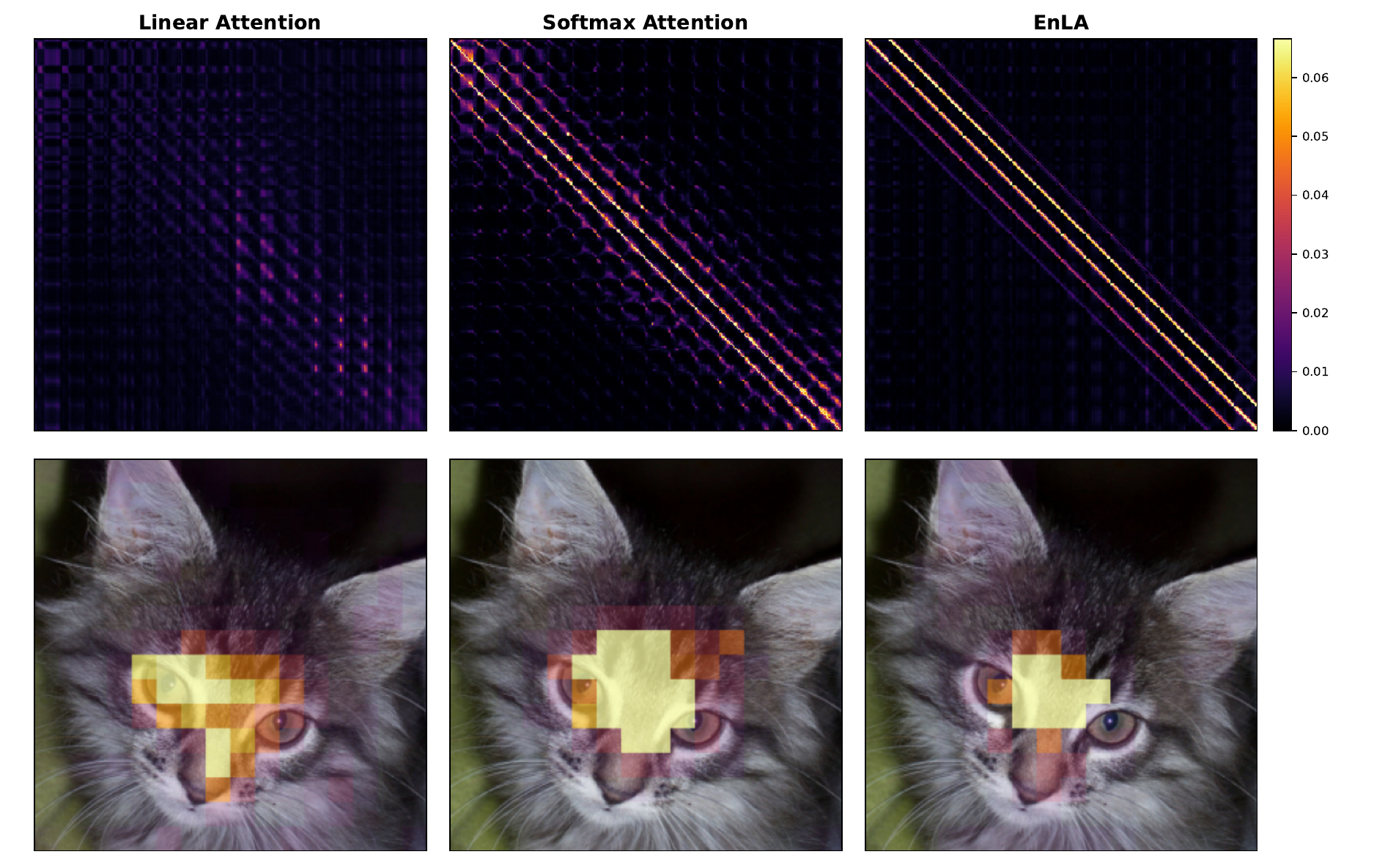}
\end{minipage}
\hfill
\begin{minipage}[t]{.49\textwidth}
\centering
{\small\textbf{Block 20}}\par\vspace{2pt}
\includegraphics[width=\linewidth]{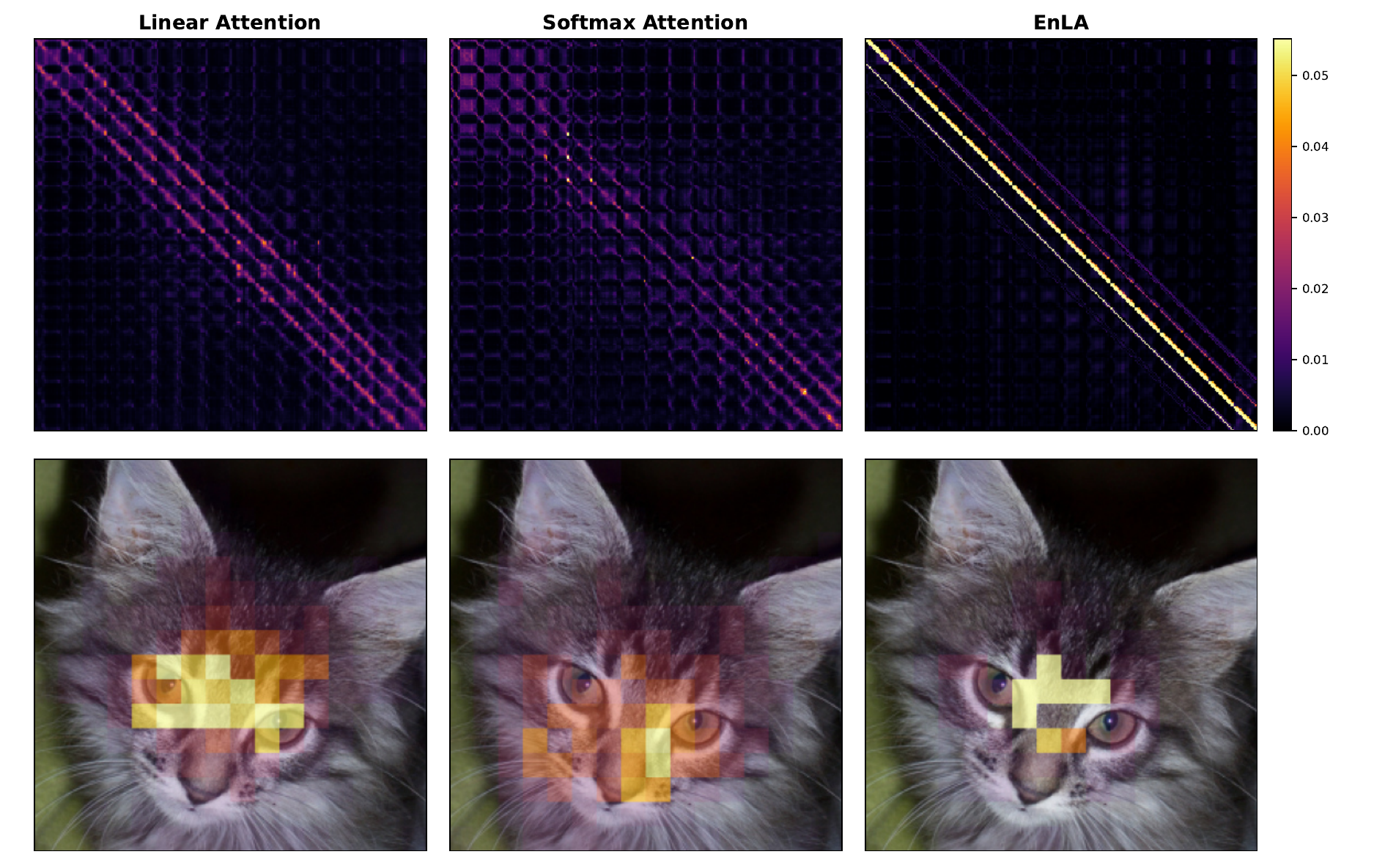}
\end{minipage}
\par\vspace{6pt}
\begin{minipage}[t]{.49\textwidth}
\centering
{\small\textbf{Block 21}}\par\vspace{2pt}
\includegraphics[width=\linewidth]{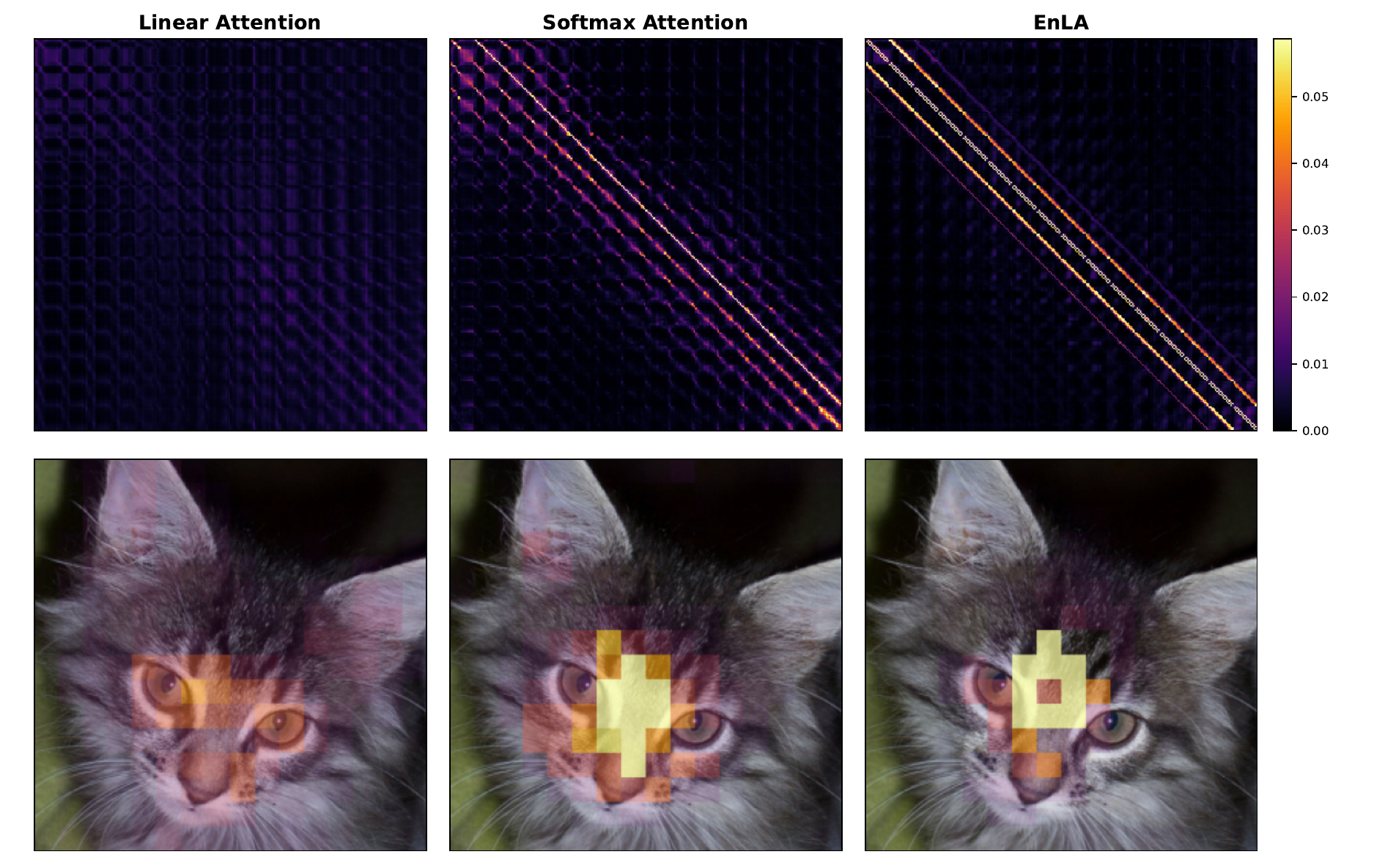}
\end{minipage}
\hfill
\begin{minipage}[t]{.49\textwidth}
\centering
{\small\textbf{Block 22}}\par\vspace{2pt}
\includegraphics[width=\linewidth]{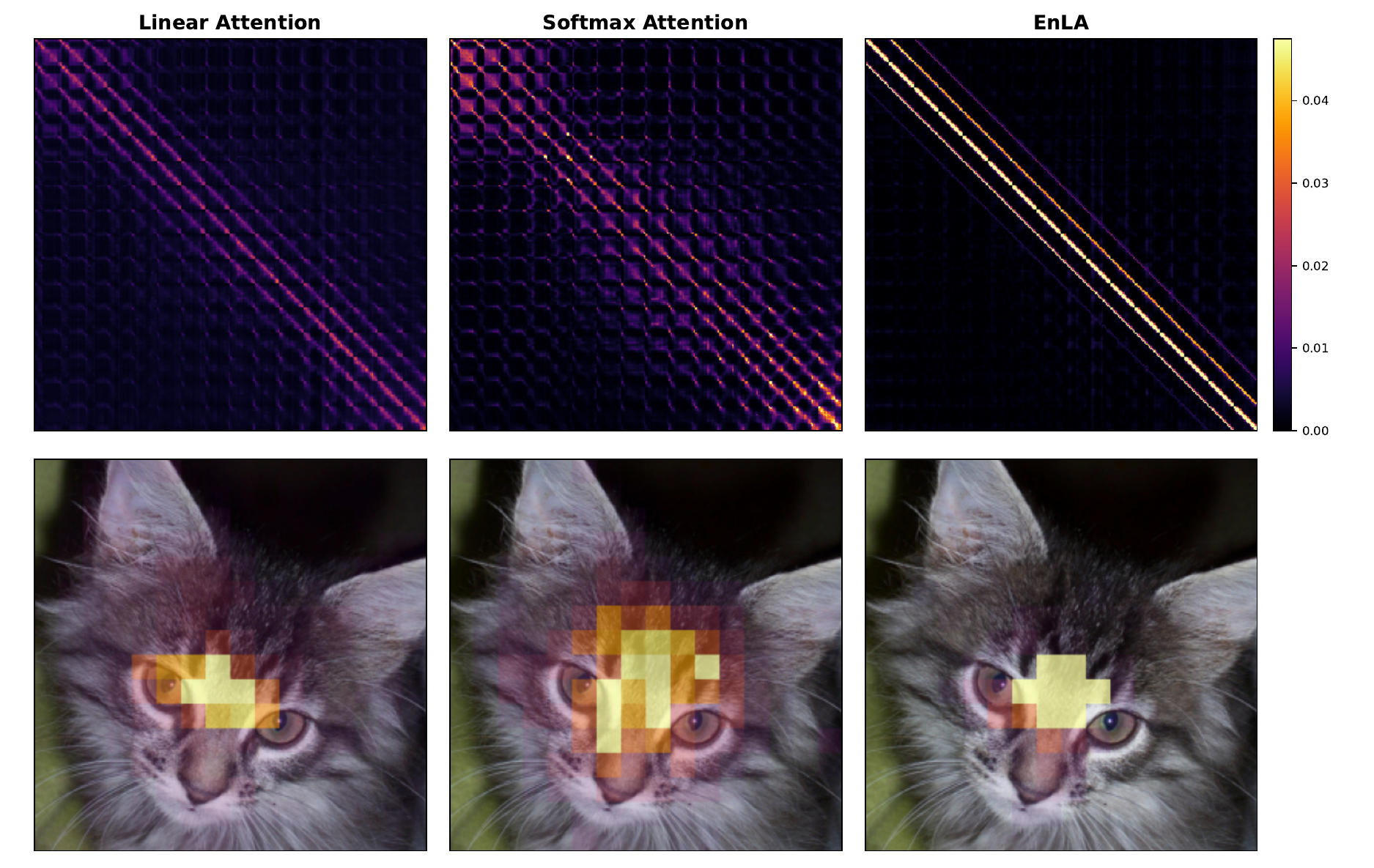}
\end{minipage}
\par\vspace{6pt}
\begin{minipage}[t]{.49\textwidth}
\centering
{\small\textbf{Block 23}}\par\vspace{2pt}
\includegraphics[width=\linewidth]{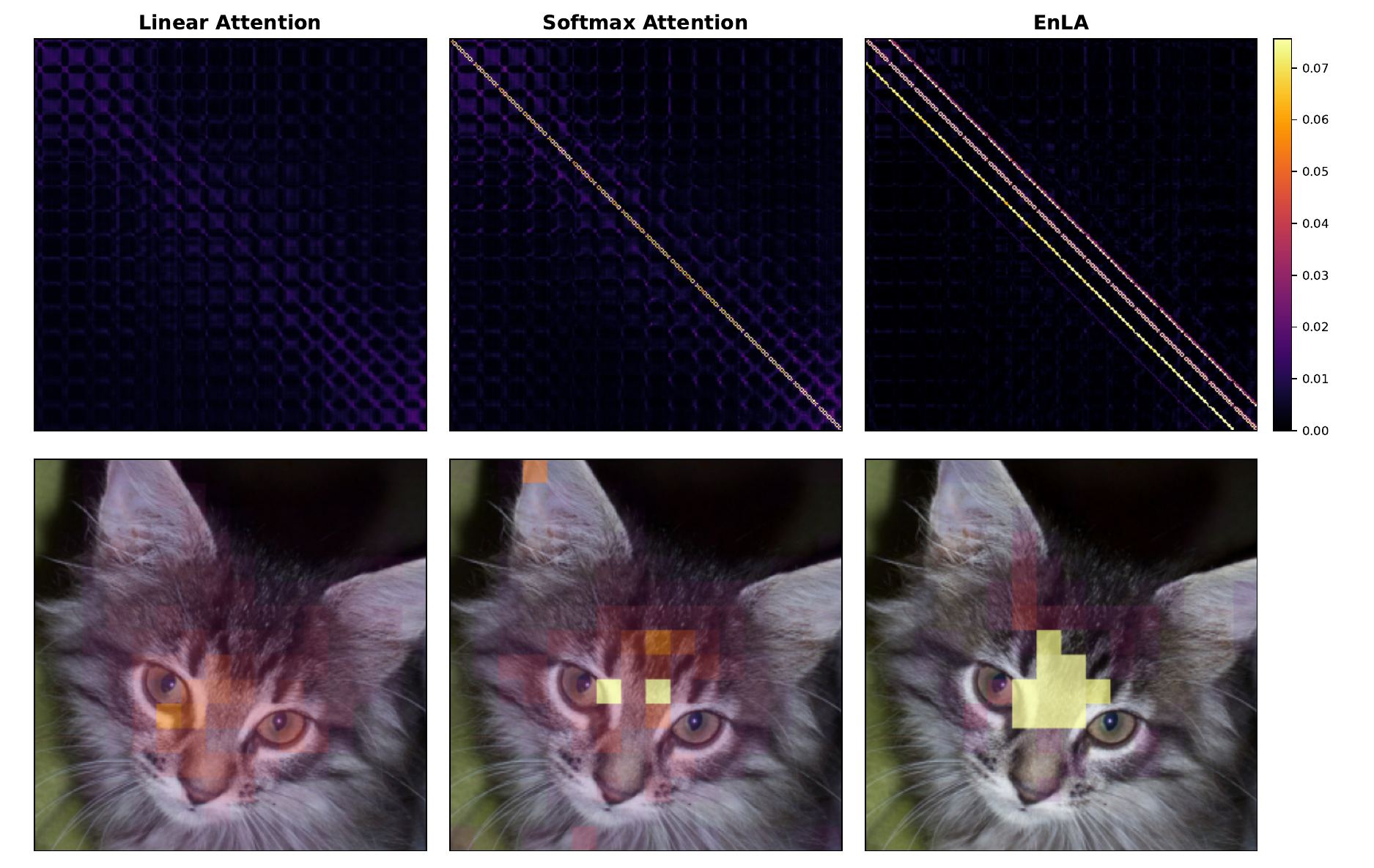}
\end{minipage}
\hfill
\begin{minipage}[t]{.49\textwidth}
\centering
{\small\textbf{Block 24}}\par\vspace{2pt}
\includegraphics[width=\linewidth]{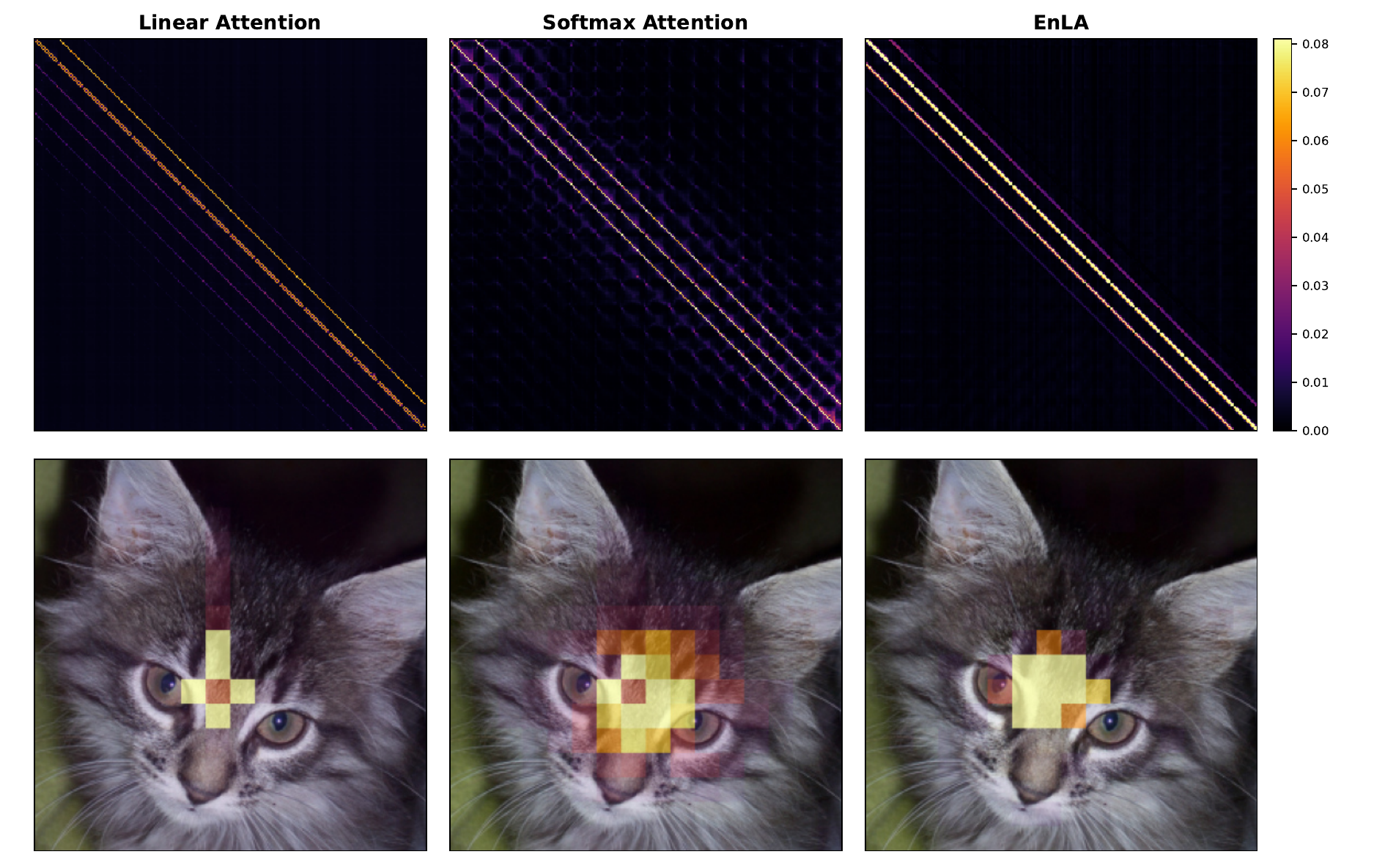}
\end{minipage}
\par\vspace{6pt}
\caption{Block-wise extension of the left part of the main paper's Figure~1: effective token-mixing operators in DiT-XL/2 at $256\times256$ resolution and $t=500$, blocks 19--24. Each panel compares linear attention, softmax attention, and EnLA from left to right. Top: row-normalized absolute magnitudes averaged over heads, with a shared color scale within each block. Bottom: the selected query's spatial magnitude map overlaid on the input image.}
\label{fig:supp_attention_blocks_4}
\end{figure*}
\begin{figure*}[p]
\centering
\begin{minipage}[t]{.49\textwidth}
\centering
{\small\textbf{Block 25}}\par\vspace{2pt}
\includegraphics[width=\linewidth]{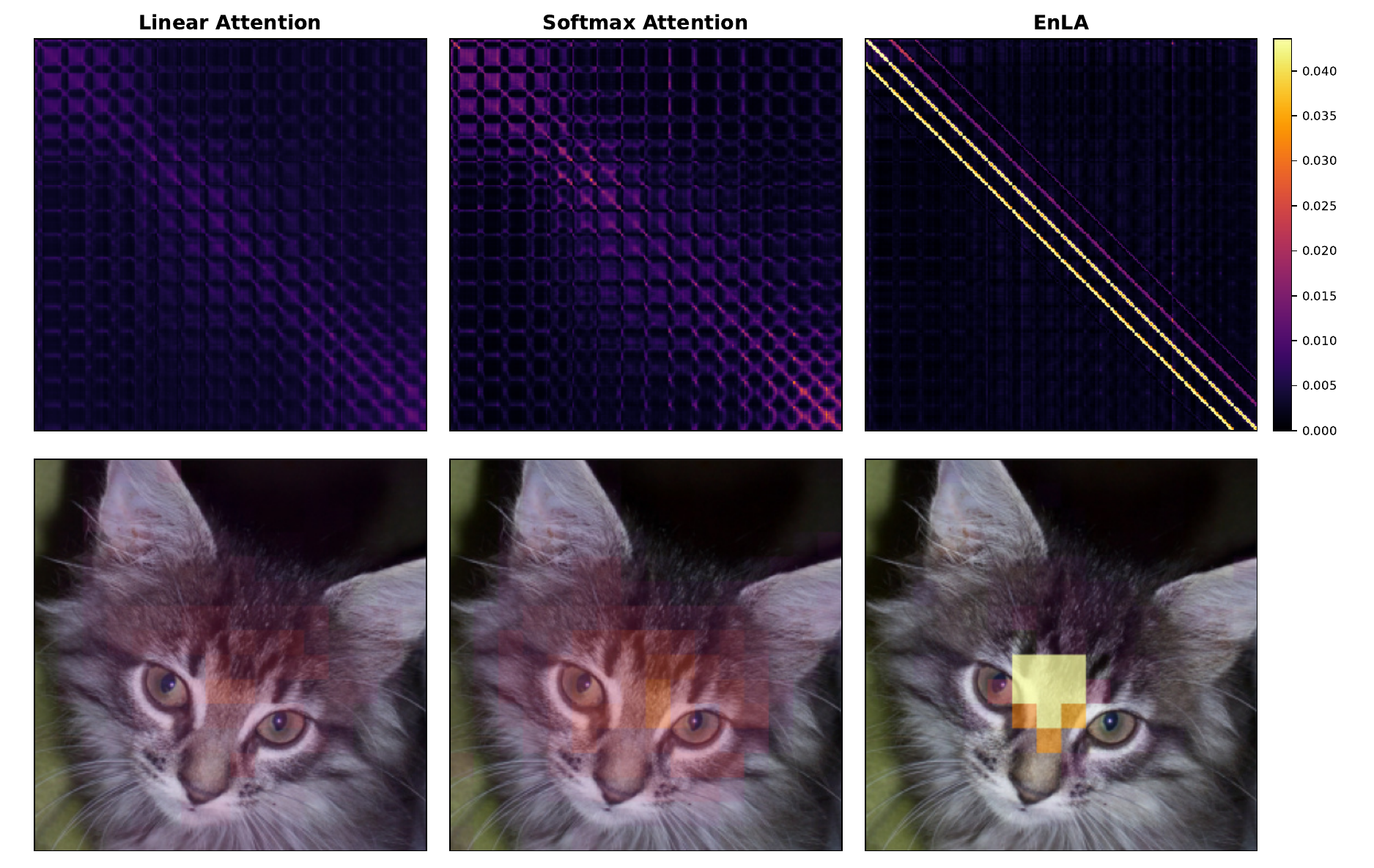}
\end{minipage}
\hfill
\begin{minipage}[t]{.49\textwidth}
\centering
{\small\textbf{Block 26}}\par\vspace{2pt}
\includegraphics[width=\linewidth]{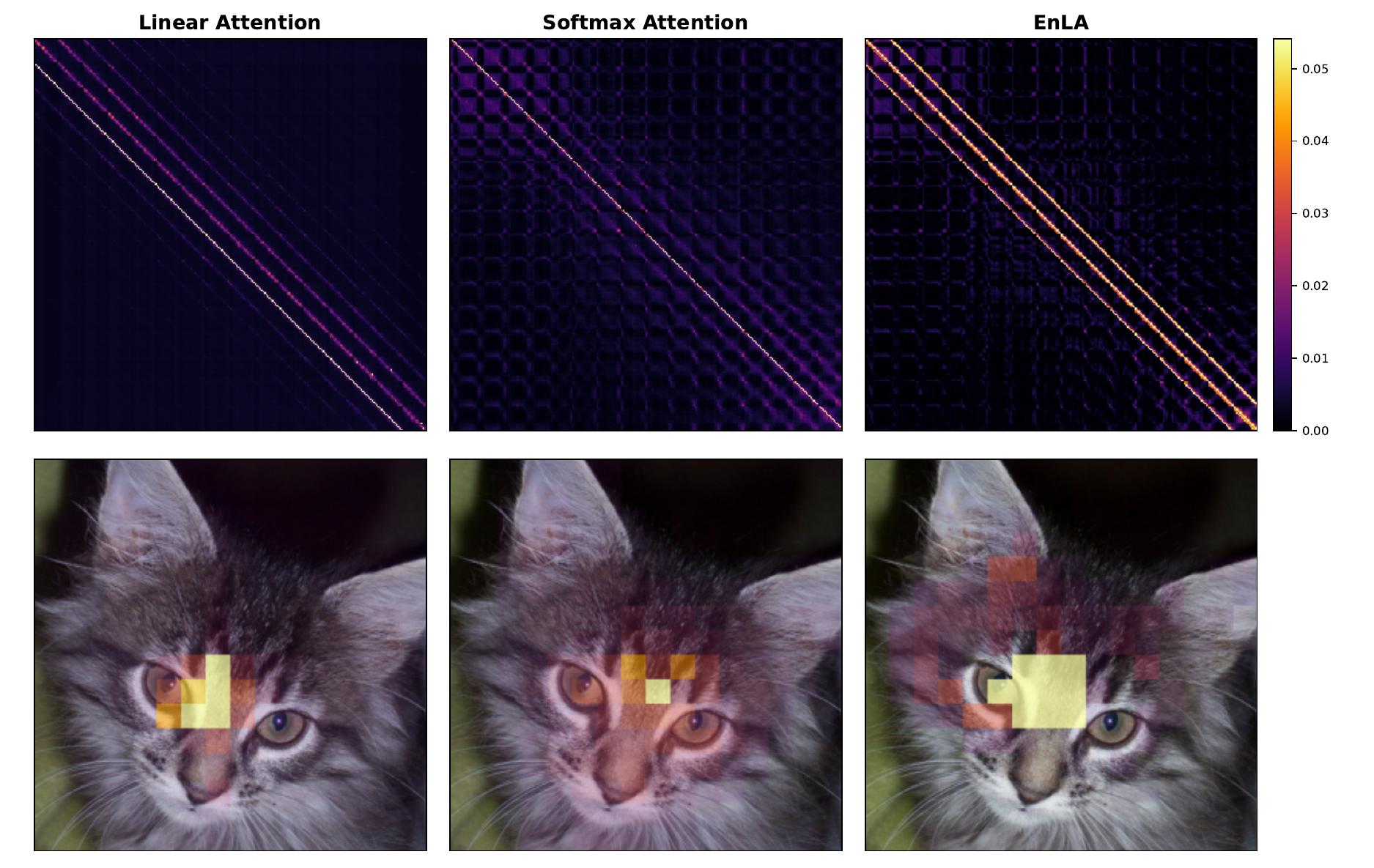}
\end{minipage}
\par\vspace{6pt}
\begin{minipage}[t]{.49\textwidth}
\centering
{\small\textbf{Block 27}}\par\vspace{2pt}
\includegraphics[width=\linewidth]{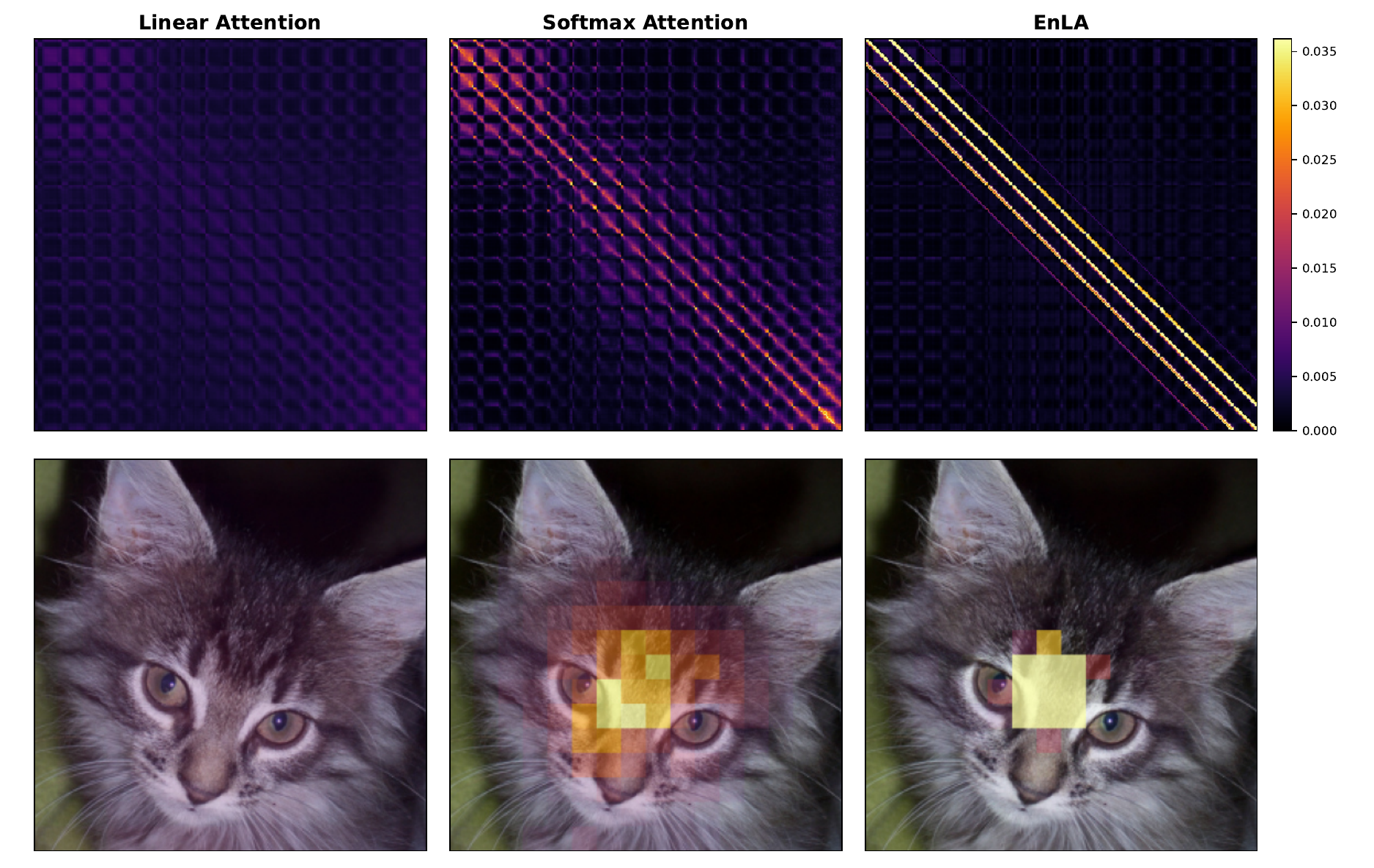}
\end{minipage}
\hfill
\begin{minipage}[t]{.49\textwidth}
\centering
{\small\textbf{Block 28}}\par\vspace{2pt}
\includegraphics[width=\linewidth]{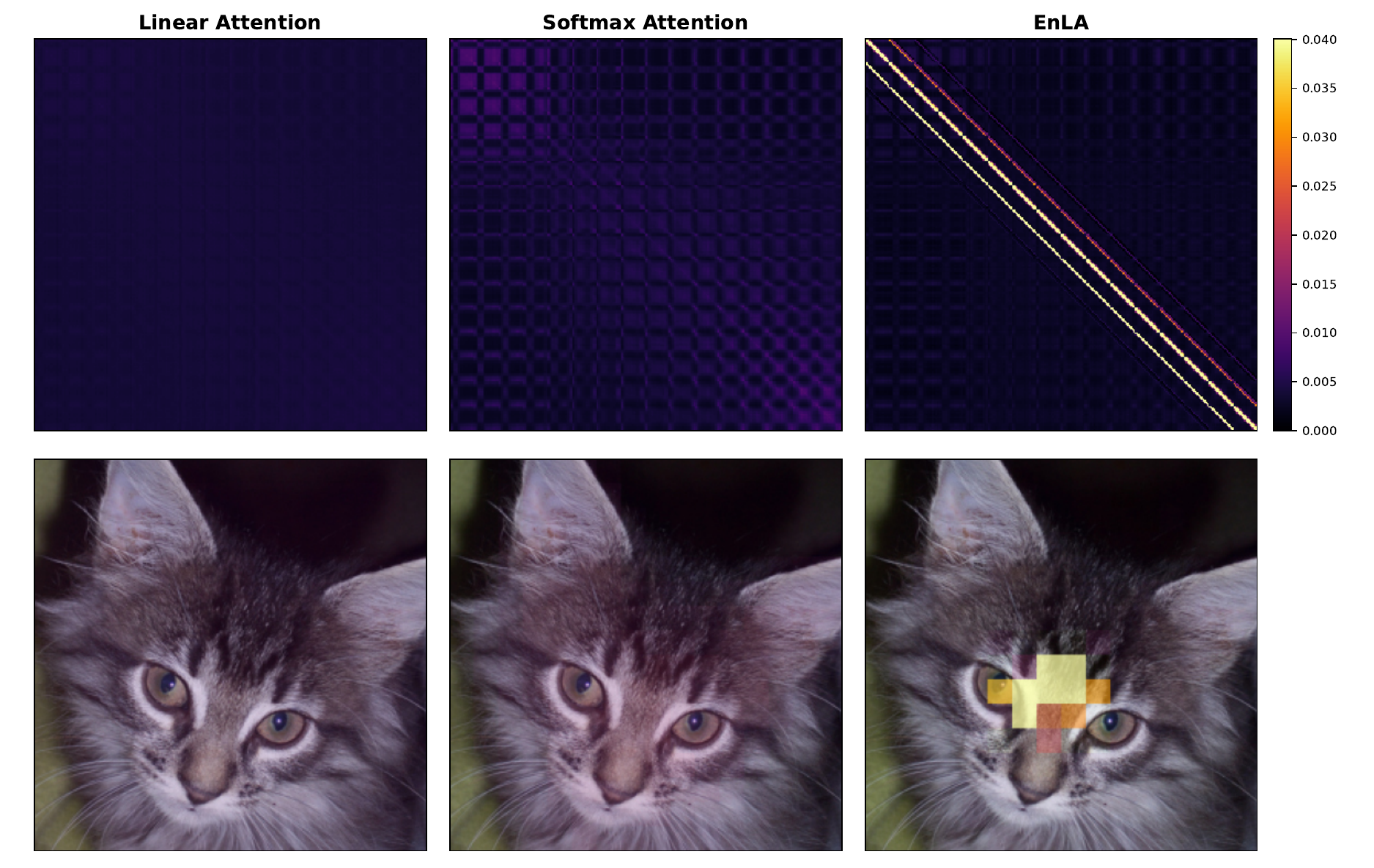}
\end{minipage}
\par\vspace{6pt}
\caption{Block-wise extension of the left part of the main paper's Figure~1: effective token-mixing operators in DiT-XL/2 at $256\times256$ resolution and $t=500$, blocks 25--28. Each panel compares linear attention, softmax attention, and EnLA from left to right. Top: row-normalized absolute magnitudes averaged over heads, with a shared color scale within each block. Bottom: the selected query's spatial magnitude map overlaid on the input image.}
\label{fig:supp_attention_blocks_5}
\end{figure*}

\begin{figure*}[p]
\centering
\captionsetup{skip=2pt}
\includegraphics[width=\textwidth]{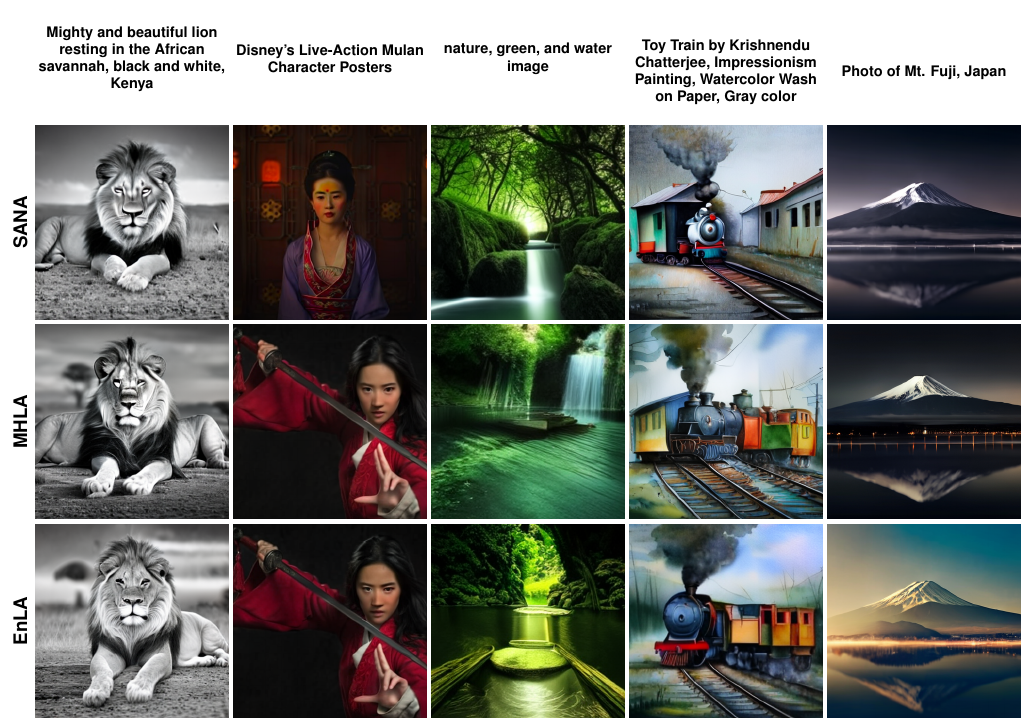}
\caption{Additional text-to-image samples. Rows show SANA, MHLA, and EnLA from top to bottom, with the prompt above each column. Sampling seeds, guidance scales, and sampling steps are matched across models for each prompt.}
\label{fig:supp_t2i_qualitative_1}
\label{fig:supp_t2i_qualitative_2}
\end{figure*}
\begin{figure*}[p]
\centering
\captionsetup{skip=2pt}
\includegraphics[width=\textwidth]{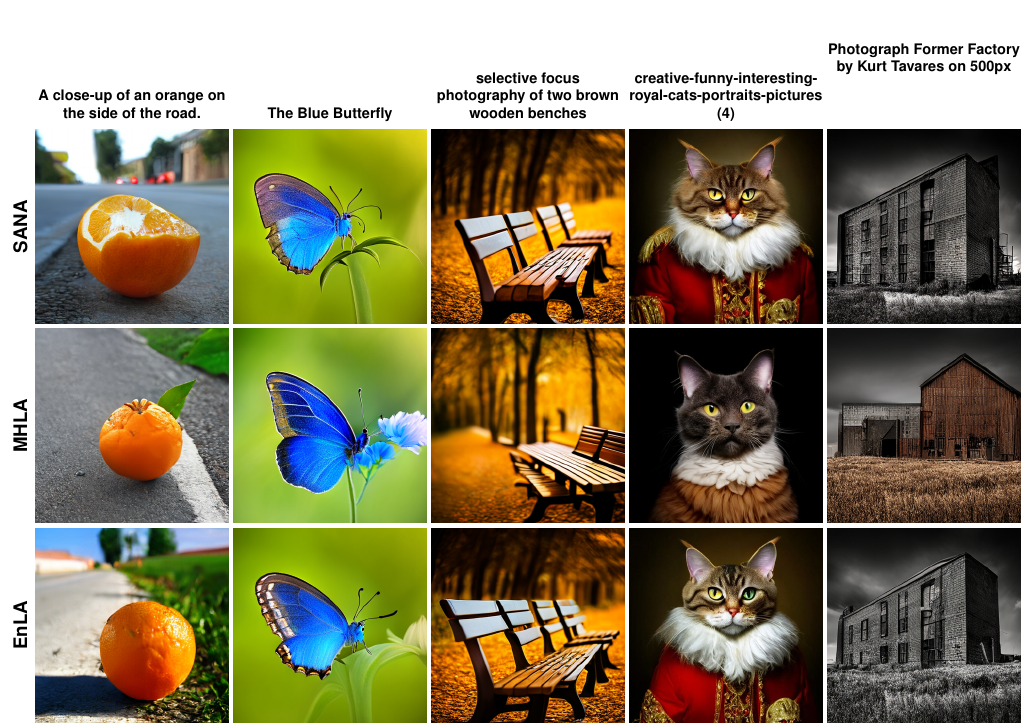}
\caption{Additional text-to-image samples. Rows show SANA, MHLA, and EnLA from top to bottom, with the prompt above each column. Sampling seeds, guidance scales, and sampling steps are matched across models for each prompt.}
\label{fig:supp_t2i_qualitative_3}
\label{fig:supp_t2i_qualitative_4}
\label{fig:supp_t2i_qualitative_5}
\label{fig:supp_t2i_qualitative_6}
\end{figure*}

\begin{table*}[t]
\centering
\caption{EnLA results on ImageNet-1K class-conditional generation without classifier-free guidance.}
\vspace{-5pt}
\label{tab:supp-c2i-nocfg}
\footnotesize
\setlength{\tabcolsep}{5pt}
\renewcommand{\arraystretch}{1.05}
\begin{tabular*}{\textwidth}{@{\extracolsep{\fill}}lllccccc@{}}
\toprule
Scale & Resolution & Method & IS $\uparrow$ & FID $\downarrow$ & sFID $\downarrow$ & Prec. $\uparrow$ & Rec. $\uparrow$ \\
\midrule
\multirow{5}{*}{DiT-S/2} & \multirow{3}{*}{$256{\times}256$} & EnLA & 22.93 & 61.15 & 11.43 & 0.396 & 0.582 \\
 &  & EnLA + SMN ($\phi(q)$) & 23.52 & 60.49 & 11.65 & 0.400 & \textbf{0.588} \\
 &  & EnLA + Gating & \textbf{24.76} & \textbf{58.43} & \textbf{11.36} & \textbf{0.411} & \textbf{0.588} \\
\cmidrule{2-8}
 & \multirow{2}{*}{$512{\times}512$} & EnLA & 19.57 & 75.90 & 14.40 & 0.406 & 0.513 \\
 &  & EnLA + Gating & \textbf{20.84} & \textbf{73.26} & \textbf{14.27} & \textbf{0.430} & \textbf{0.530} \\
\midrule
\multirow{2}{*}{DiT-B/2} & \multirow{2}{*}{$256{\times}256$} & EnLA & 38.38 & 38.36 & \textbf{8.06} & 0.506 & \textbf{0.630} \\
 &  & EnLA + Gating & \textbf{41.16} & \textbf{35.89} & \textbf{8.06} & \textbf{0.524} & 0.627 \\
\midrule
\multirow{2}{*}{DiT-L/2} & \multirow{2}{*}{$256{\times}256$} & EnLA & 63.04 & 21.86 & \textbf{6.09} & 0.608 & \textbf{0.633} \\
 &  & EnLA + Gating & \textbf{65.46} & \textbf{20.92} & 6.40 & \textbf{0.621} & 0.627 \\
\midrule
\multirow{2}{*}{DiT-XL/2} & \multirow{2}{*}{$256{\times}256$} & EnLA & 71.47 & 18.40 & \textbf{5.85} & 0.637 & \textbf{0.634} \\
 &  & EnLA + Gating & \textbf{73.72} & \textbf{17.77} & 6.29 & \textbf{0.643} & 0.632 \\
\bottomrule
\end{tabular*}
\vspace{-5pt}
\end{table*}

\begin{table*}[t]
\centering
\caption{ImageNet-1K class-conditional generation \emph{with classifier-free guidance}
(CFG $=1.5$).}
\label{tab:supp-c2i-cfg}
\footnotesize
\setlength{\tabcolsep}{5pt}
\renewcommand{\arraystretch}{1.05}
\begin{tabular*}{\textwidth}{@{\extracolsep{\fill}}lllccccc@{}}
\toprule
Scale & Resolution & Method & IS $\uparrow$ & FID $\downarrow$ & sFID $\downarrow$ & Prec. $\uparrow$ & Rec. $\uparrow$ \\
\midrule
\multirow{5}{*}{DiT-S/2} & \multirow{3}{*}{$256{\times}256$} & EnLA & 41.14 & 37.71 & 8.56 & 0.520 & 0.547 \\
 &  & EnLA + SMN ($\phi(q)$) & 41.87 & 37.11 & 8.67 & 0.524 & \textbf{0.551} \\
 &  & EnLA + Gating & \textbf{46.94} & \textbf{34.12} & \textbf{8.43} & \textbf{0.547} & 0.547 \\
\cmidrule{2-8}
 & \multirow{2}{*}{$512{\times}512$} & EnLA & 30.43 & 52.17 & 11.50 & 0.544 & \textbf{0.537} \\
 &  & EnLA + Gating & \textbf{34.22} & \textbf{48.64} & \textbf{11.08} & \textbf{0.566} & 0.523 \\
\midrule
\multirow{2}{*}{DiT-B/2} & \multirow{2}{*}{$256{\times}256$} & EnLA & 84.63 & 16.71 & 6.01 & 0.677 & 0.541 \\
 &  & EnLA + Gating & \textbf{91.28} & \textbf{14.89} & \textbf{5.98} & \textbf{0.696} & \textbf{0.548} \\
\midrule
\multirow{2}{*}{DiT-L/2} & \multirow{2}{*}{$256{\times}256$} & EnLA & 141.75 & 6.98 & \textbf{4.81} & 0.781 & \textbf{0.524} \\
 &  & EnLA + Gating & \textbf{150.09} & \textbf{6.23} & 4.89 & \textbf{0.790} & \textbf{0.524} \\
\midrule
\multirow{2}{*}{DiT-XL/2} & \multirow{2}{*}{$256{\times}256$} & EnLA & 163.56 & 5.41 & 4.69 & 0.801 & \textbf{0.524} \\
 &  & EnLA + Gating & \textbf{168.66} & \textbf{4.98} & \textbf{4.67} & \textbf{0.810} & 0.523 \\
\bottomrule
\end{tabular*}
\end{table*}

\begin{table*}[t]
\centering
\caption{Parameter counts and GFLOPs for DiT variants across model scales and image resolutions.}
\label{tab:supp_params_flops}
\vspace{-5pt}
\footnotesize
\setlength{\tabcolsep}{5pt}
\renewcommand{\arraystretch}{1.05}
\begin{tabular*}{\textwidth}{@{\extracolsep{\fill}}lllcc@{}}
\toprule
Scale & Resolution & Method & Params (M) & GFLOPs \\
\midrule
\multirow{6}{*}{DiT-S/2} & \multirow{3}{*}{$256{\times}256$} & Softmax & 32.96 & 6.07 \\
 && Linear & 32.96 & 5.61 \\
 && EnLA & 33.90 & 5.87 \\
\cmidrule{2-5}
 & \multirow{3}{*}{$512{\times}512$} & Softmax & 33.26 & 31.48 \\
 && Linear & 33.26 & 22.42 \\
 && EnLA & 34.20 & 23.45 \\
\midrule
\multirow{3}{*}{DiT-B/2} & \multirow{3}{*}{$256{\times}256$} & Softmax & 130.51 & 23.02 \\
 && Linear & 130.51 & 22.12 \\
 && EnLA & 133.26 & 22.86 \\
\midrule
\multirow{3}{*}{DiT-L/2} & \multirow{3}{*}{$256{\times}256$} & Softmax & 458.10 & 80.75 \\
 && Linear & 458.10 & 78.34 \\
 && EnLA & 466.99 & 80.72 \\
\midrule
\multirow{3}{*}{DiT-XL/2} & \multirow{3}{*}{$256{\times}256$} & Softmax & 675.13 & 118.69 \\
 && Linear & 675.13 & 115.66 \\
 && EnLA & 687.83 & 119.05 \\
\bottomrule
\end{tabular*}
\vspace{-5pt}
\end{table*}

\begin{figure*}[t]
\centering
\includegraphics[width=\textwidth]{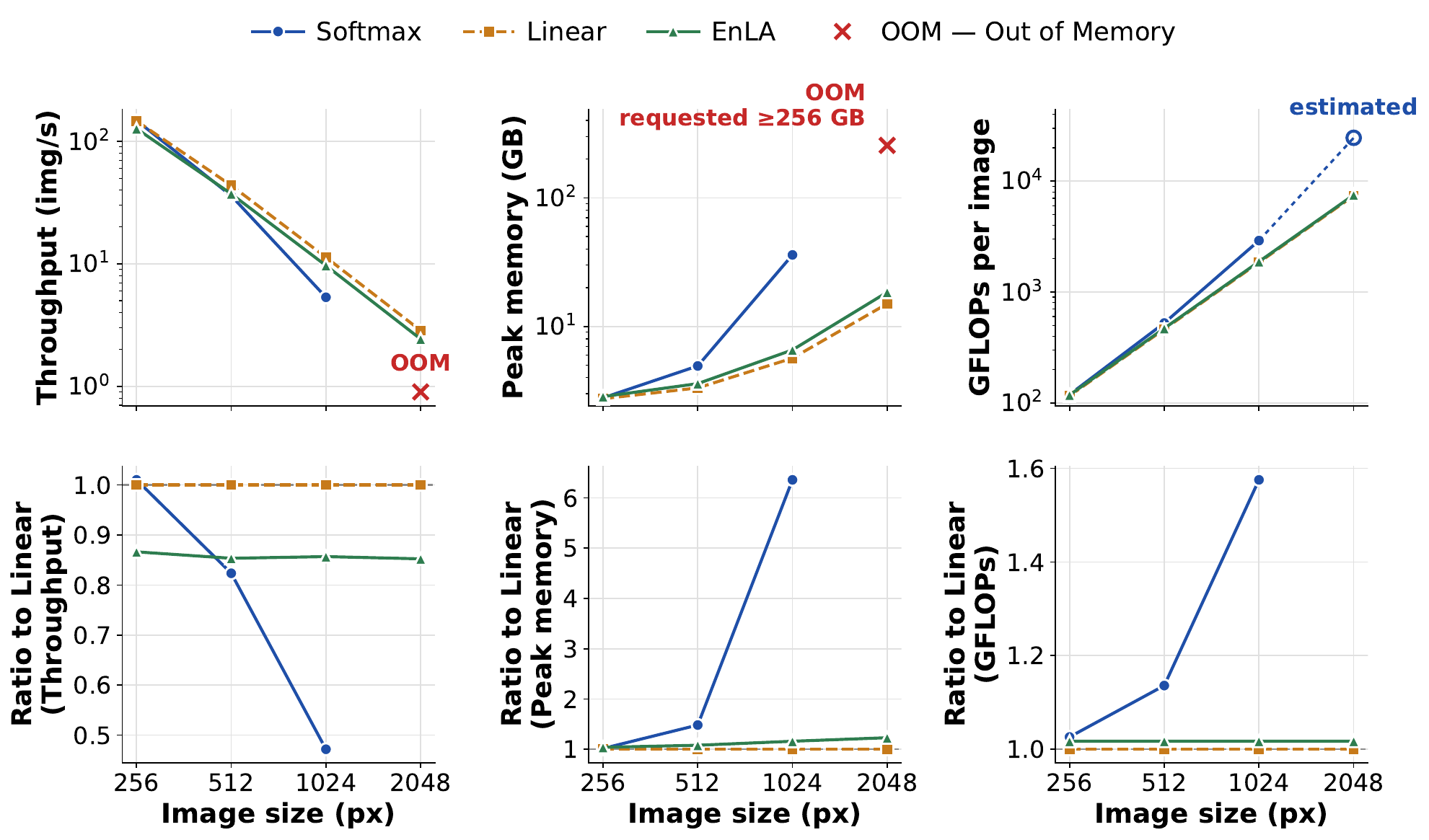}
\caption{Inference cost of DiT-XL/2 at batch size 16 on one NVIDIA H200 in full fp32. Top: throughput, peak memory, and GFLOPs per image. Bottom: corresponding ratios to vanilla linear attention. OOM indicates an out-of-memory failure; the softmax GFLOPs at $2048\times2048$ are estimated.}
\label{fig:supp_infer_full}
\end{figure*}

\begin{figure*}[!t]
    \centering
    \captionsetup{skip=2pt}
    \includegraphics[width=\textwidth,trim=0 3pt 0 0,clip]{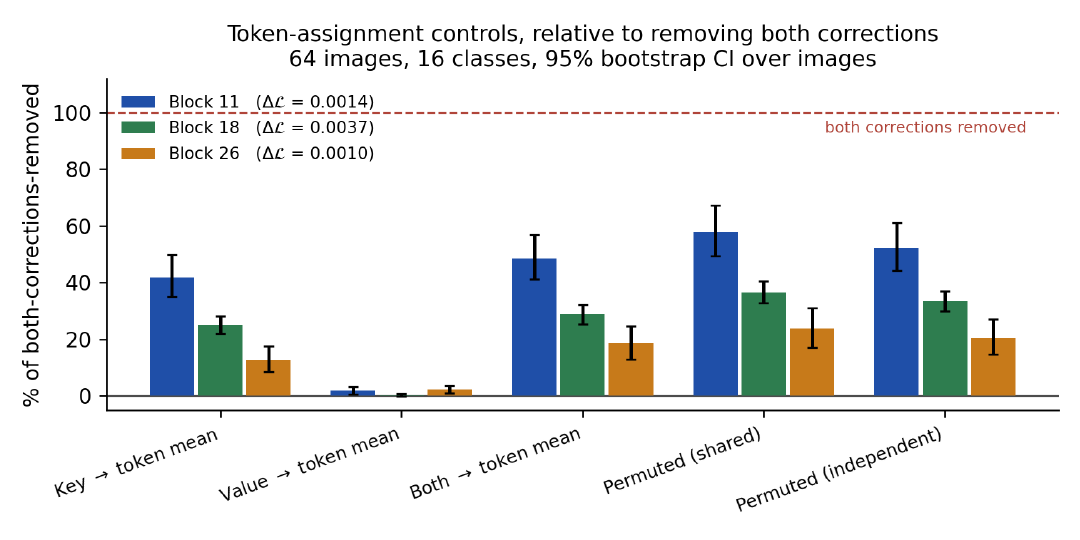}
    \caption{Effect of token-mean replacement and permutation at blocks 11, 18, and 26.
    Error increases are normalized by removing both corrections at each block (dashed line).
    Error bars show paired-bootstrap 95\% confidence intervals over images.}
    \label{fig:supp_token_controls}
\end{figure*}

\begin{table*}[t]
\centering
\caption{ImageNet-1K classification comparison with non-DeiT~\cite{touvron2021deit} vision transformer (ViT)~\cite{dosovitskiy2021vit} architectures. $\dagger$ denotes models trained from scratch using the same training recipe as EnLA.}
\label{tab:sota-in1k}
\label{tab:sota-in1k-s}
\small
\setlength{\tabcolsep}{3pt}
\renewcommand{\arraystretch}{1.05}
\begin{tabular*}{\linewidth}{@{\extracolsep{\fill}}clccc@{}}
\toprule
 & Model & Params & FLOPs & Top-1 Acc. $\uparrow$ \\
\midrule
\multirow{6}{*}{\rotatebox{90}{\shortstack{Smaller-scale\\models}}} & FL-PVT-T~\cite{Han_2023_ICCV}       & 12M   & 2.0G & 77.8 \\
 & FL-PVTv2-B1~\cite{Han_2023_ICCV}    & 13M   & 2.2G & 79.5 \\
 & MSVMamba-M~\cite{shi2024multi}     & 12M   & 1.5G & 79.8 \\
 & NAT-M~\cite{hassani2023neighborhood}          & 20M   & 2.7G & 81.8 \\
 & MAViT-T$^{\dagger}$~\cite{fan2025mala}        & 16M   & 2.5G & 82.3 \\
 & \textbf{EnLA}  & 17M & 2.7G & \textbf{82.5} \\
\midrule
\multirow{6}{*}{\rotatebox{90}{\shortstack{Larger-scale\\models}}} & FAT-B3~\cite{fan2023lightweight}         & 29M & 4.4G & 83.6 \\
 & VMamba-T~\cite{liu2024vmamba}       & 30M & 4.9G & 82.6 \\
 & MV-T~\cite{hatamizadeh2025mambavision}           & 32M & 4.4G & 82.3 \\
 & MSVMamba-T~\cite{shi2024multi}     & 32M & 5.1G & 83.0 \\
 & MAViT-S$^{\dagger}$~\cite{fan2025mala}        & 27M & 4.6G & 84.1 \\
 & \textbf{EnLA}  & 28M & 4.8G & \textbf{84.3} \\
\bottomrule
\end{tabular*}
\end{table*}

\clearpage
\begin{figure*}[p]
\centering
\setlength{\tabcolsep}{0pt}
\renewcommand{\arraystretch}{0}
\begin{tabular}{@{}cc@{}}
\includegraphics[width=0.5\textwidth]{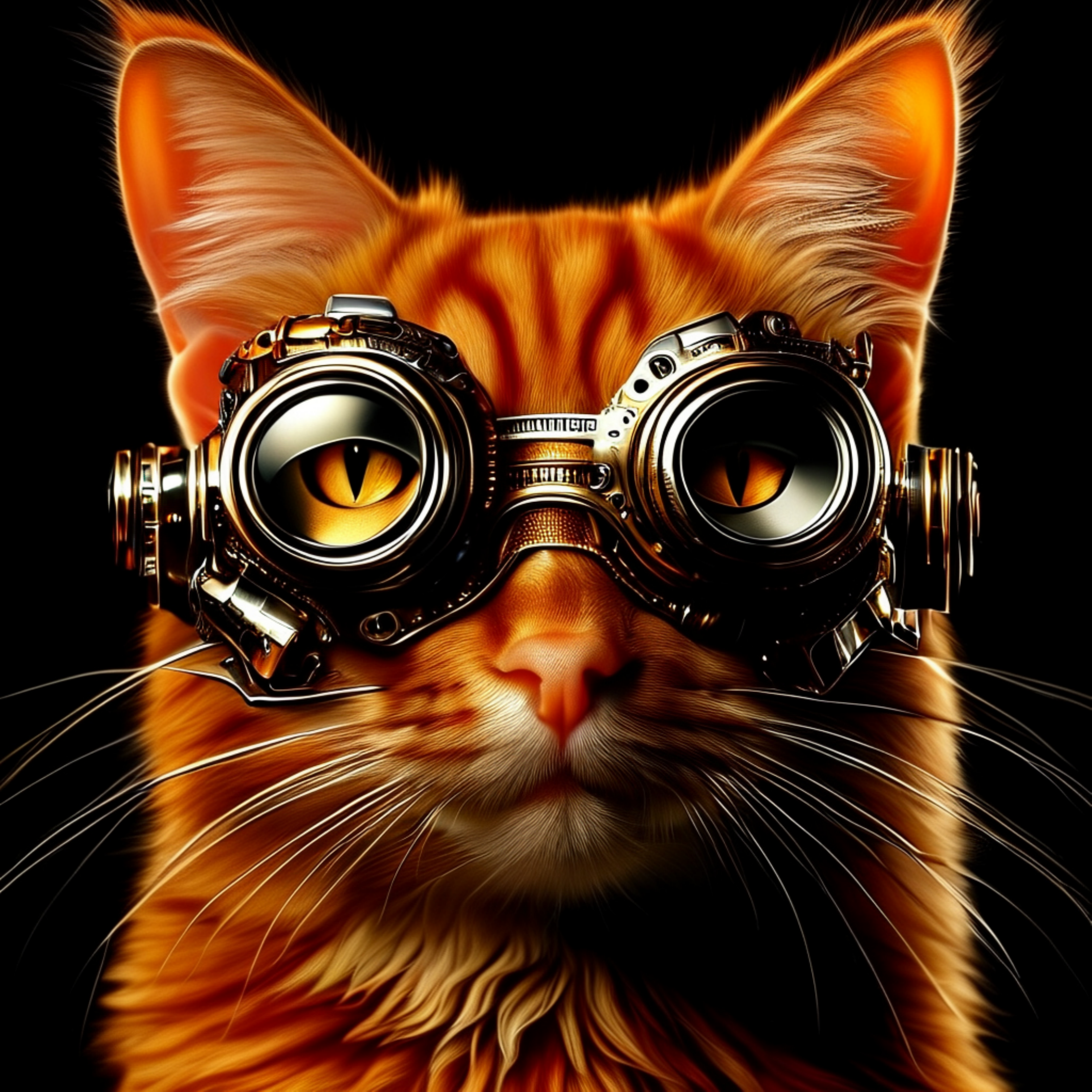}&%
\includegraphics[width=0.5\textwidth]{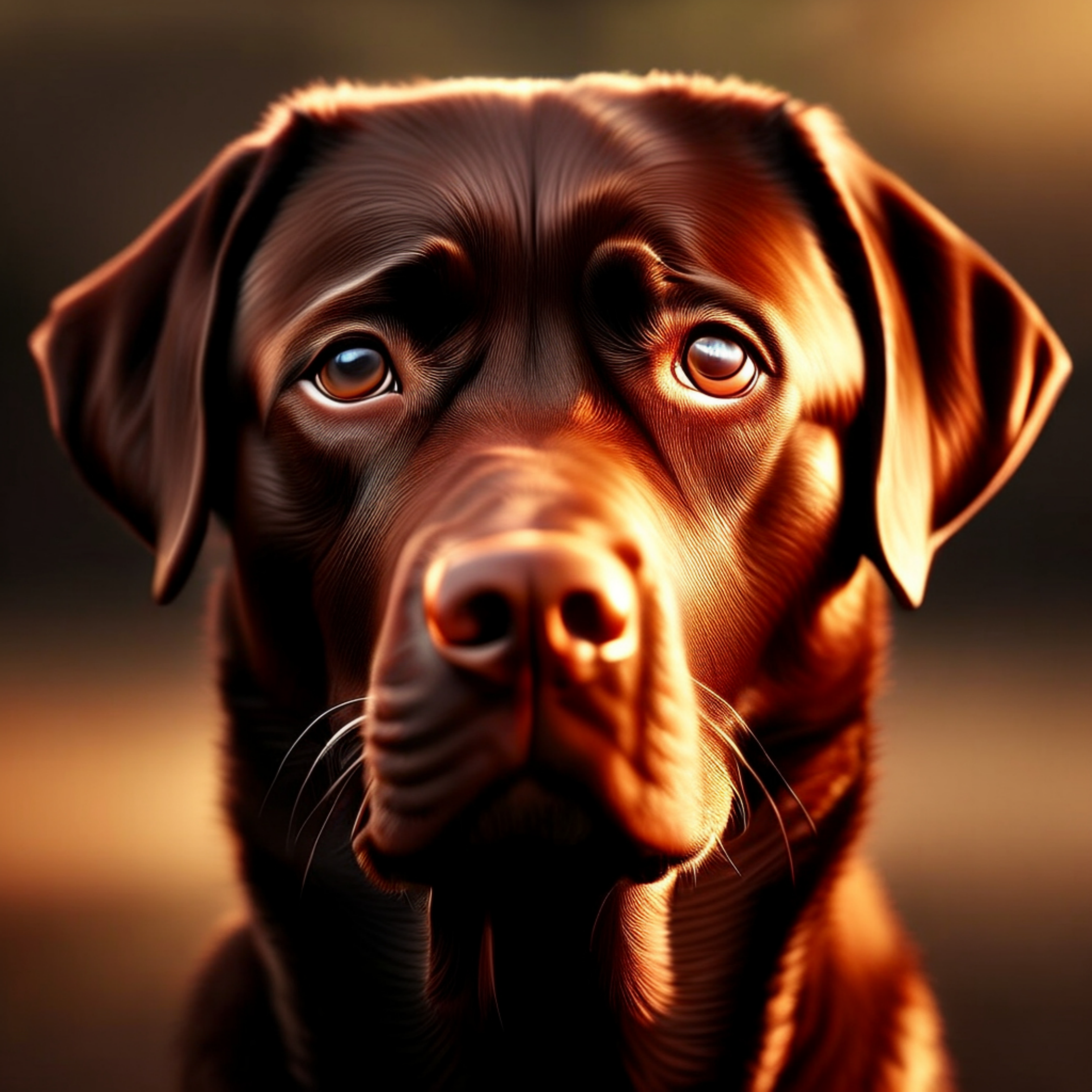}\\[-\lineskip]
\includegraphics[width=0.5\textwidth]{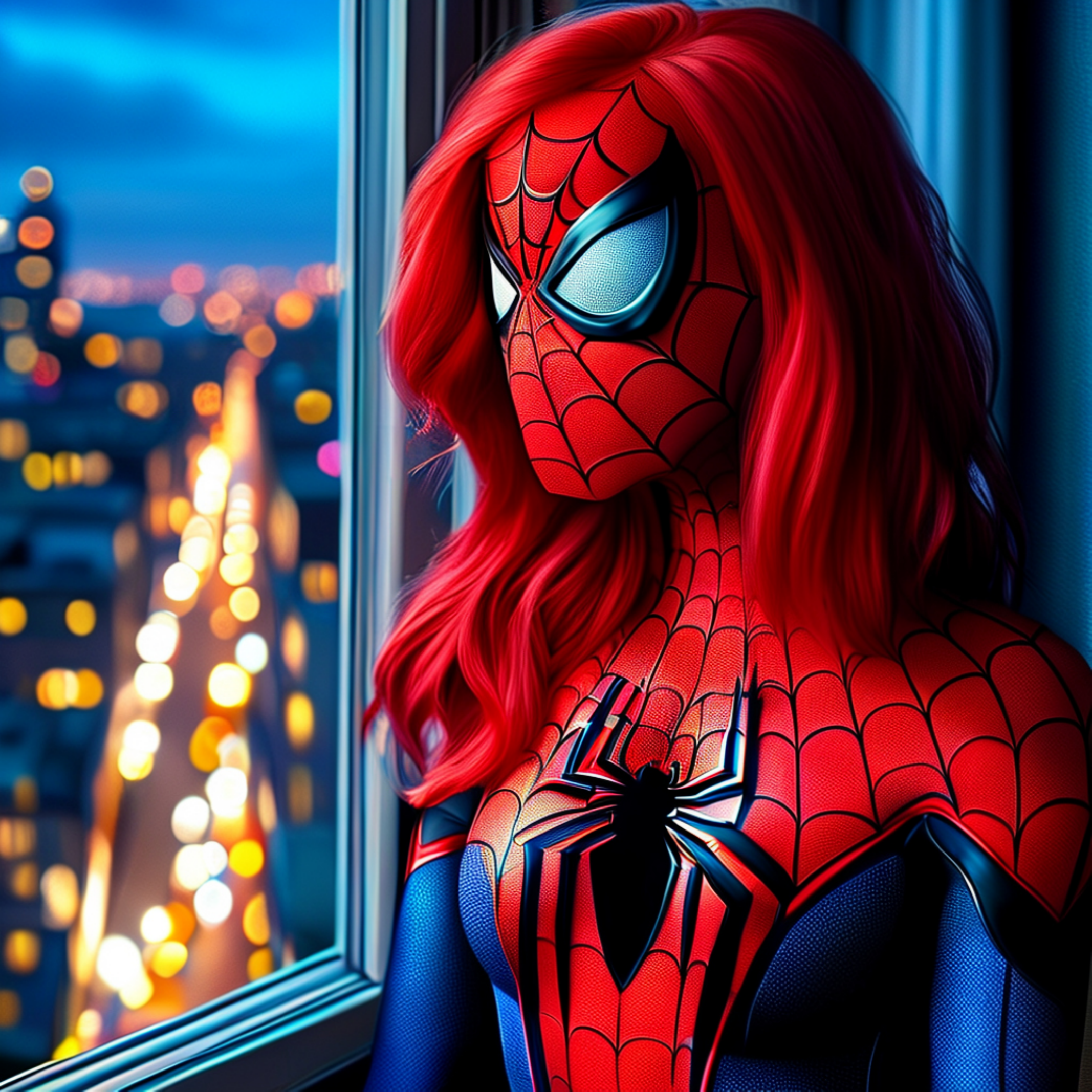}&%
\includegraphics[width=0.5\textwidth]{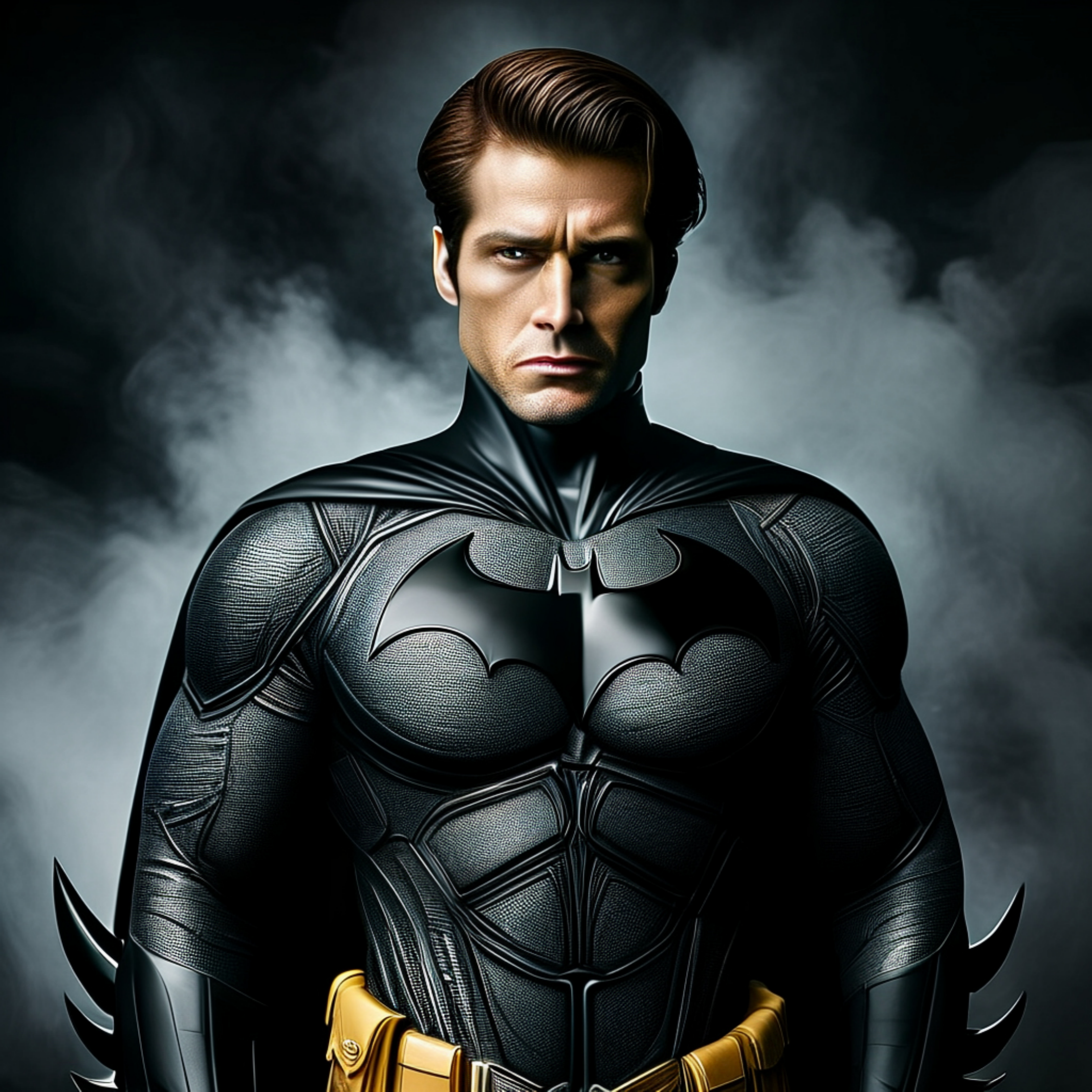}
\end{tabular}
\caption{EnLA text-to-image generation at $1024\times1024$ resolution.}
\label{fig:supp-dalle-grid}
\end{figure*}
\clearpage

\begin{figure*}[p]
\centering
\setlength{\tabcolsep}{0pt}
\renewcommand{\arraystretch}{0}
\begin{tabular}{@{}ccc@{}}
\includegraphics[width=0.333333\textwidth]{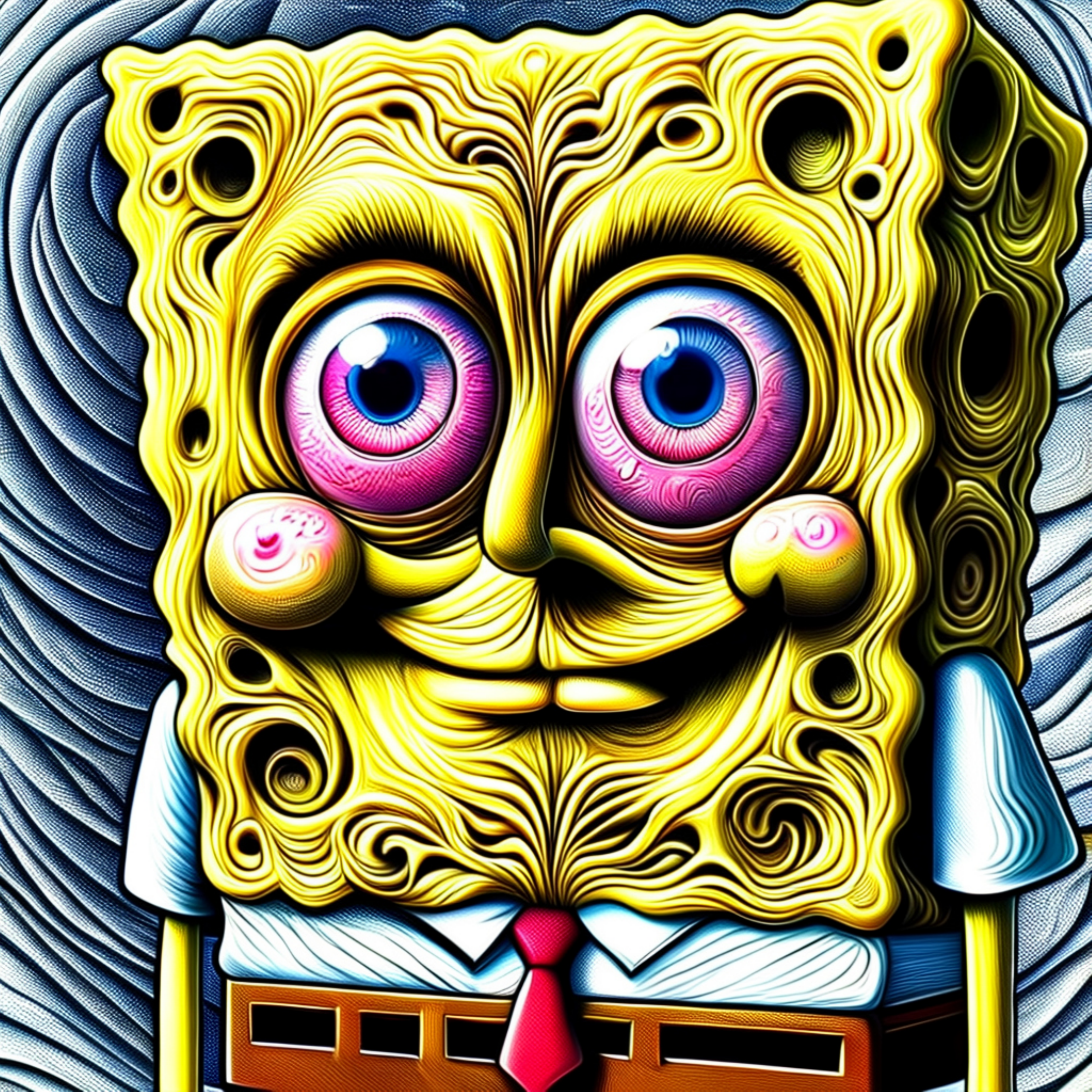}&%
\includegraphics[width=0.333333\textwidth]{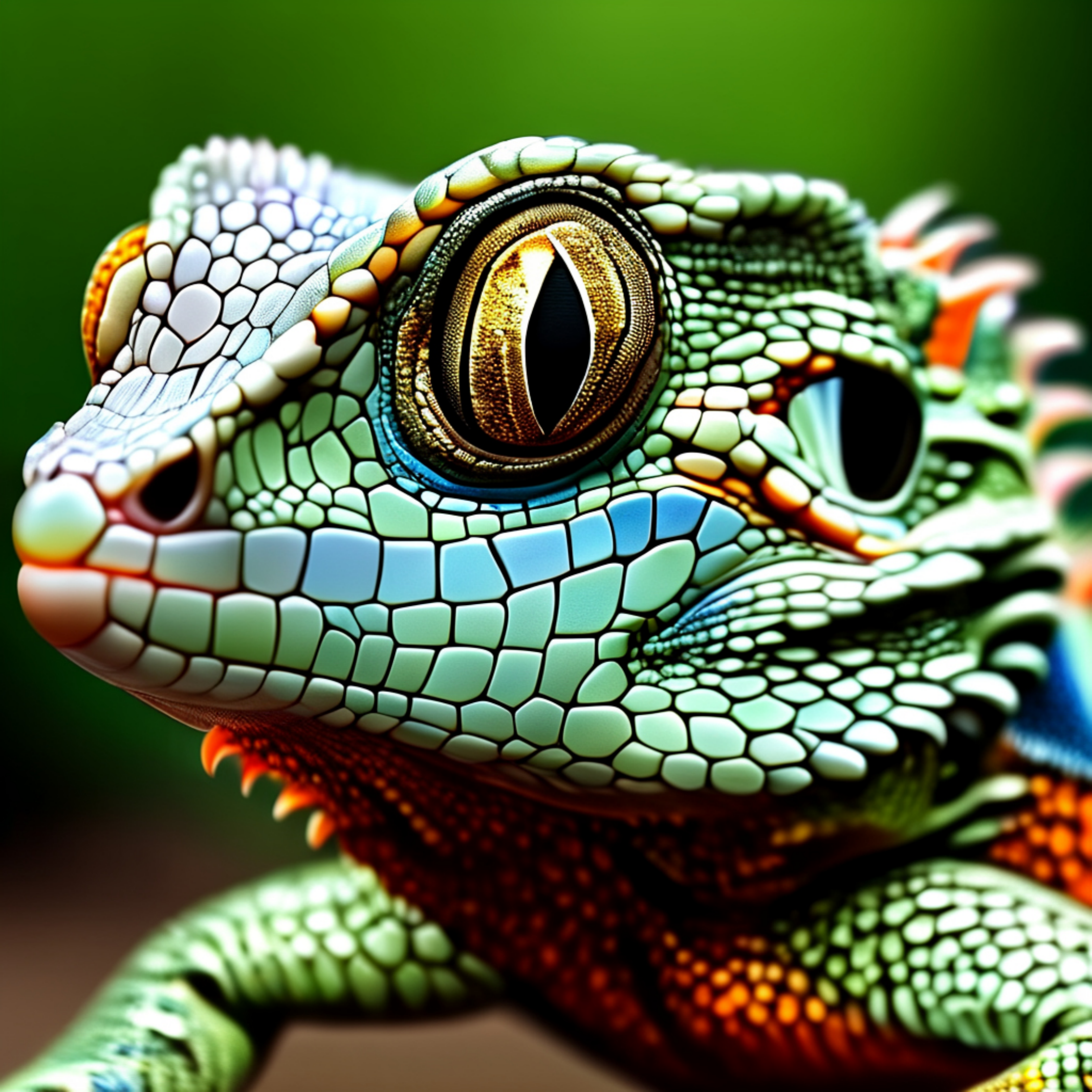}&%
\includegraphics[width=0.333333\textwidth]{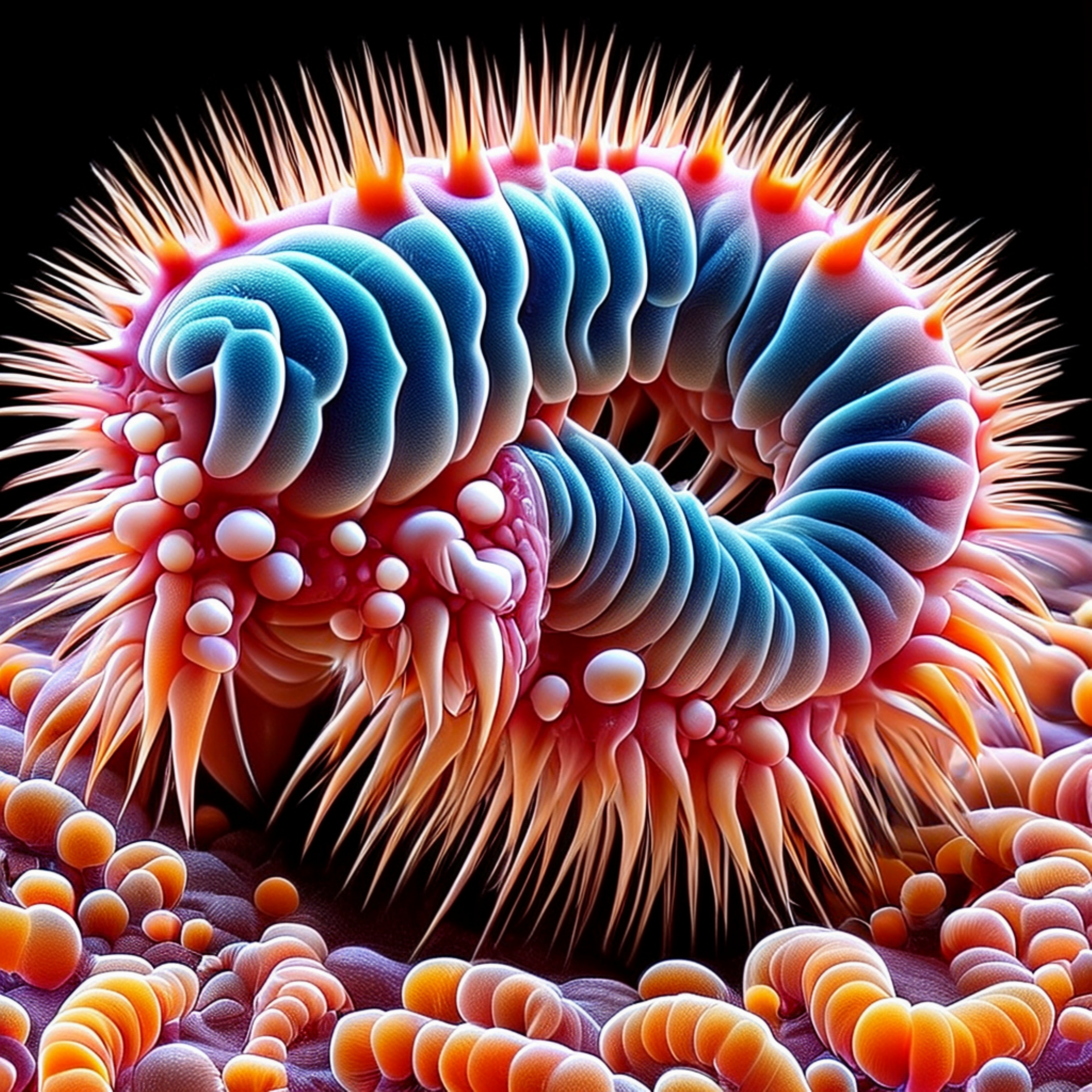}\\[-\lineskip]
\includegraphics[width=0.333333\textwidth]{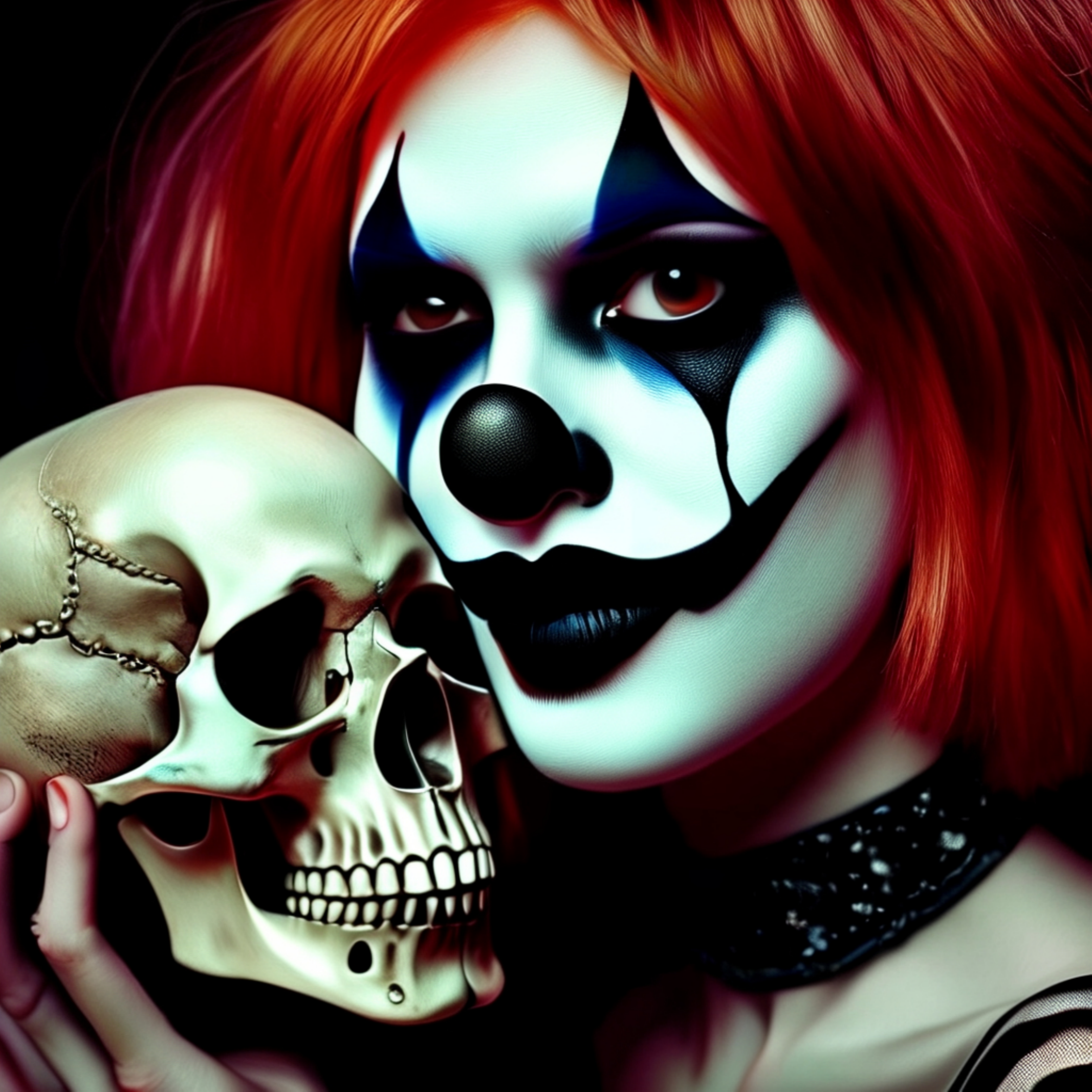}&%
\includegraphics[width=0.333333\textwidth]{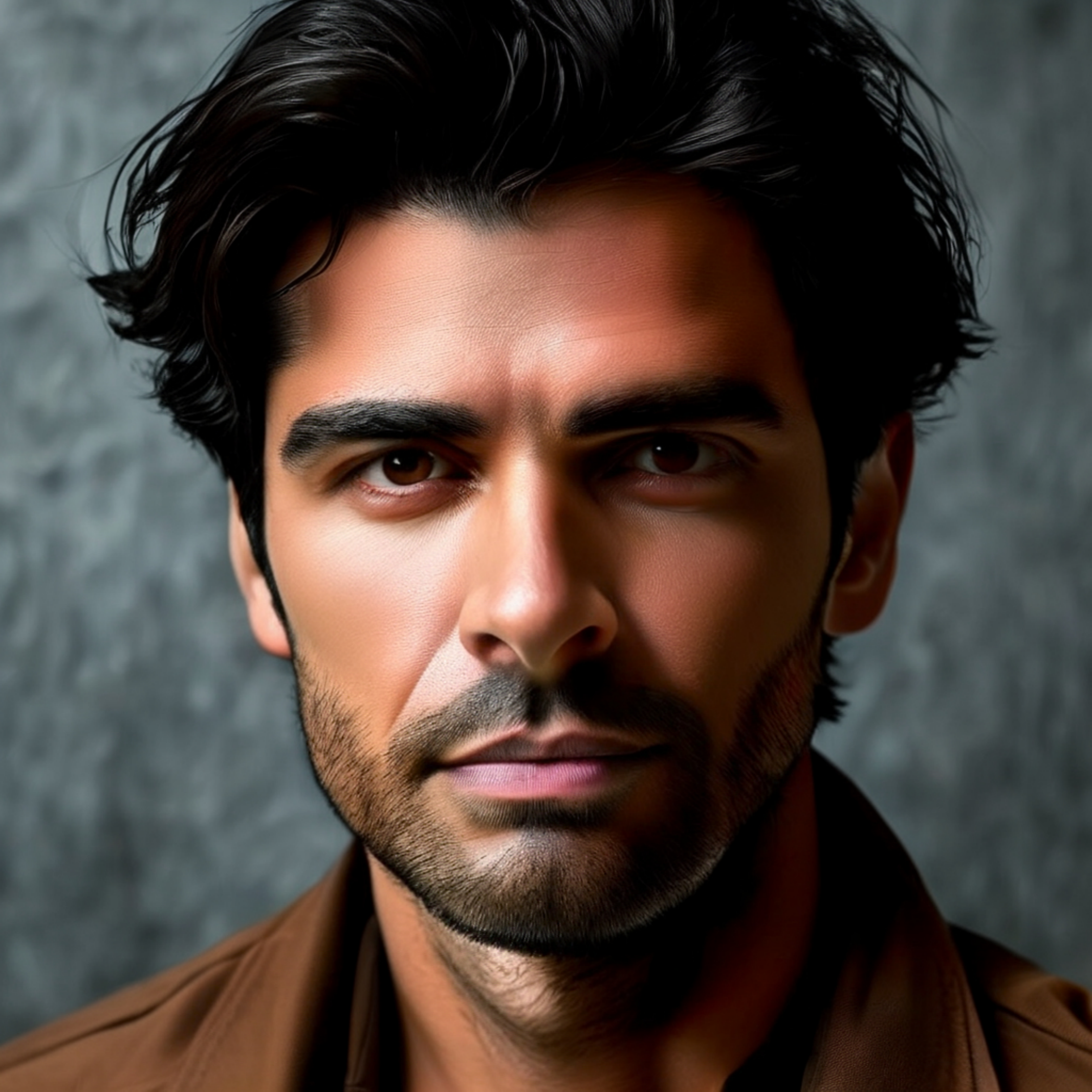}&%
\includegraphics[width=0.333333\textwidth]{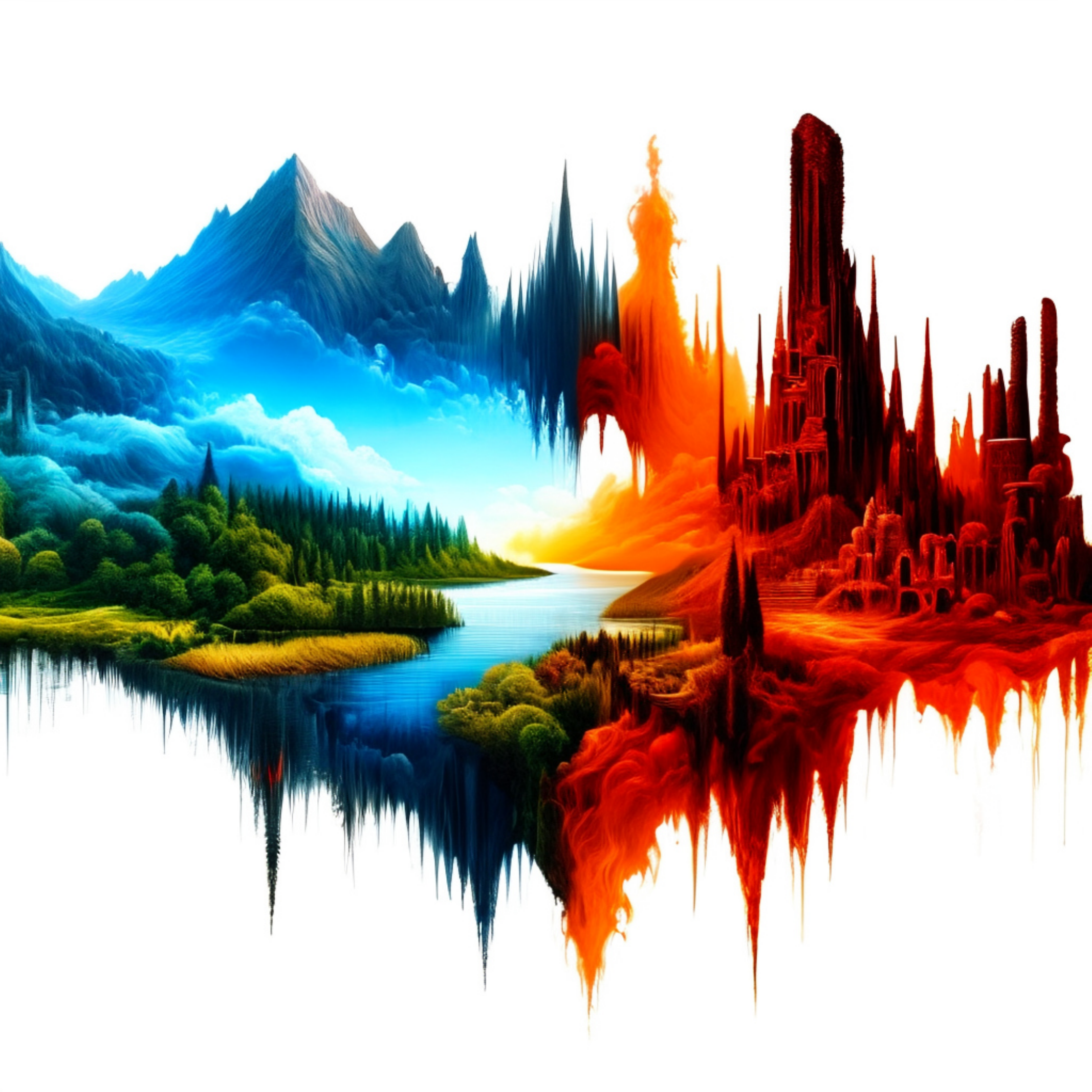}
\end{tabular}
\caption{EnLA text-to-image generation at $1024\times1024$ resolution.}
\label{fig:supp-dalle-remaining}
\end{figure*}
\clearpage

\begin{figure*}[p]
\centering
\setlength{\tabcolsep}{0pt}
\renewcommand{\arraystretch}{0}
\begin{tabular}{@{}ccc@{}}
\includegraphics[width=0.333333\textwidth]{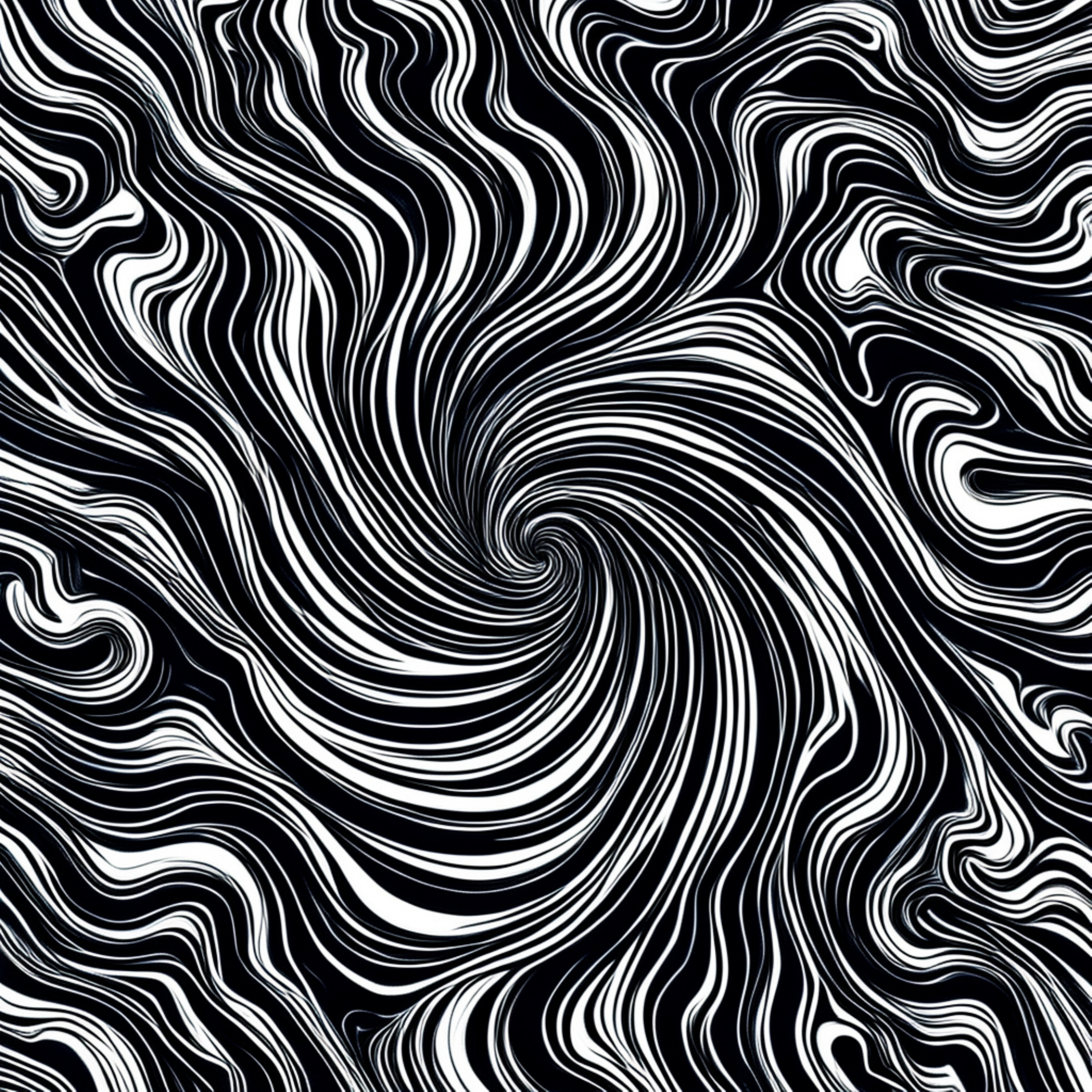}&%
\includegraphics[width=0.333333\textwidth]{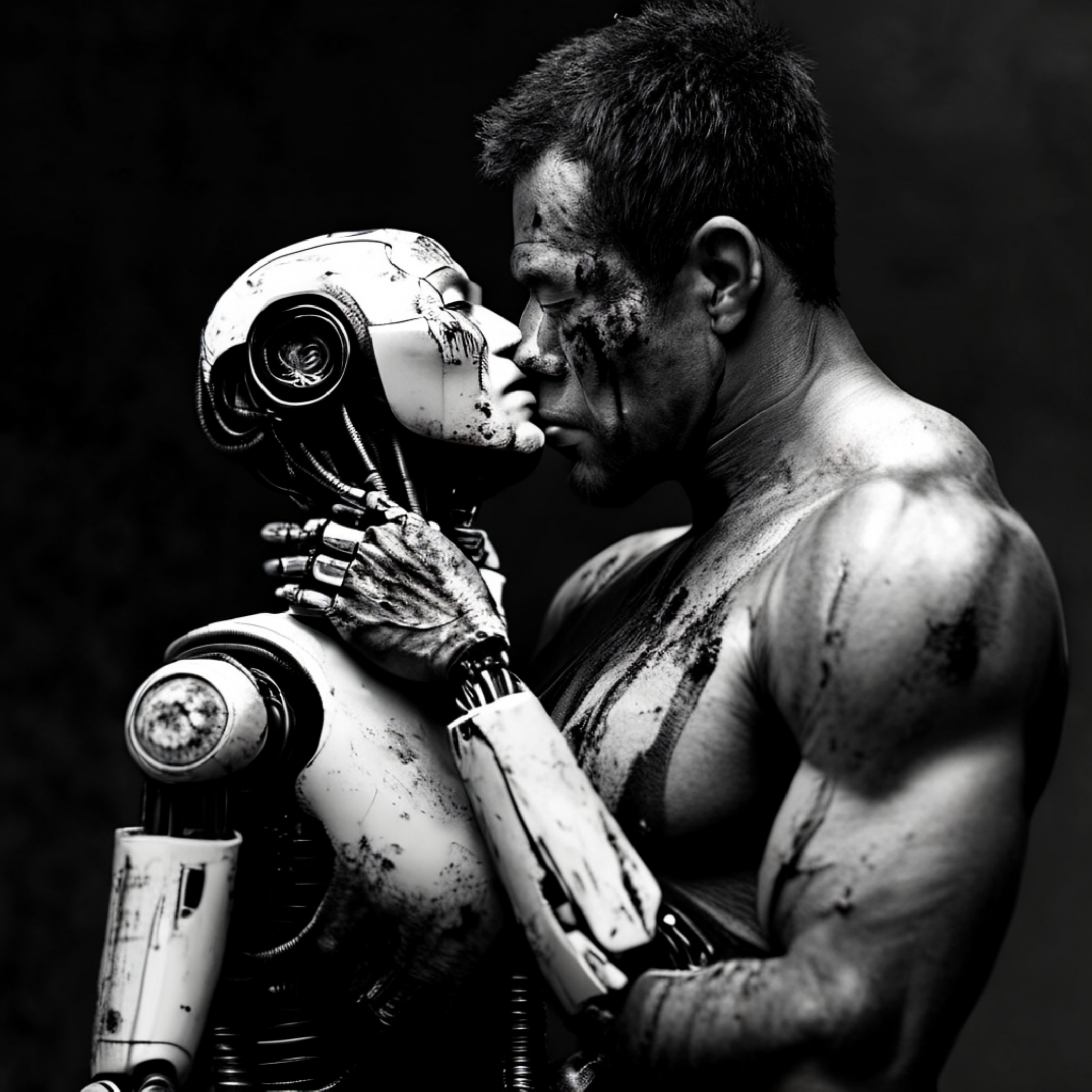}&%
\includegraphics[width=0.333333\textwidth]{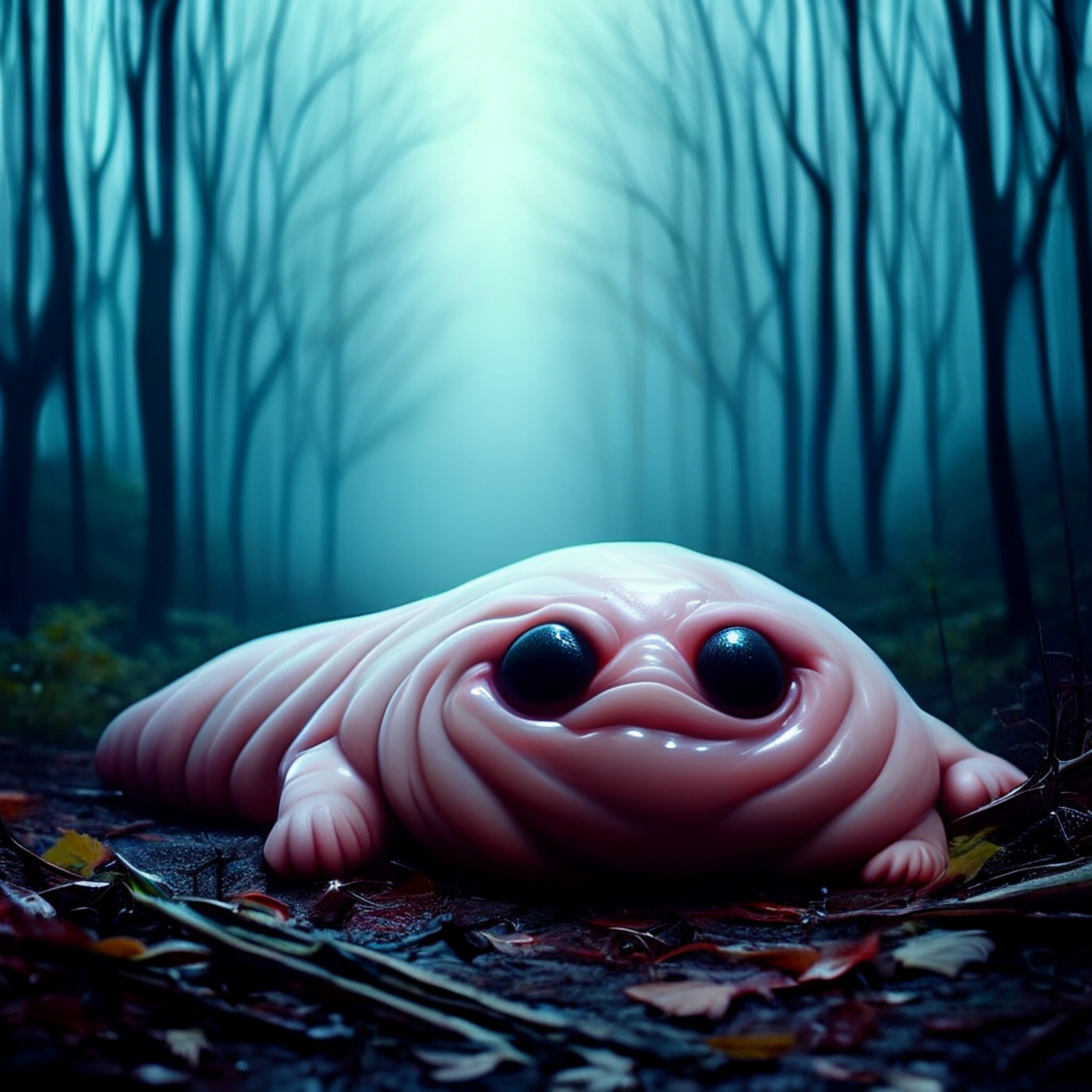}\\[-\lineskip]
\includegraphics[width=0.333333\textwidth]{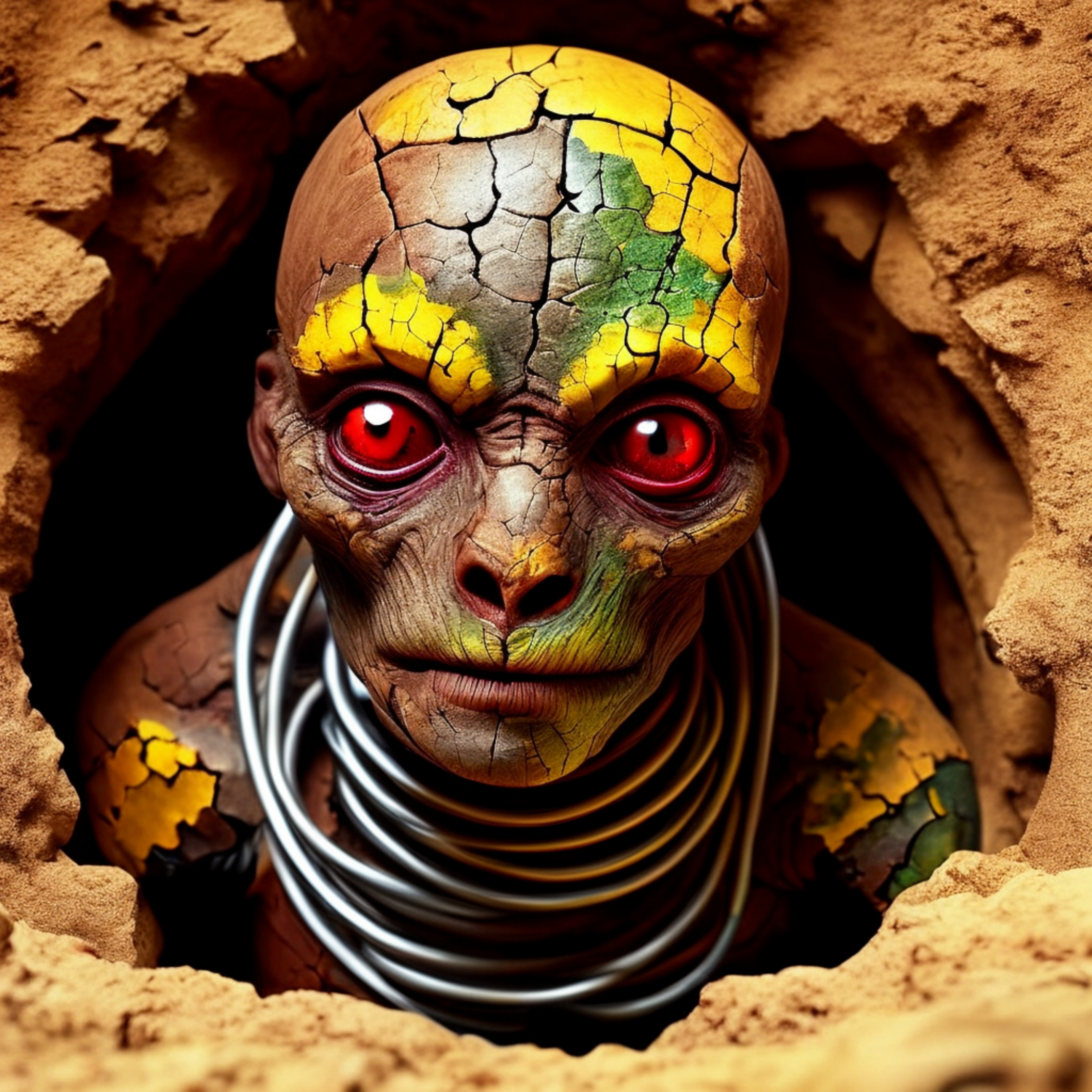}&%
\includegraphics[width=0.333333\textwidth]{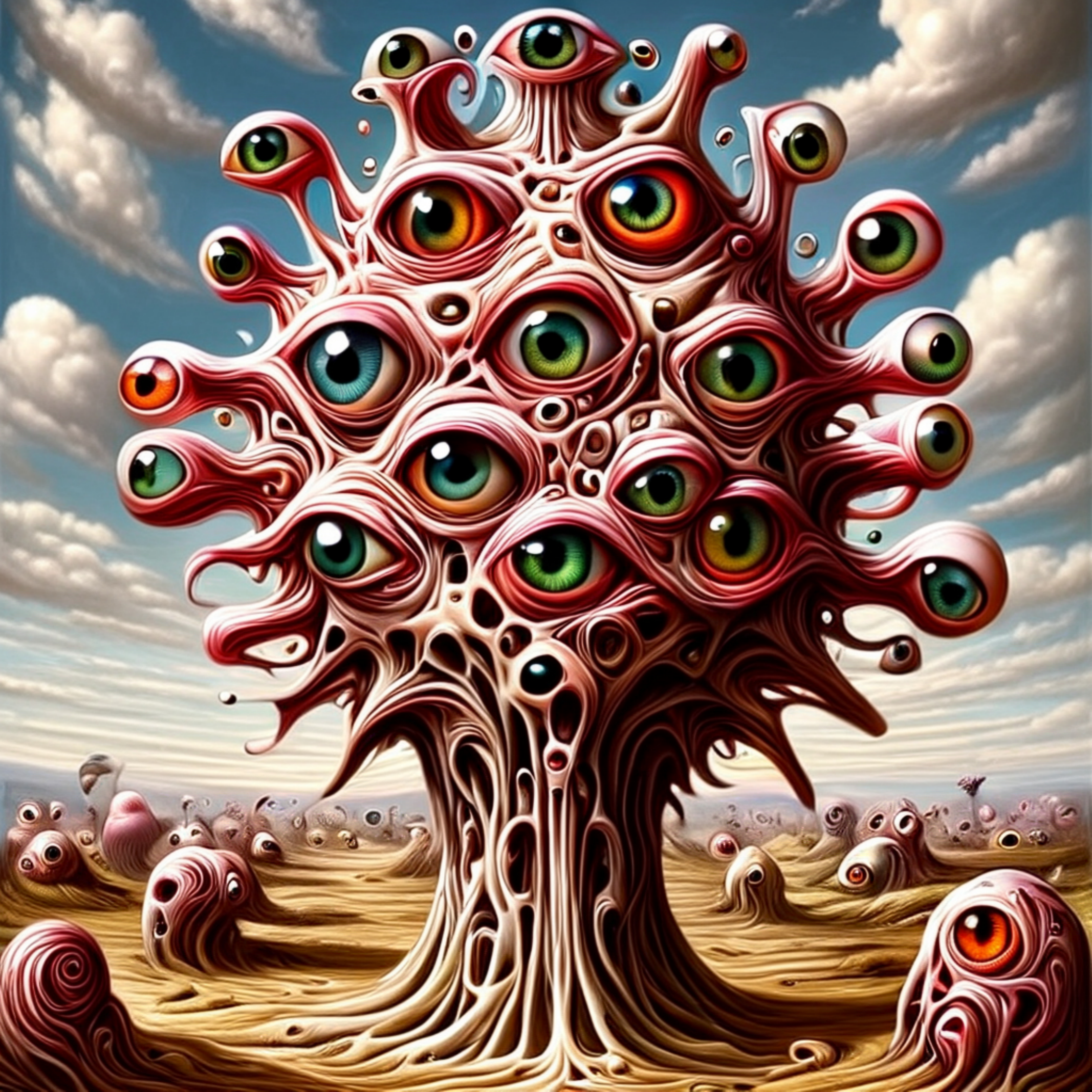}&%
\includegraphics[width=0.333333\textwidth]{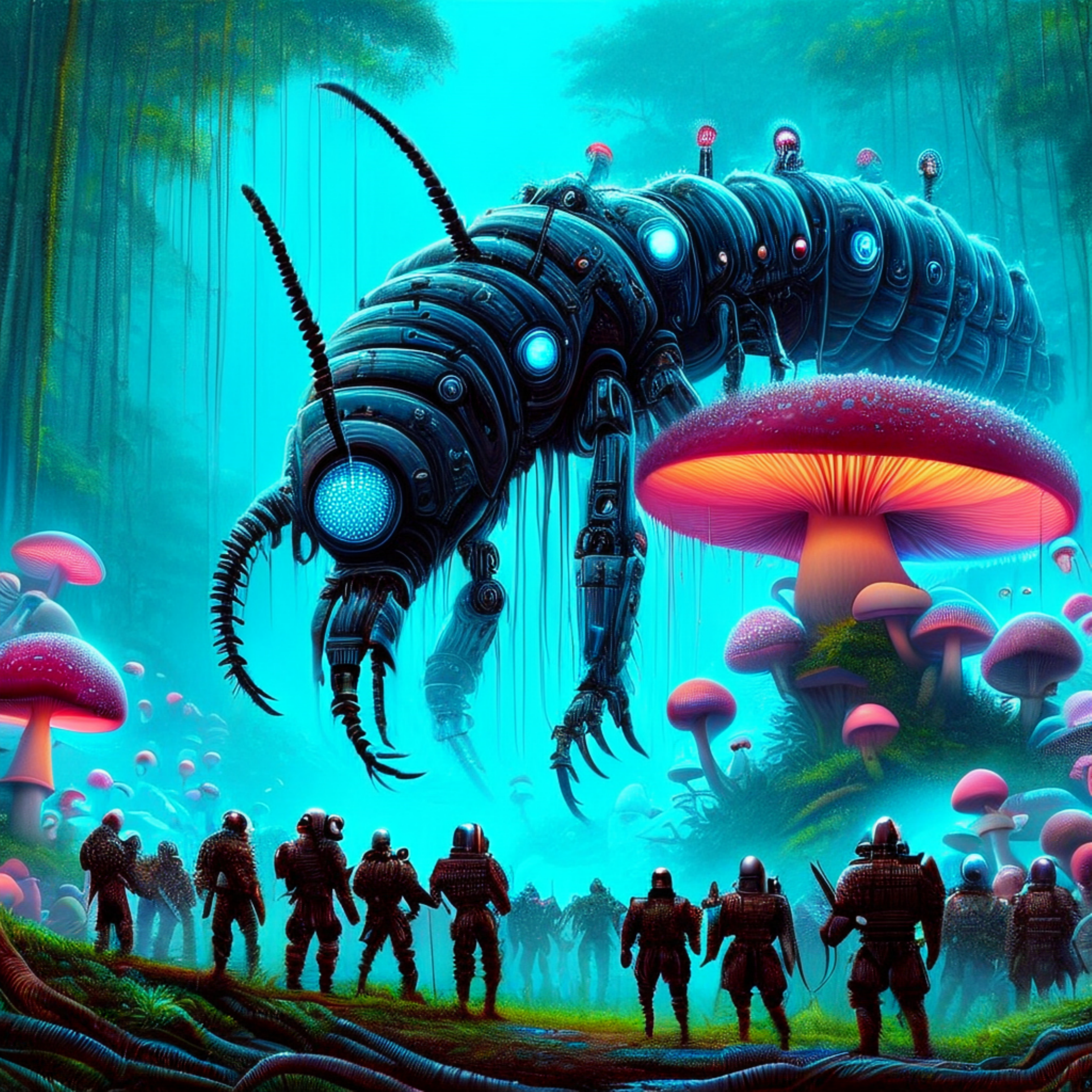}\\[-\lineskip]
\includegraphics[width=0.333333\textwidth]{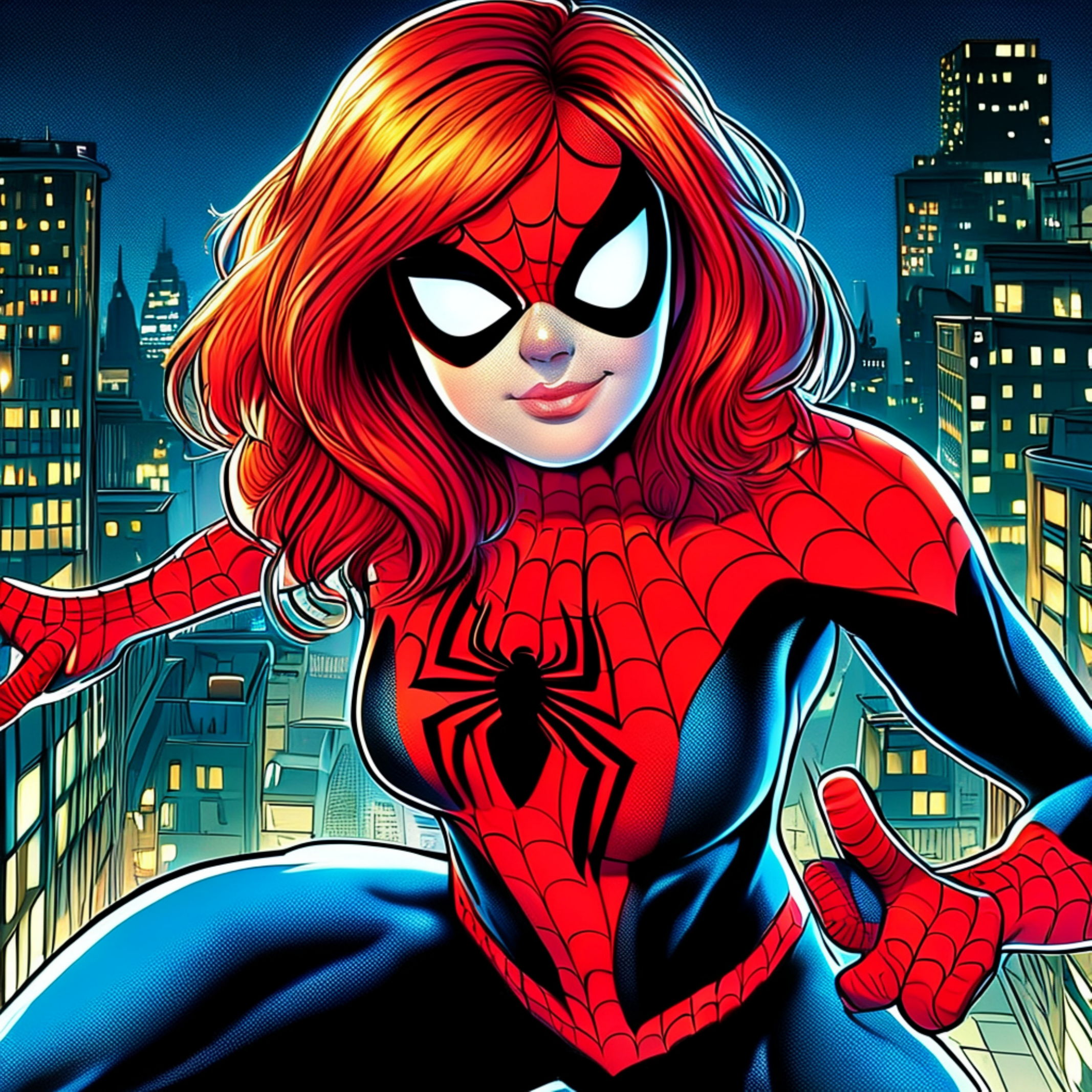}&%
\includegraphics[width=0.333333\textwidth]{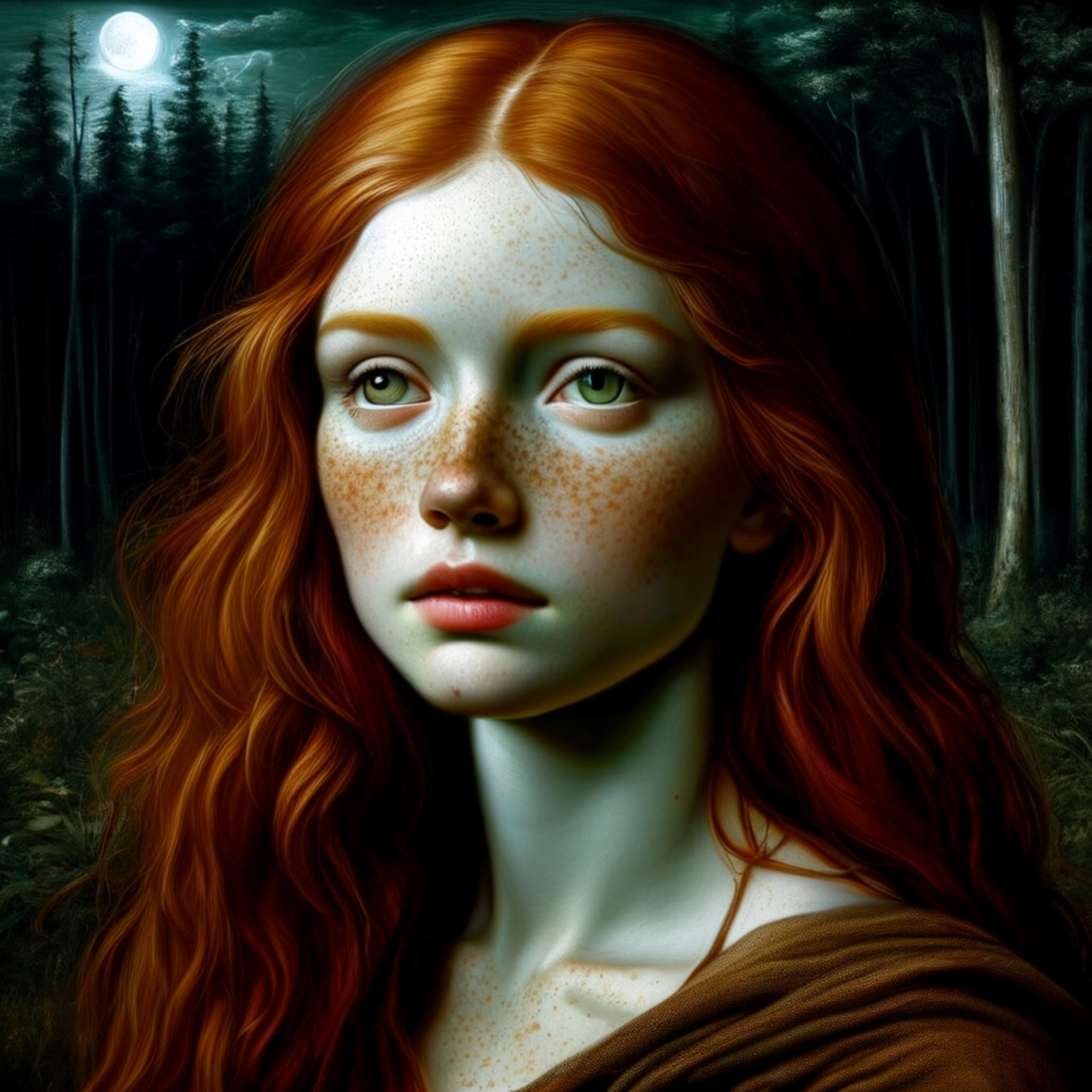}&%
\includegraphics[width=0.333333\textwidth]{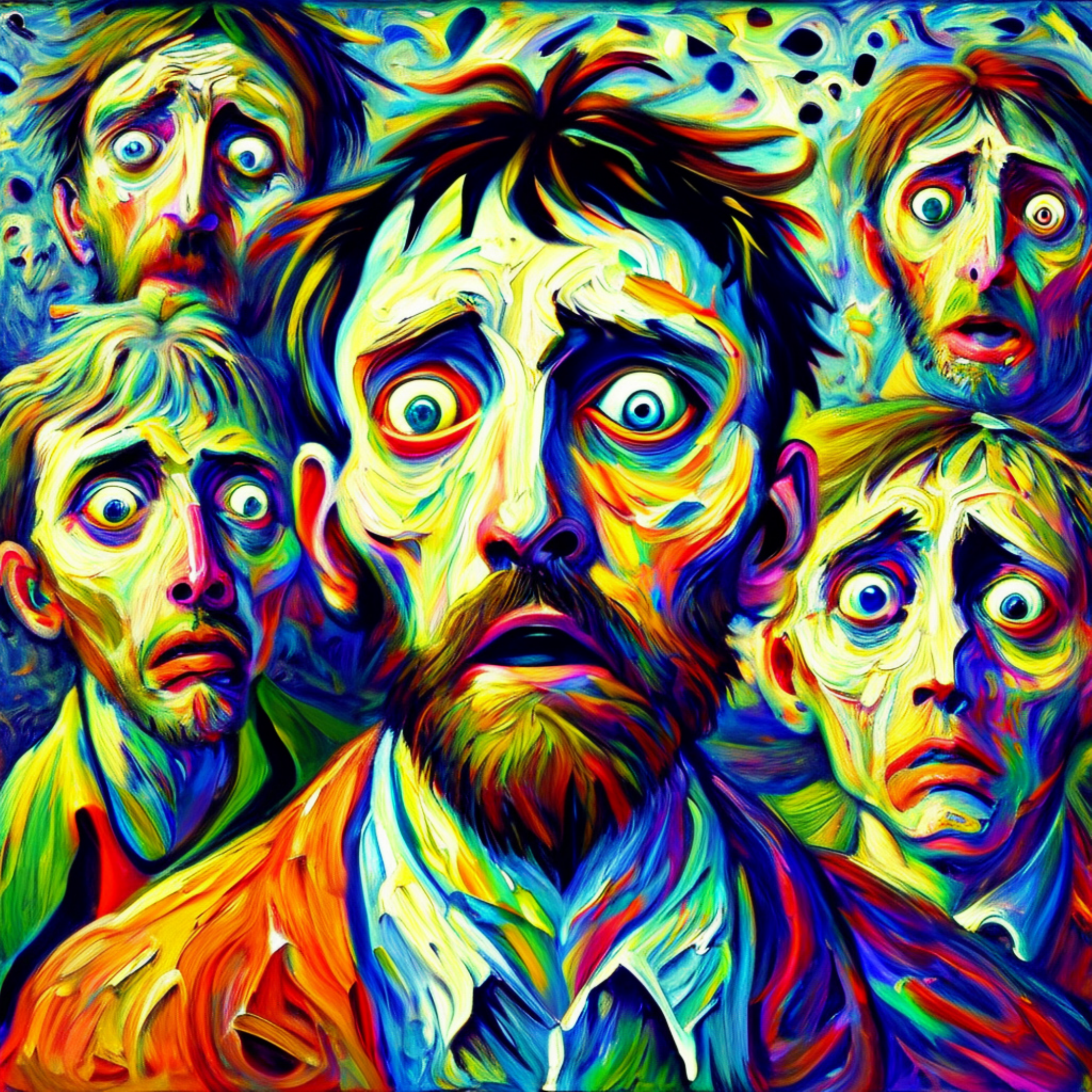}
\end{tabular}
\caption{EnLA text-to-image generation at $1024\times1024$ resolution.}
\label{fig:supp-dalle-remaining-2}
\end{figure*}
\clearpage

\begin{figure*}[p]
\centering
\setlength{\tabcolsep}{0pt}
\renewcommand{\arraystretch}{0}
\begin{tabular}{@{}ccc@{}}
\includegraphics[width=0.333333\textwidth]{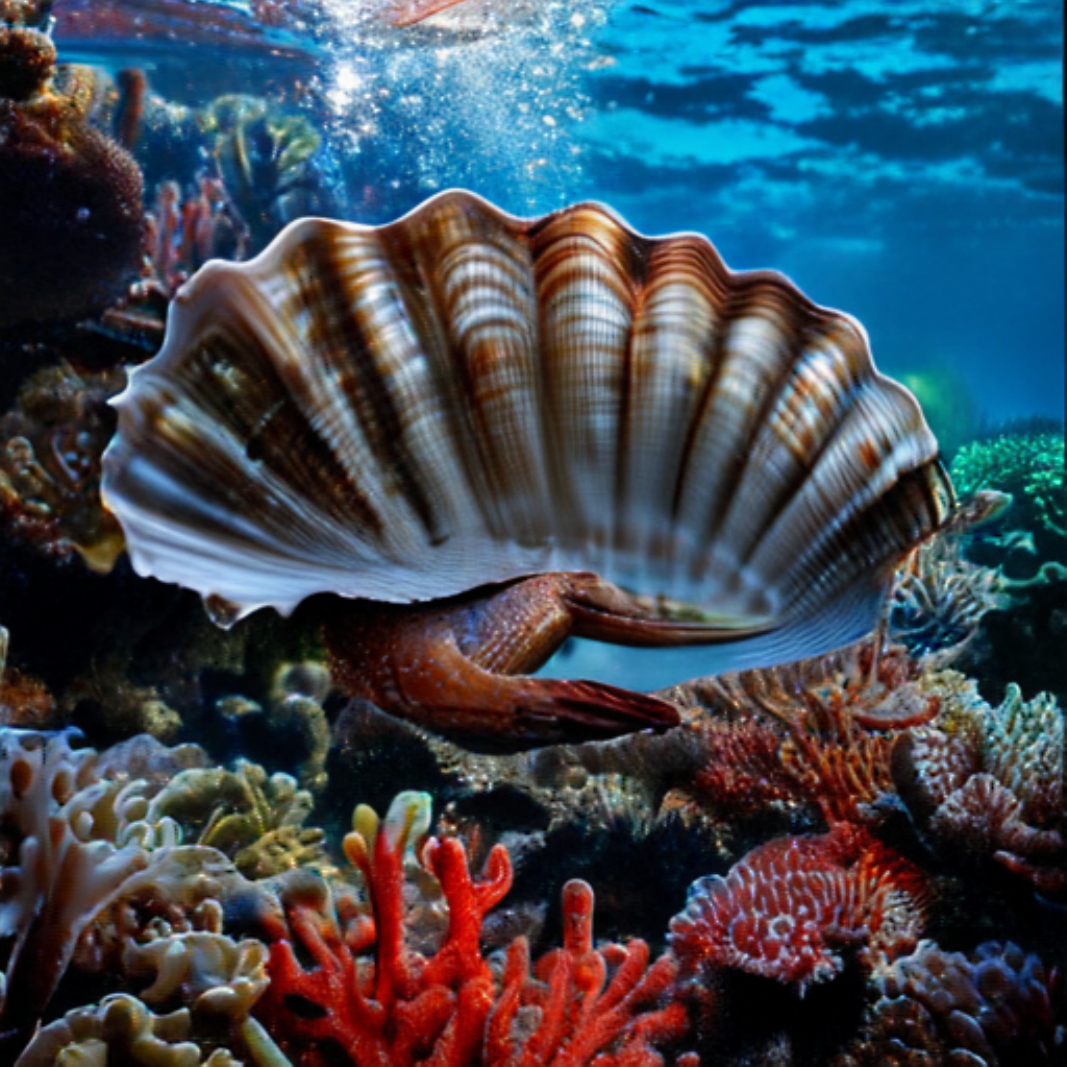}&%
\includegraphics[width=0.333333\textwidth]{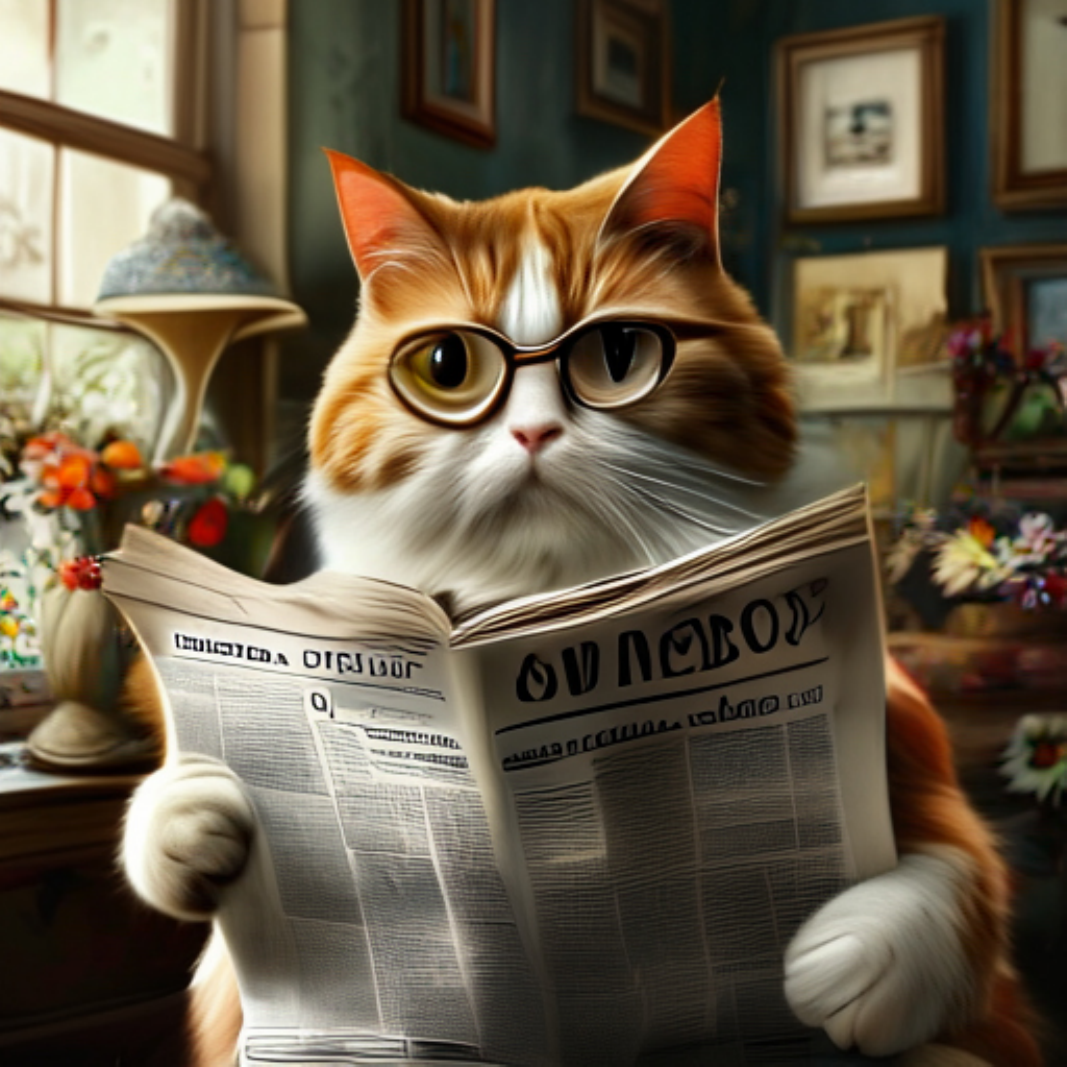}&%
\includegraphics[width=0.333333\textwidth]{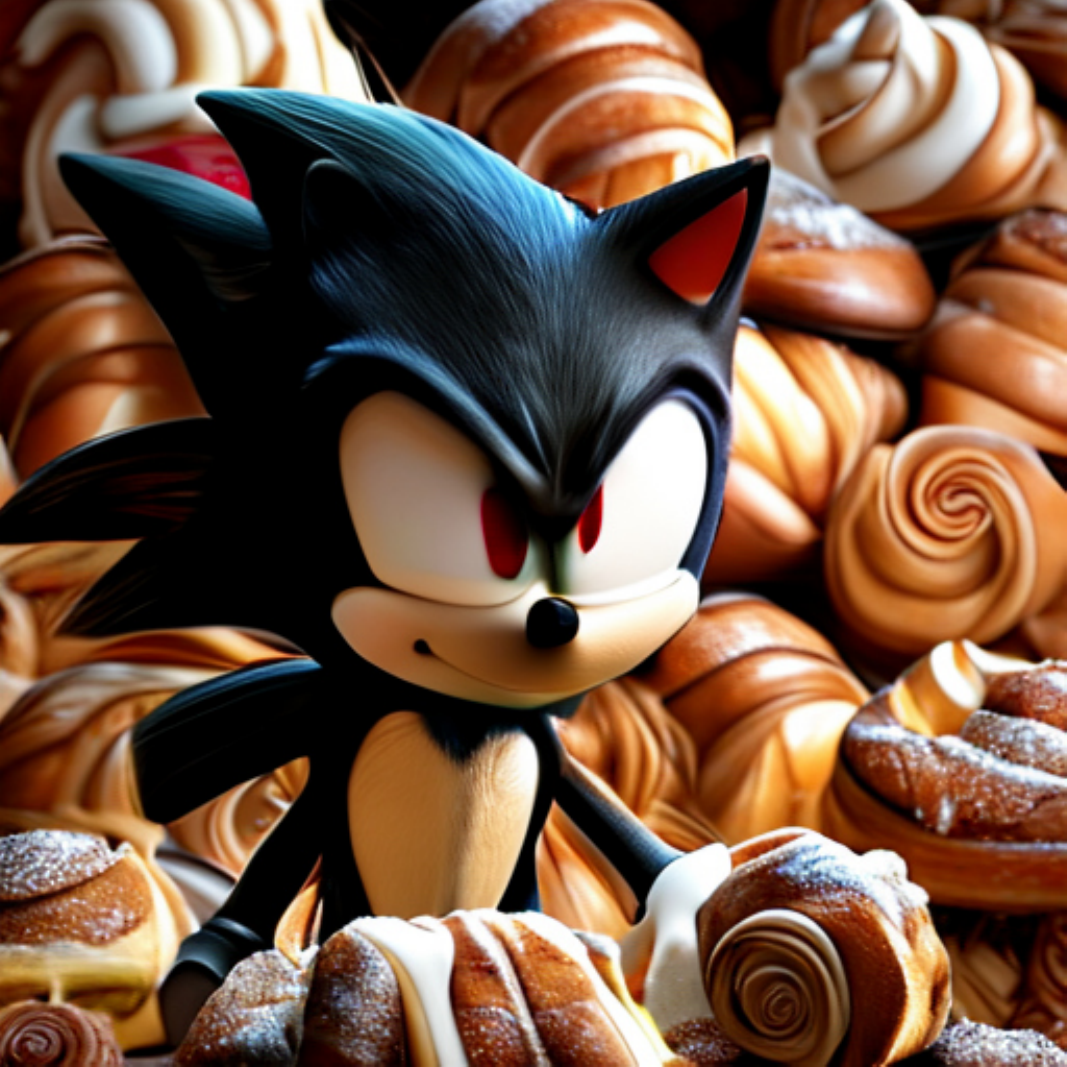}\\[-\lineskip]
\includegraphics[width=0.333333\textwidth]{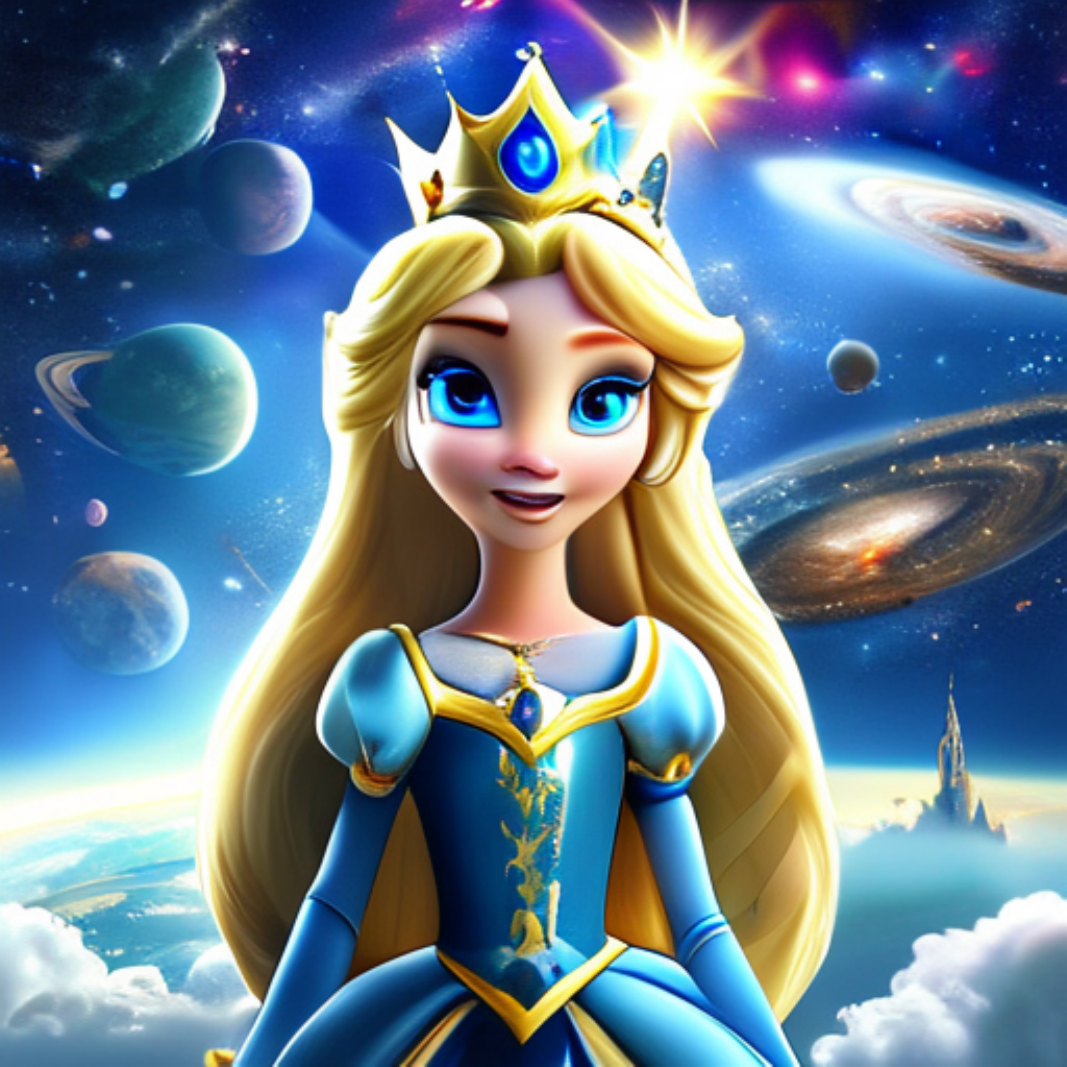}&%
\includegraphics[width=0.333333\textwidth]{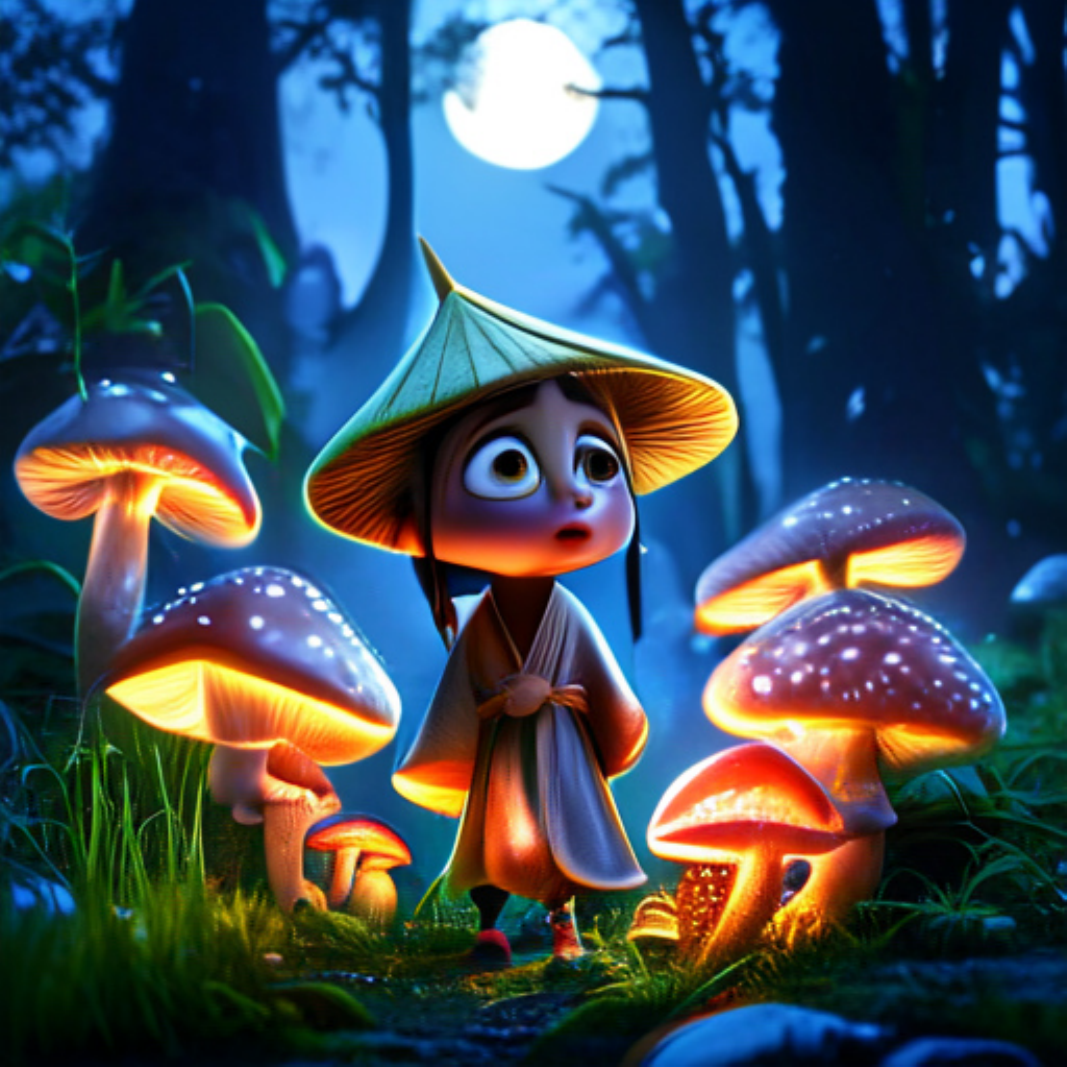}&%
\includegraphics[width=0.333333\textwidth]{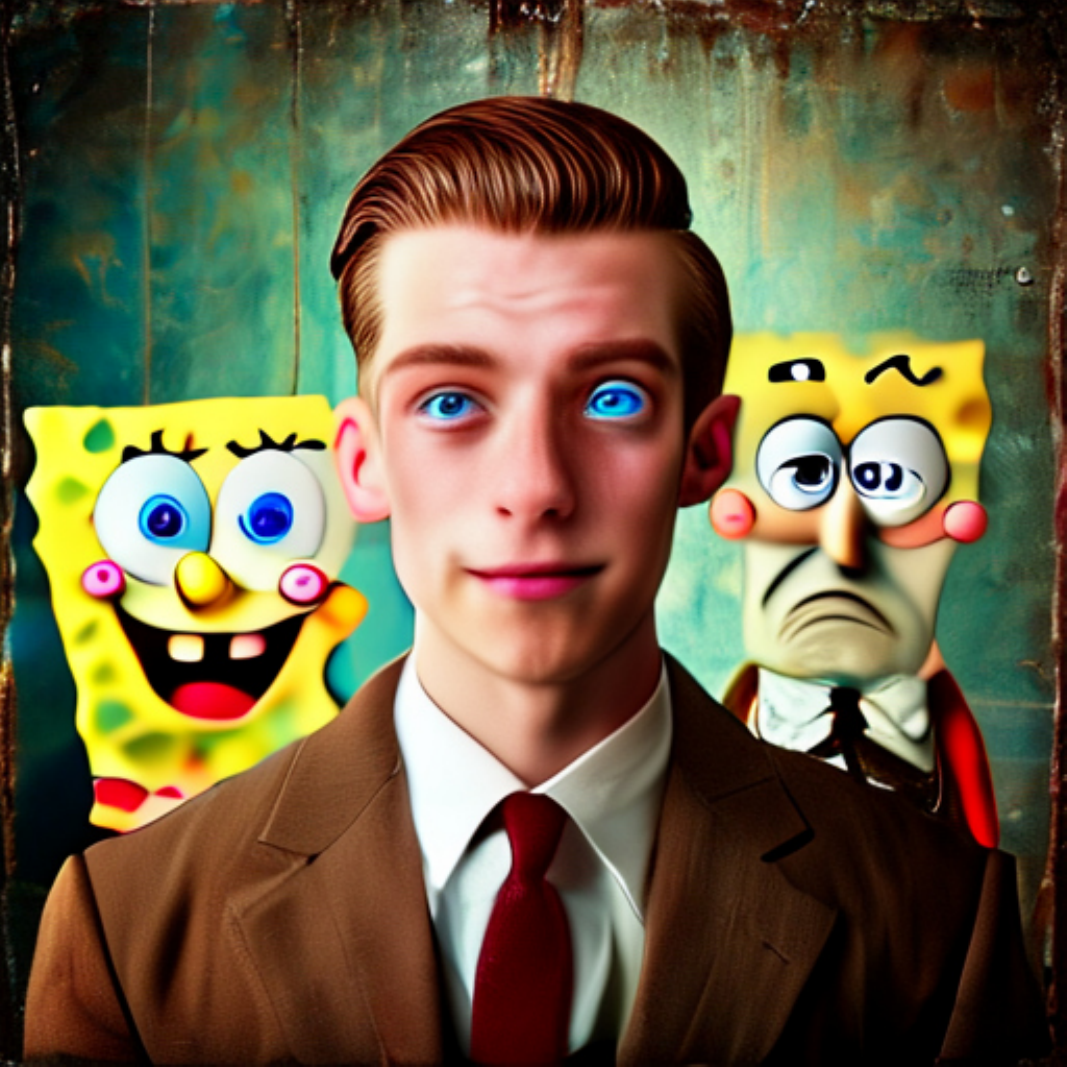}\\[-\lineskip]
\includegraphics[width=0.333333\textwidth]{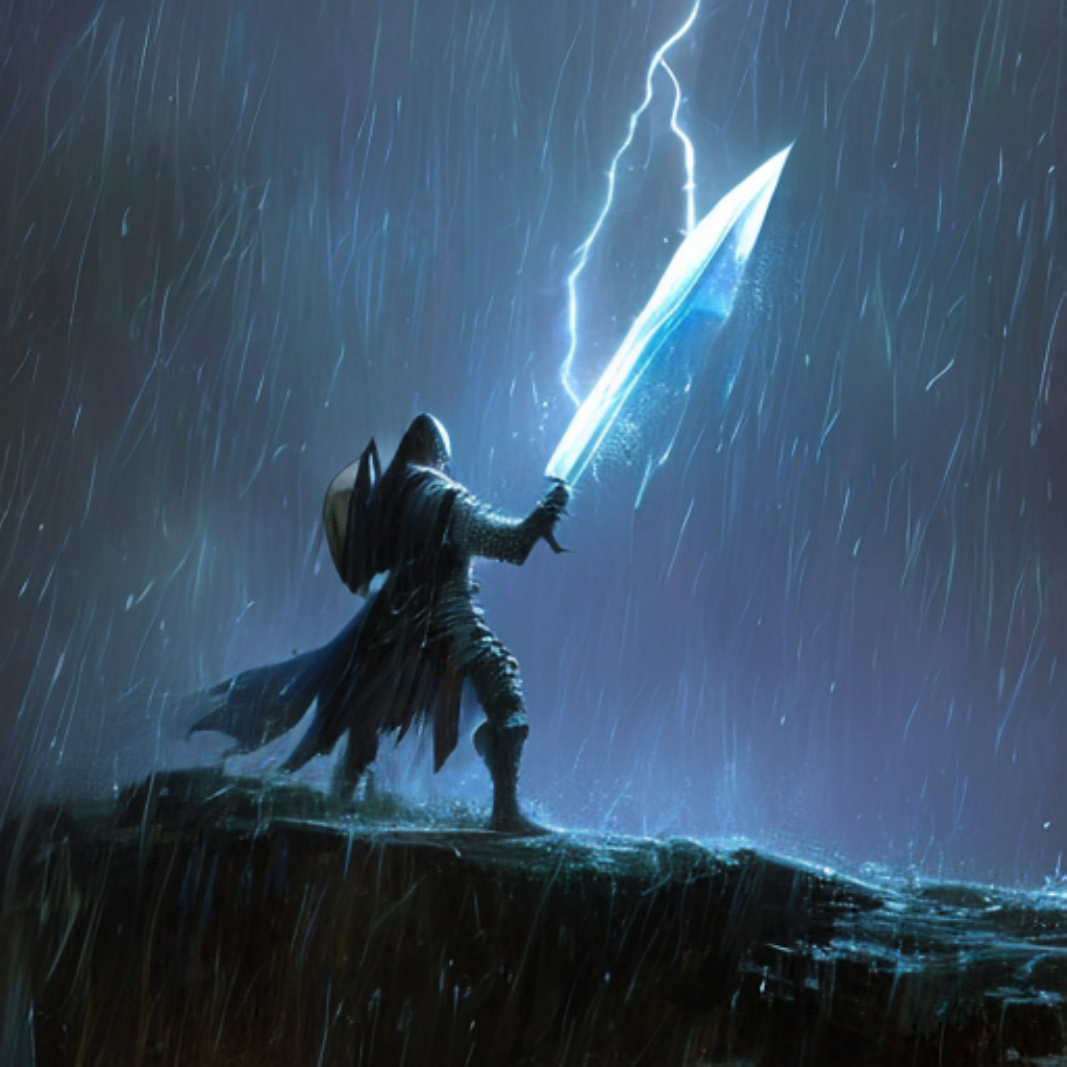}&%
\includegraphics[width=0.333333\textwidth]{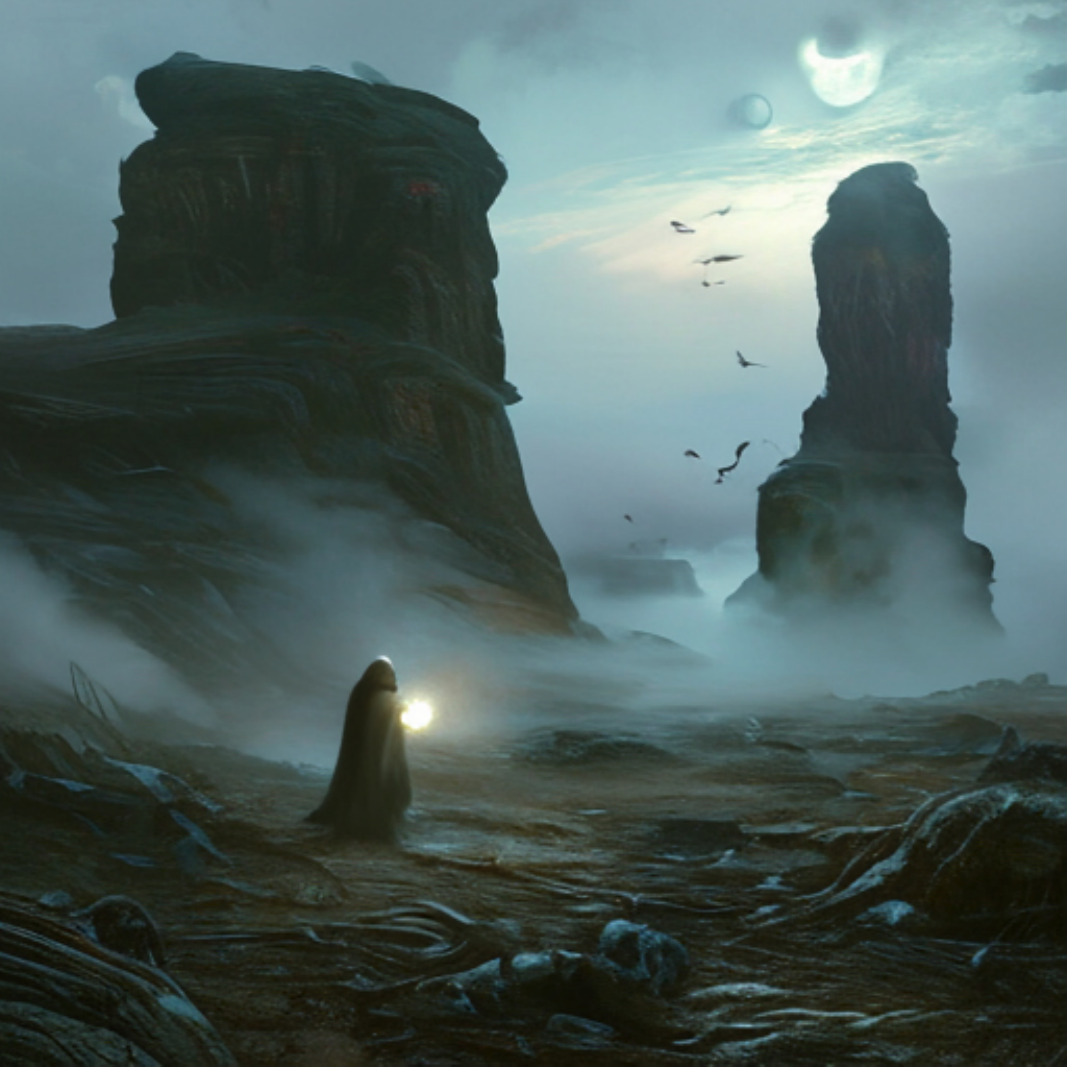}&%
\includegraphics[width=0.333333\textwidth]{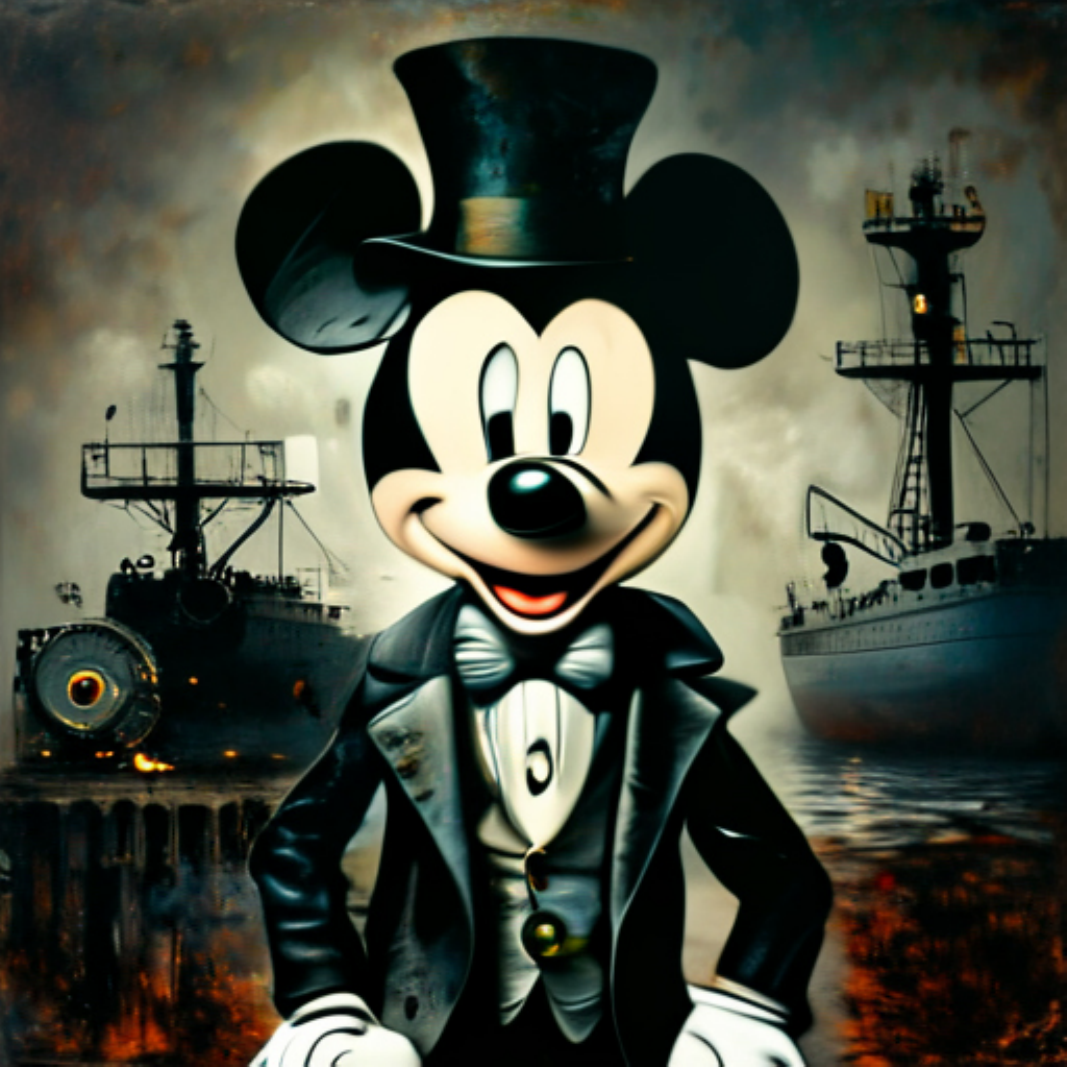}
\end{tabular}
\caption{EnLA text-to-image generation at $512\times512$ resolution.}
\label{fig:supp-midjourney-1}
\end{figure*}
\clearpage

\begin{figure*}[p]
\centering
\setlength{\tabcolsep}{0pt}
\renewcommand{\arraystretch}{0}
\begin{tabular}{@{}ccc@{}}
\includegraphics[width=0.333333\textwidth]{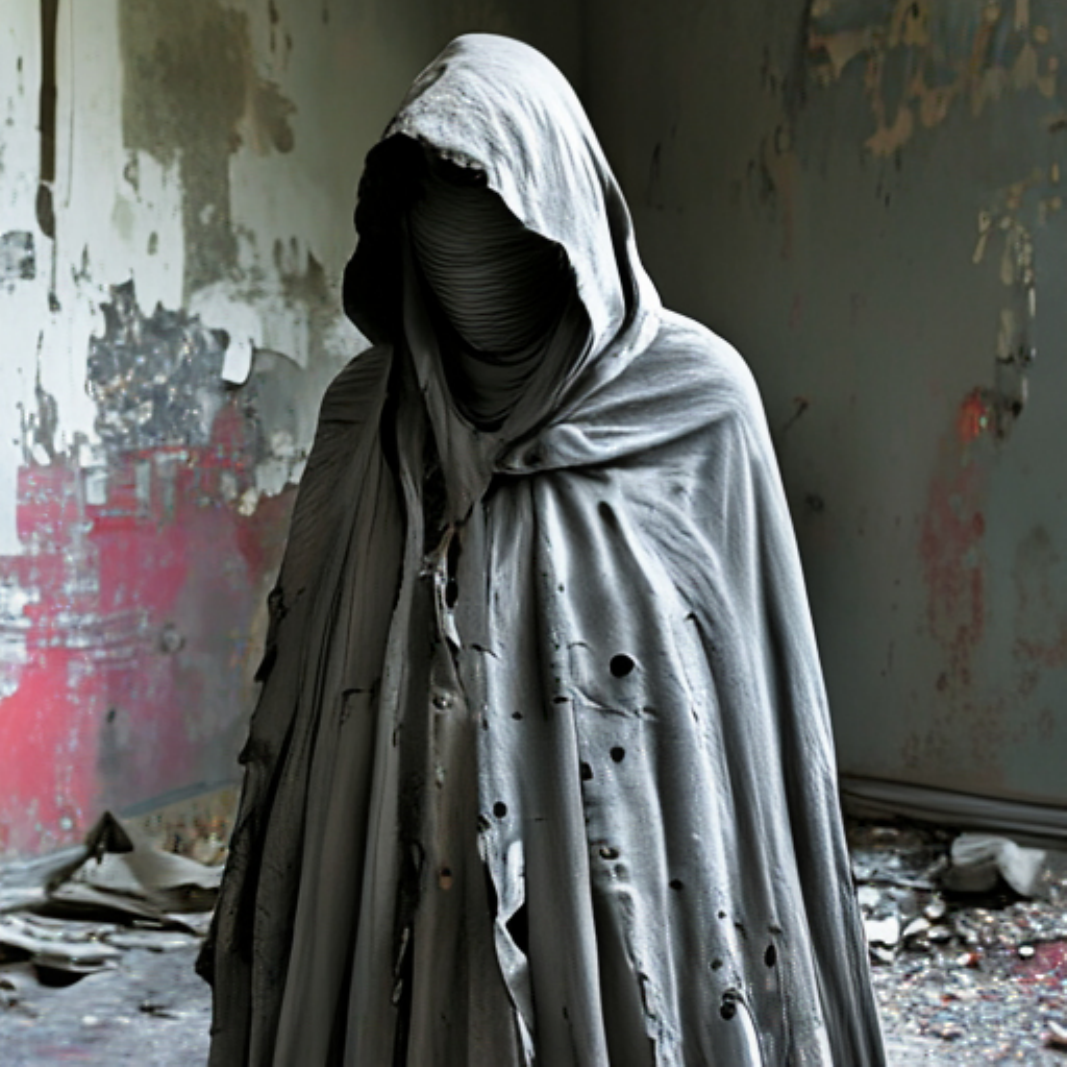}&%
\includegraphics[width=0.333333\textwidth]{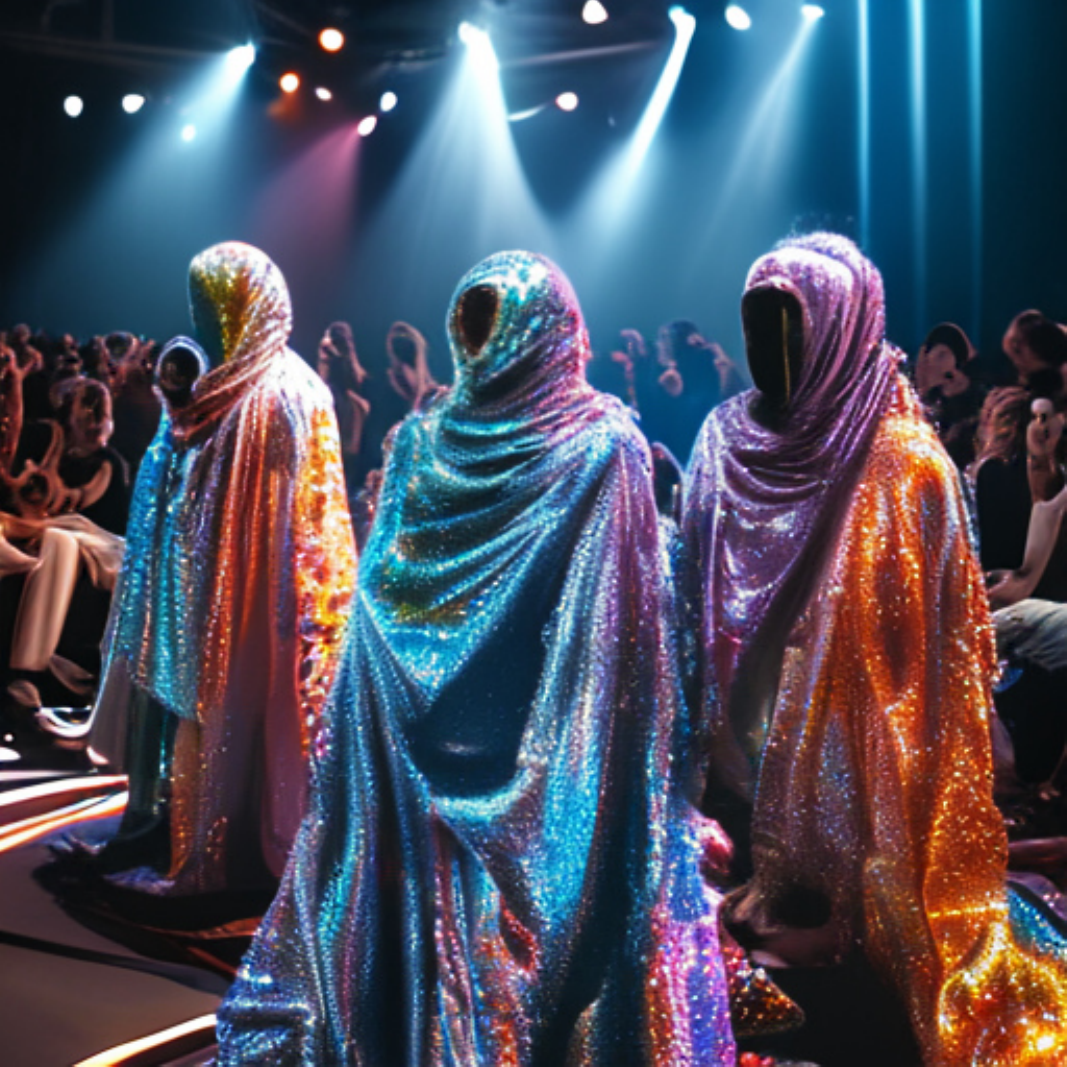}&%
\includegraphics[width=0.333333\textwidth]{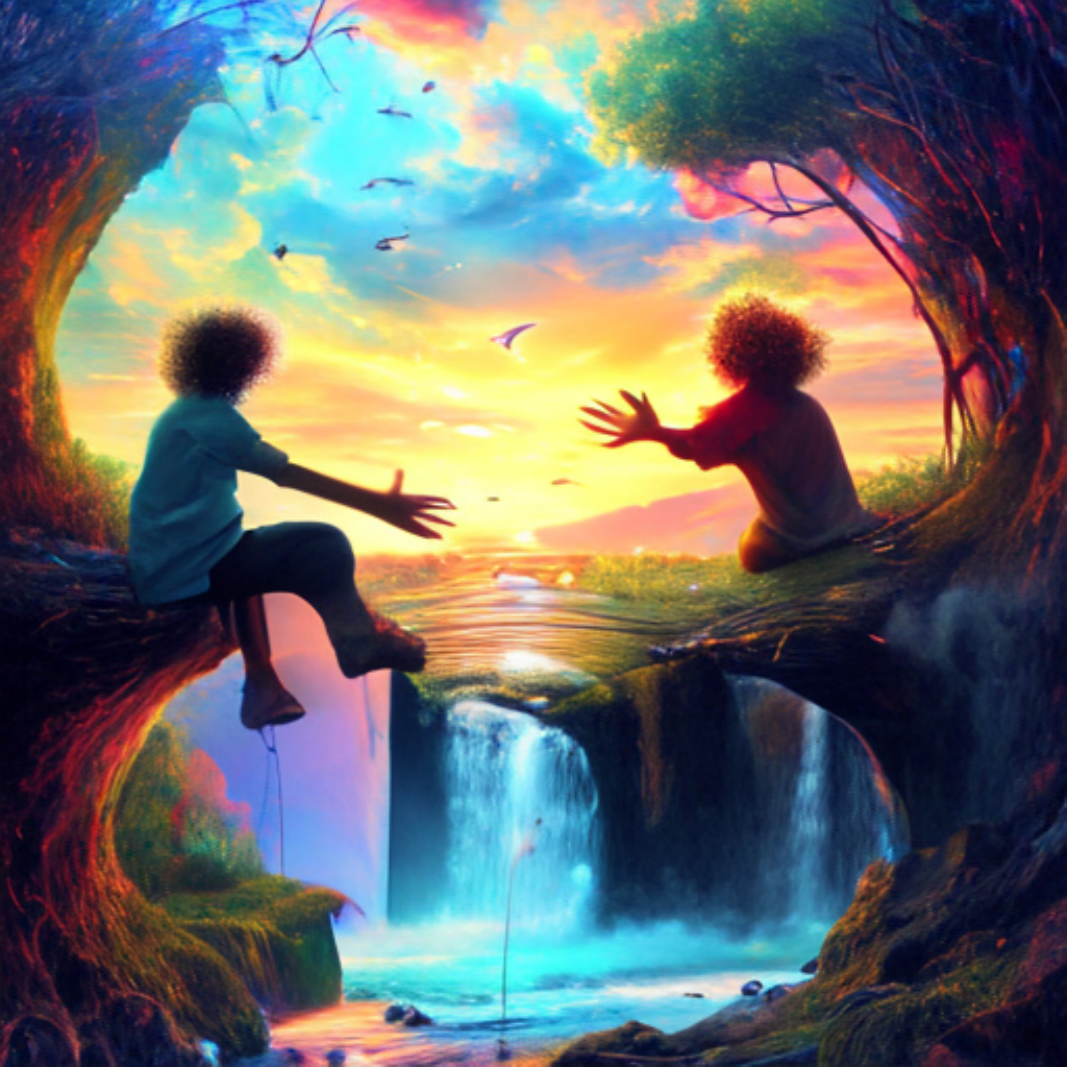}\\[-\lineskip]
\includegraphics[width=0.333333\textwidth]{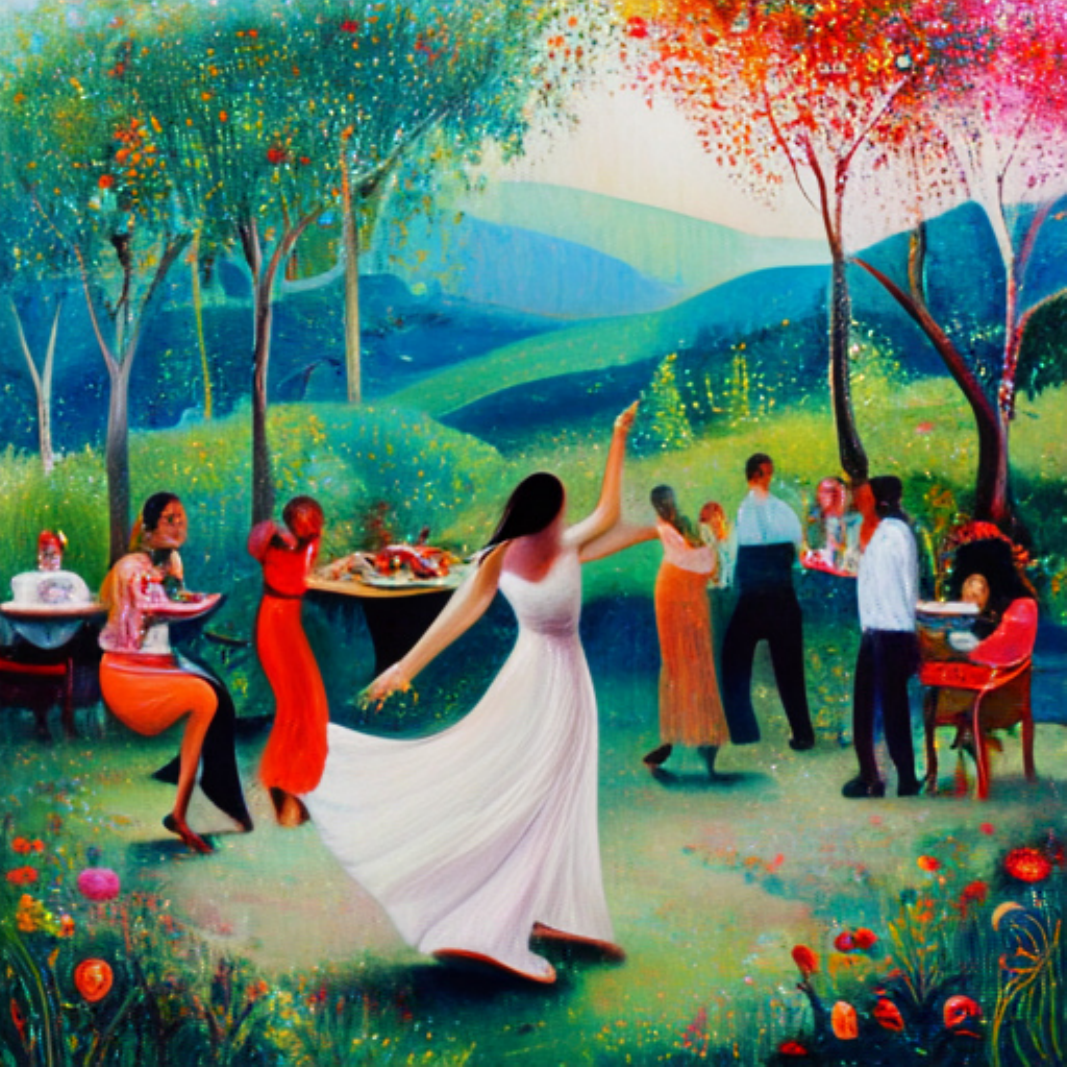}&%
\includegraphics[width=0.333333\textwidth]{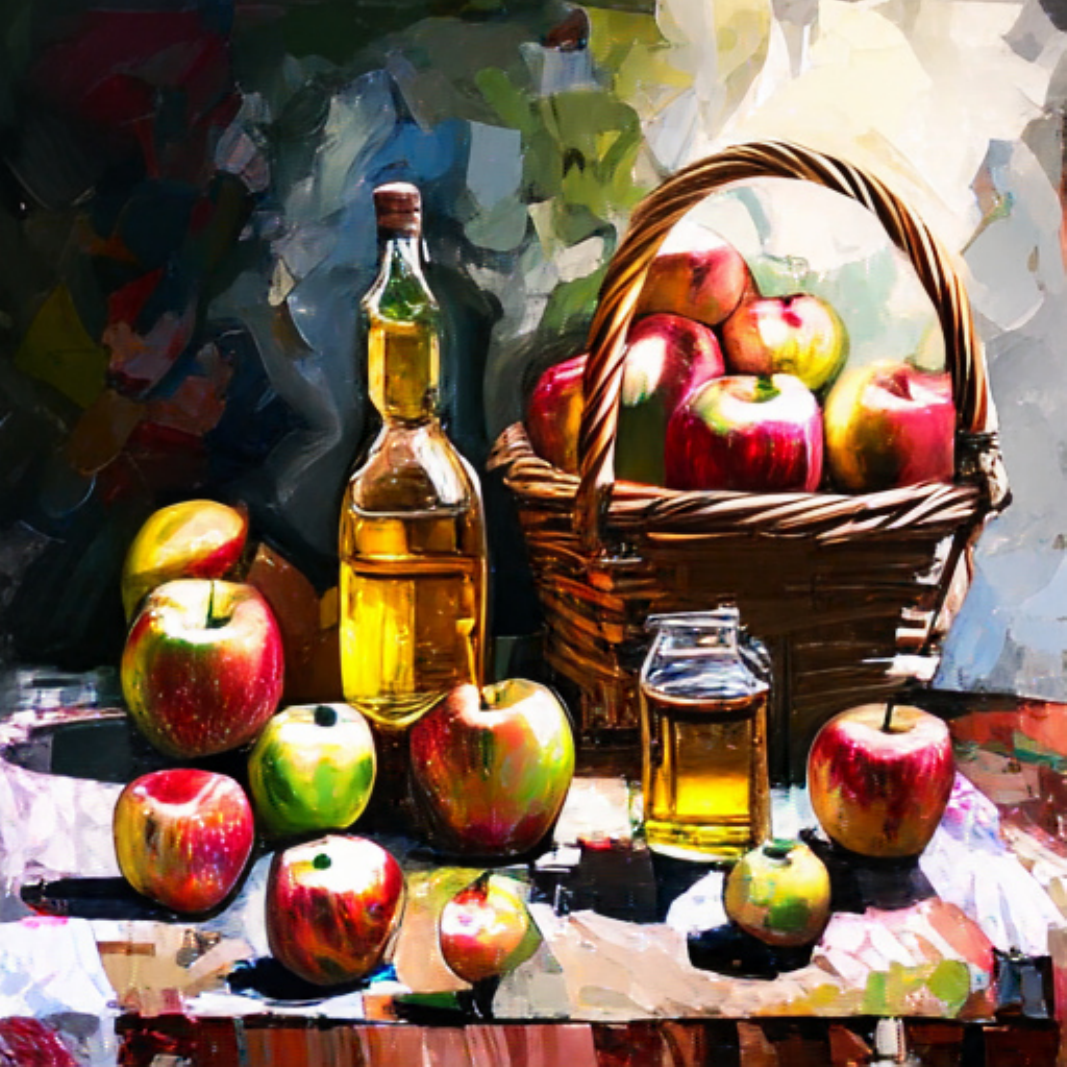}&%
\includegraphics[width=0.333333\textwidth]{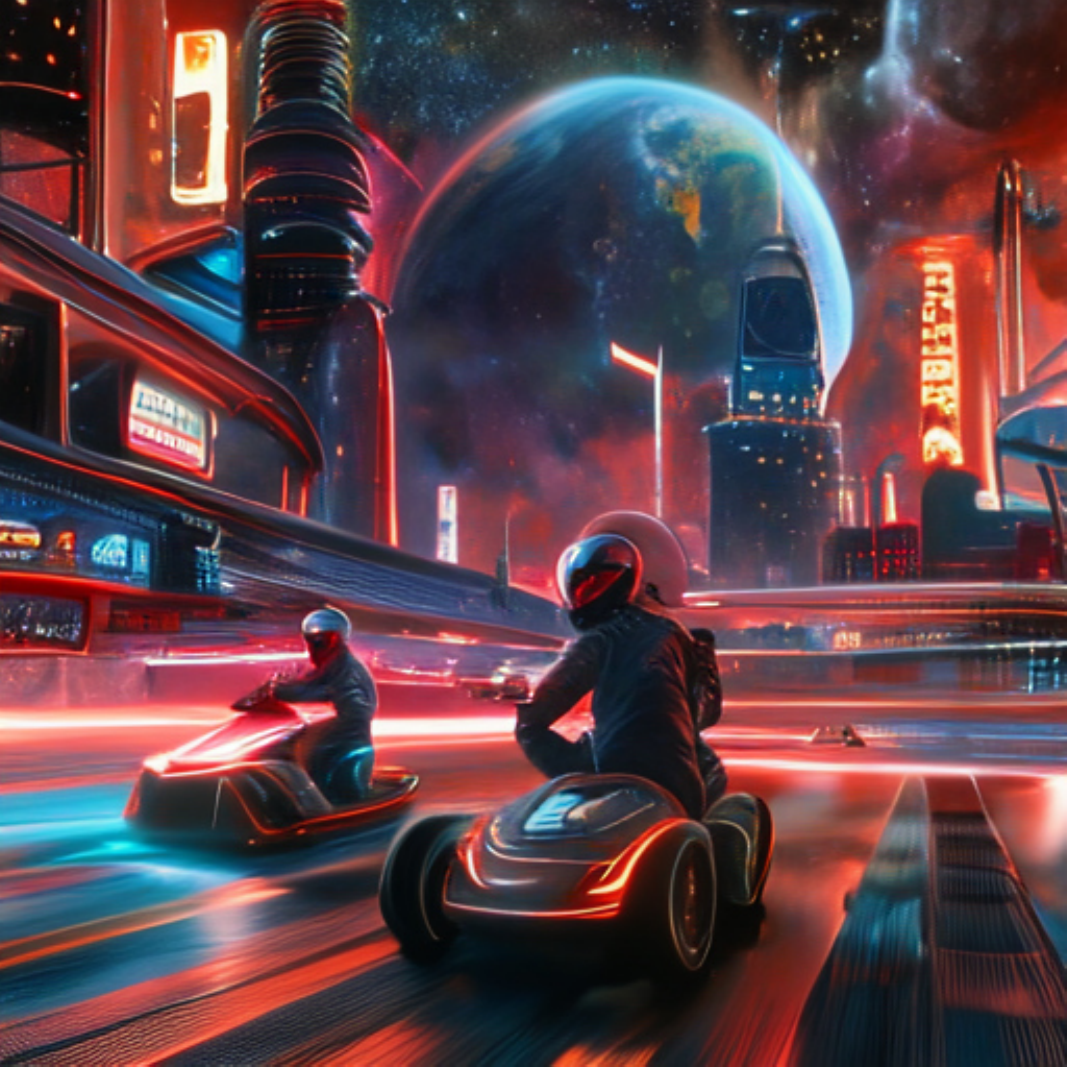}\\[-\lineskip]
\includegraphics[width=0.333333\textwidth]{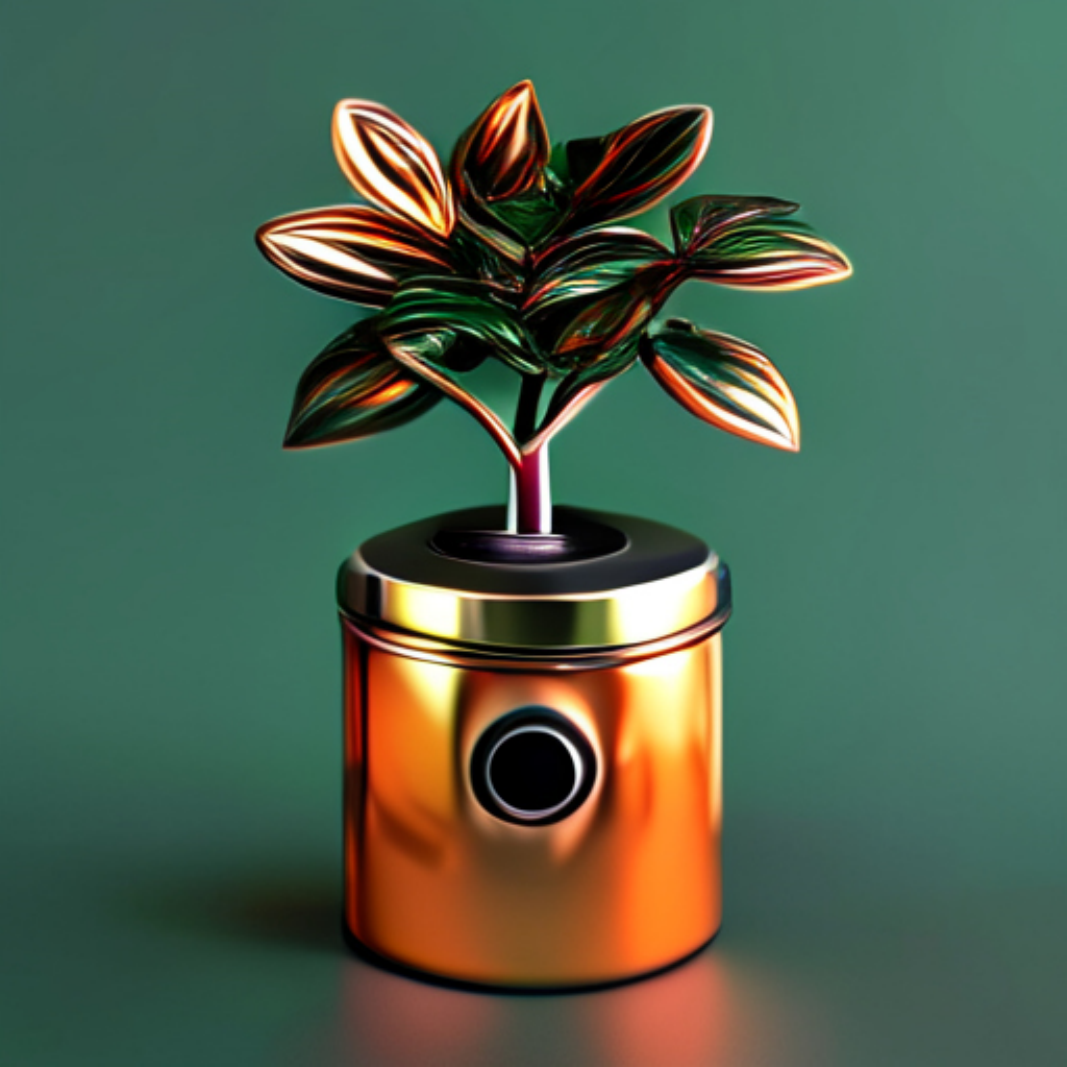}&%
\includegraphics[width=0.333333\textwidth]{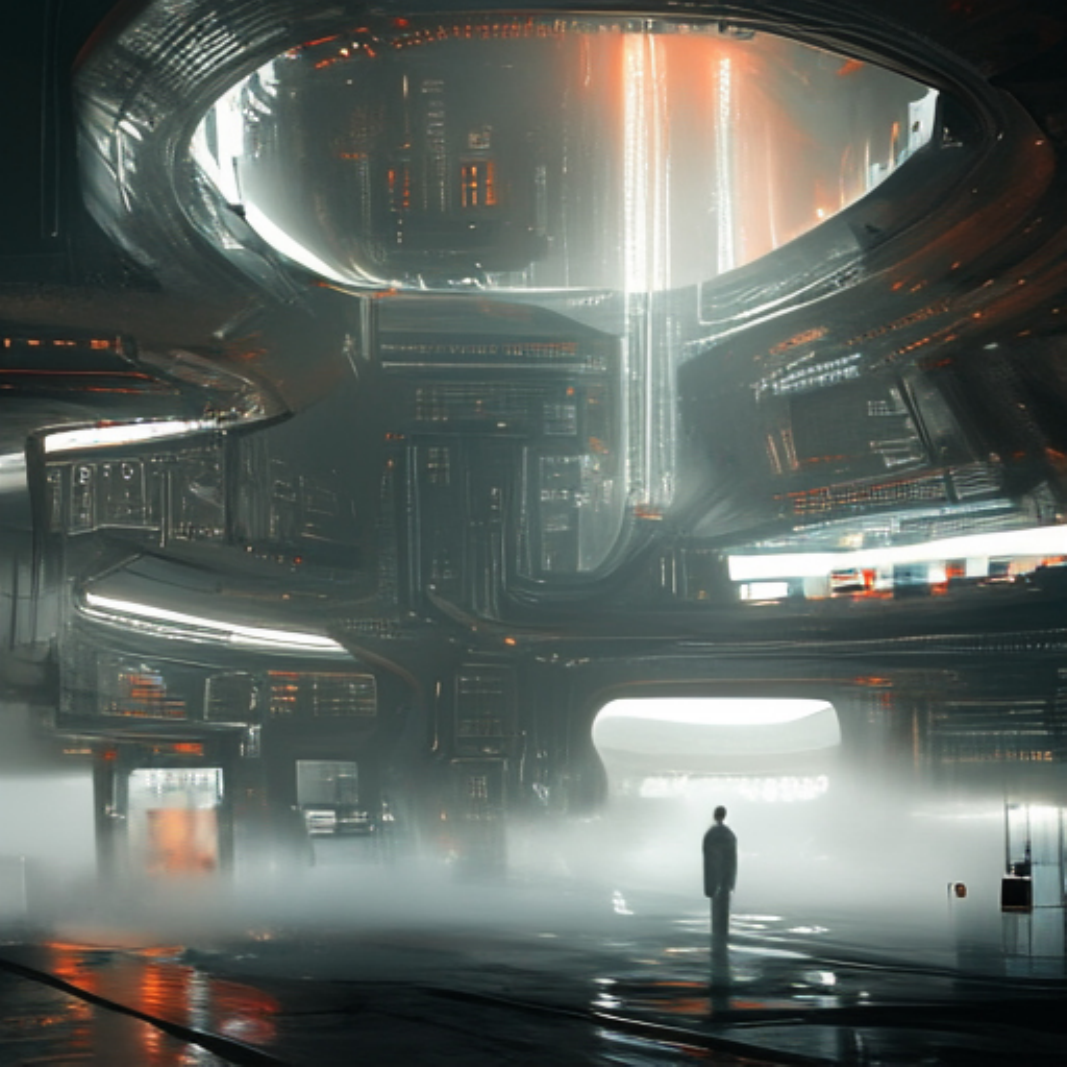}&%
\includegraphics[width=0.333333\textwidth]{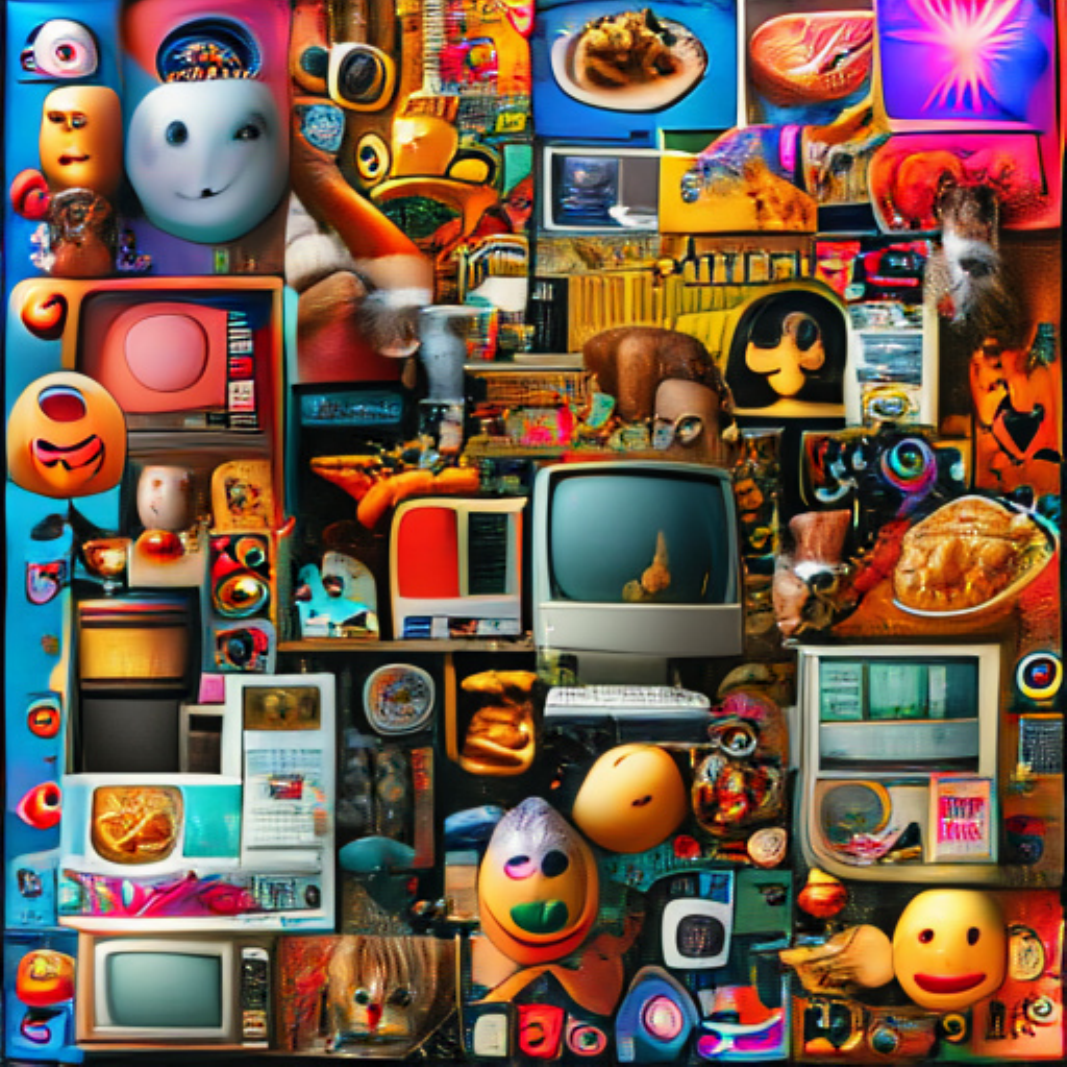}
\end{tabular}
\caption{EnLA text-to-image generation at $512\times512$ resolution.}
\label{fig:supp-midjourney-2}
\end{figure*}
\clearpage

\begin{figure*}[p]
\centering
\setlength{\tabcolsep}{0pt}
\renewcommand{\arraystretch}{0}
\begin{tabular}{@{}ccc@{}}
\includegraphics[width=0.333333\textwidth]{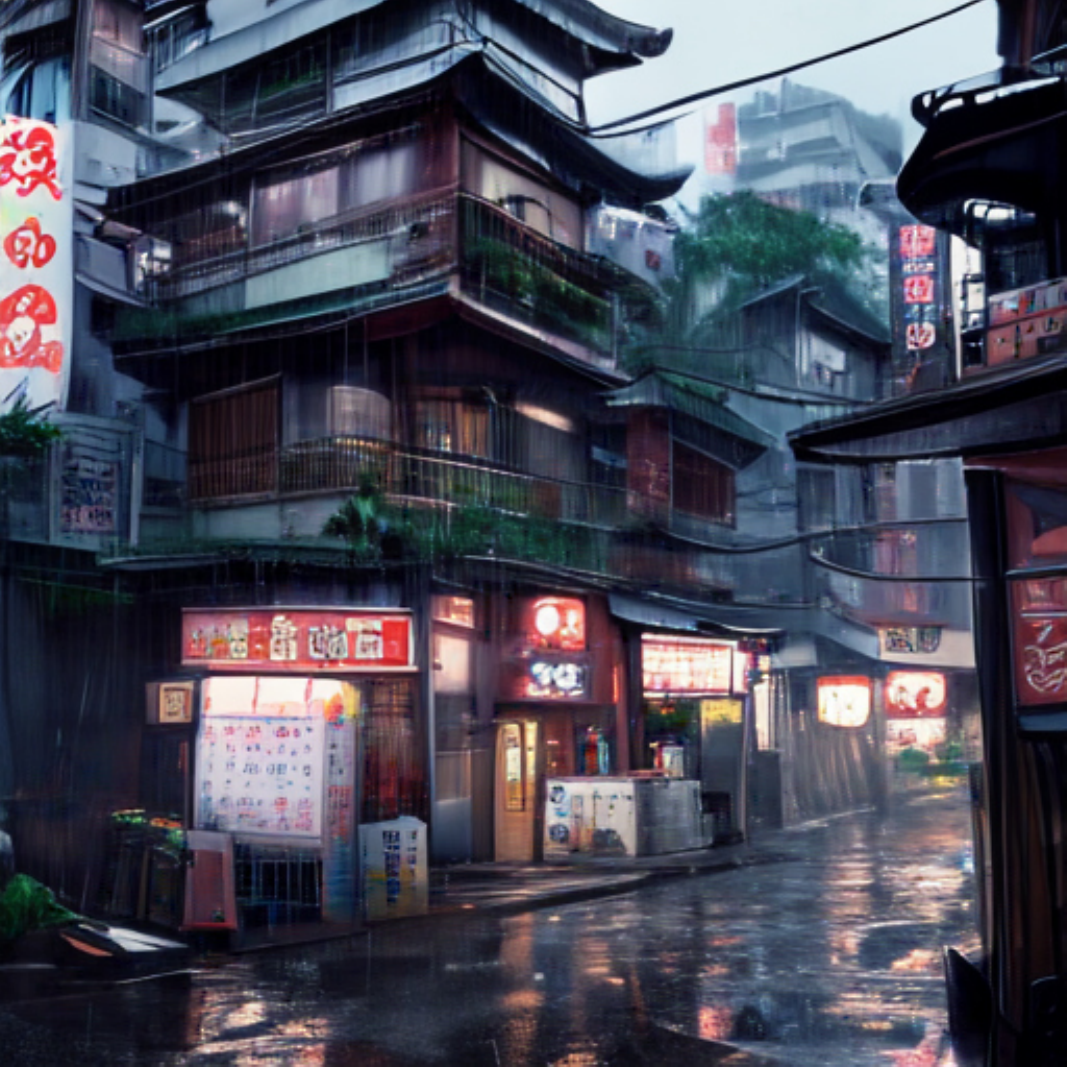}&%
\includegraphics[width=0.333333\textwidth]{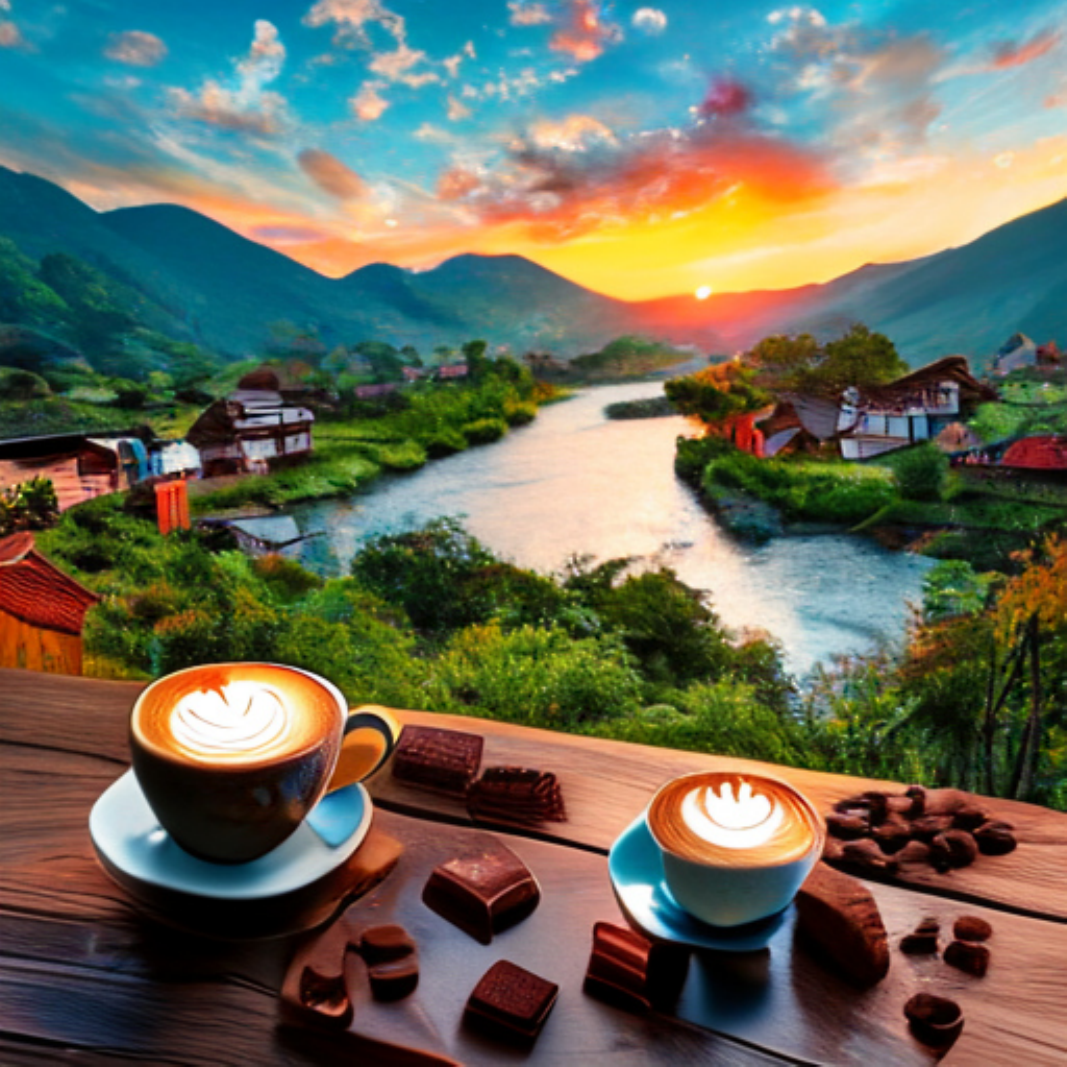}&%
\includegraphics[width=0.333333\textwidth]{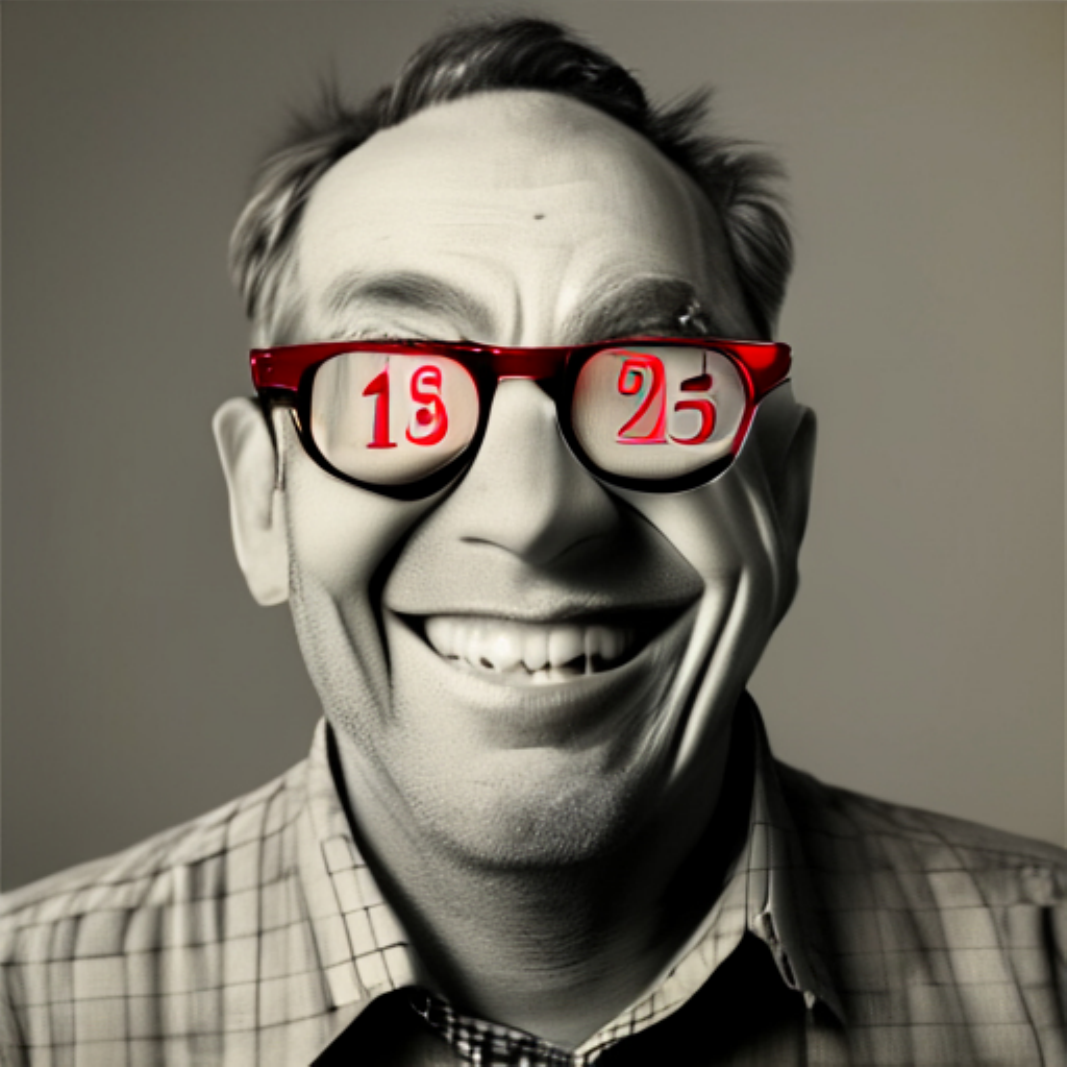}\\[-\lineskip]
\multicolumn{3}{c}{%
\includegraphics[width=0.333333\textwidth]{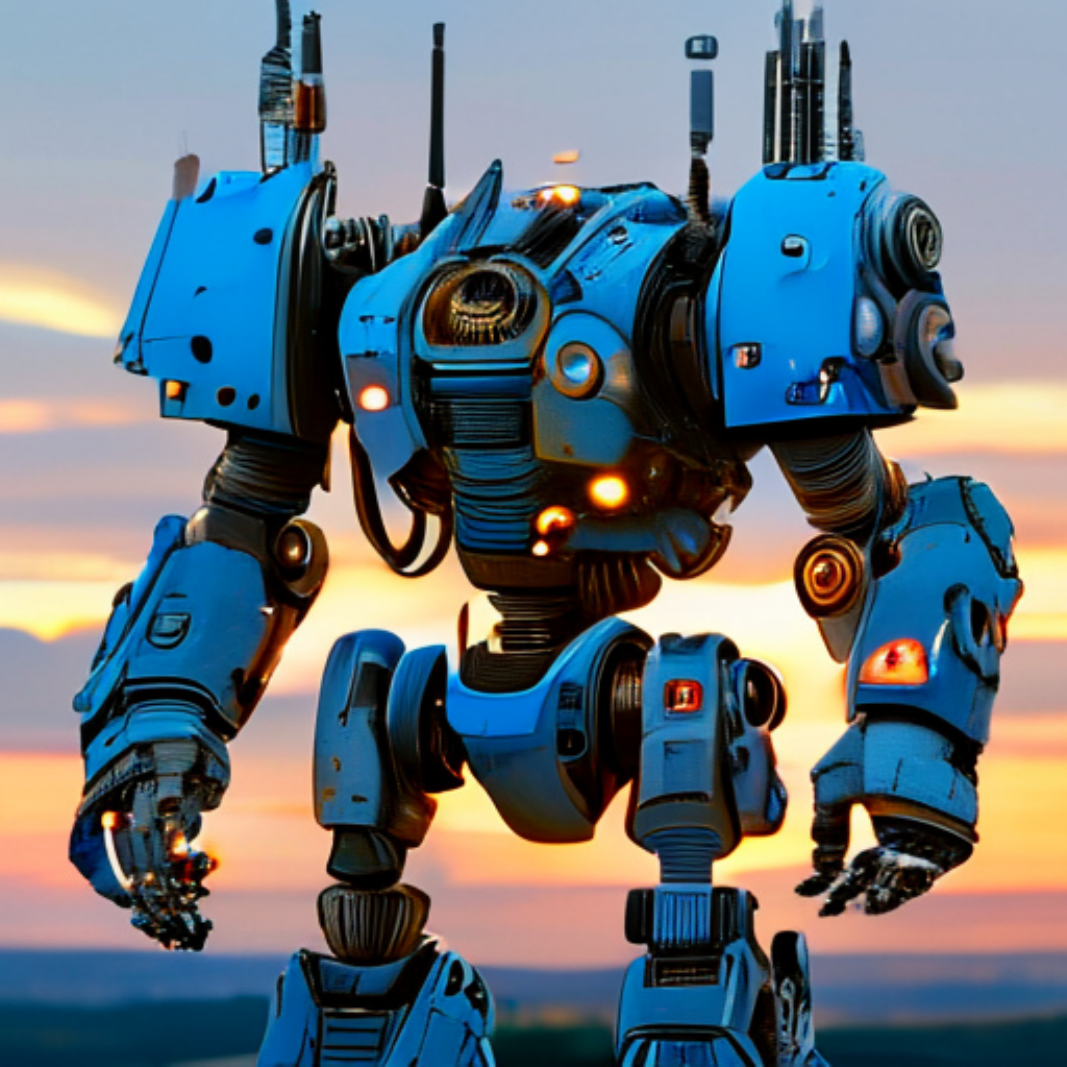}%
\includegraphics[width=0.333333\textwidth]{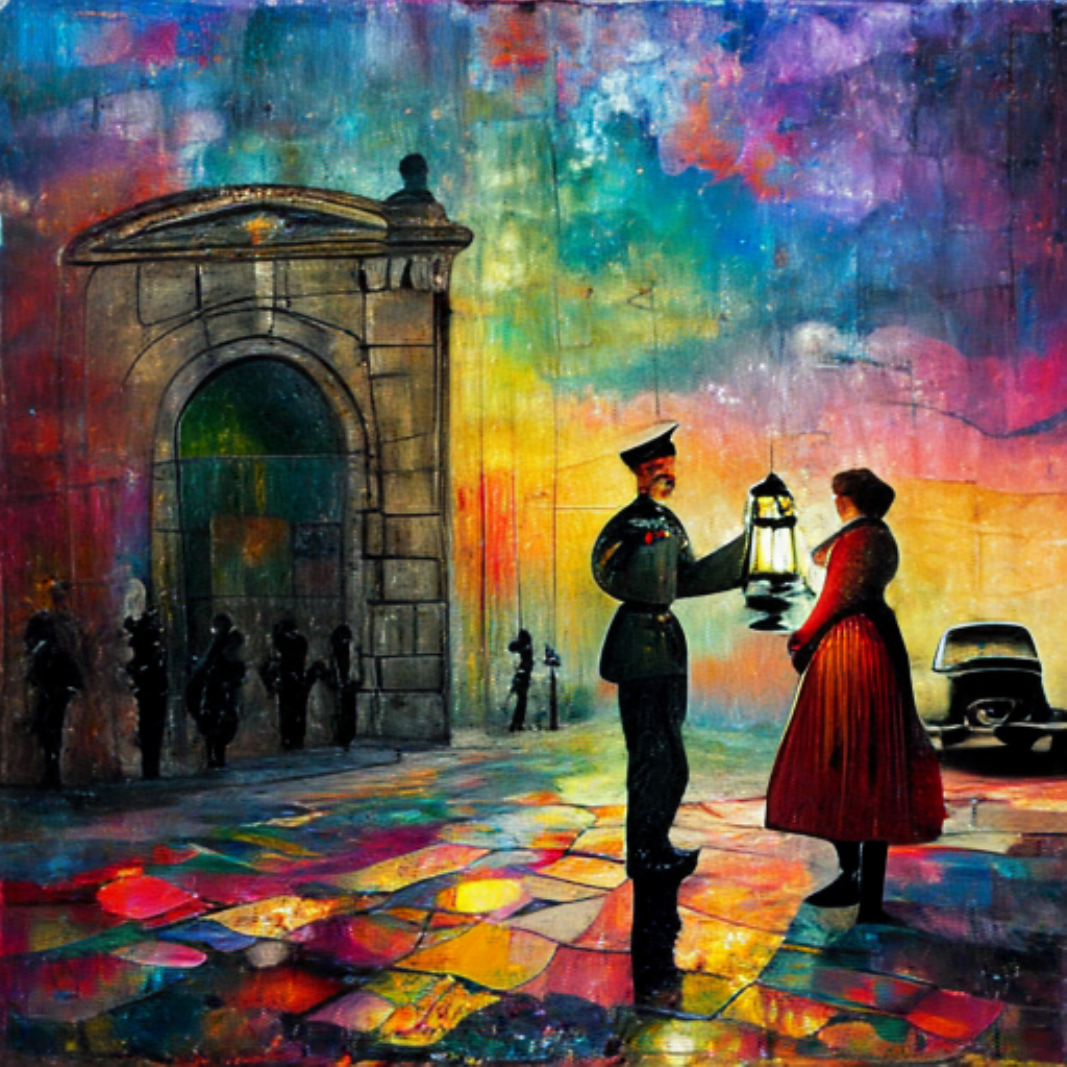}}
\end{tabular}
\caption{EnLA text-to-image generation at $512\times512$ resolution.}
\label{fig:supp-midjourney-3}
\end{figure*}
\clearpage

\begin{figure*}[p]
\centering
\setlength{\tabcolsep}{0pt}
\renewcommand{\arraystretch}{0}
\begin{tabular}{@{}ccc@{}}
\includegraphics[width=0.333333\textwidth]{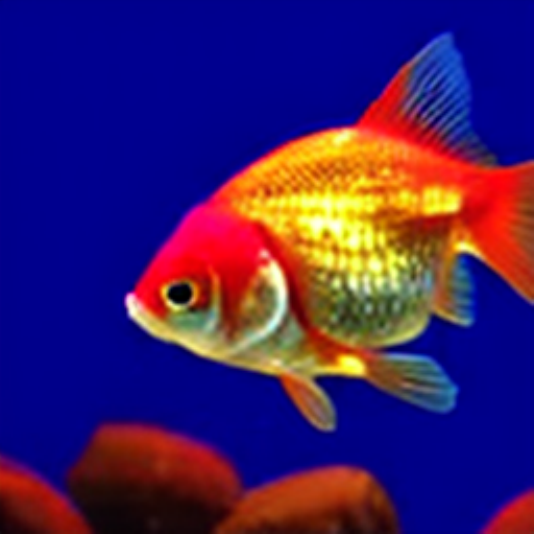}&%
\includegraphics[width=0.333333\textwidth]{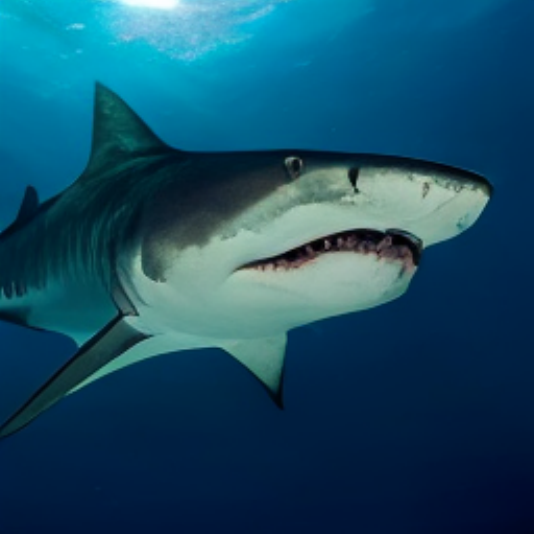}&%
\includegraphics[width=0.333333\textwidth]{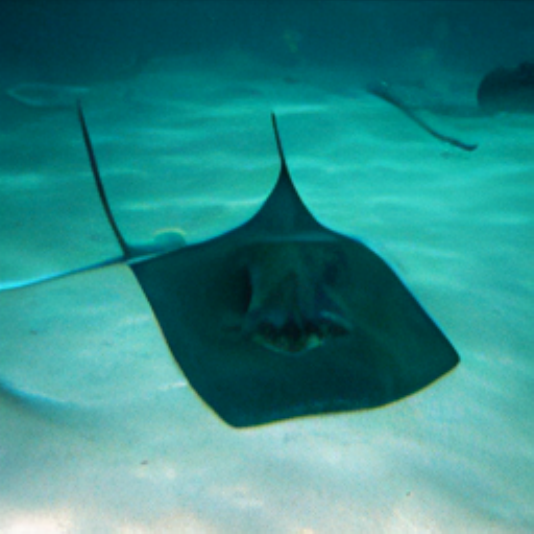}\\[-\lineskip]
\includegraphics[width=0.333333\textwidth]{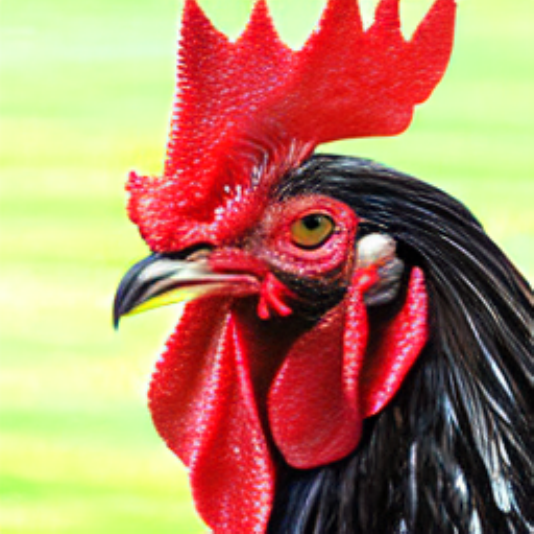}&%
\includegraphics[width=0.333333\textwidth]{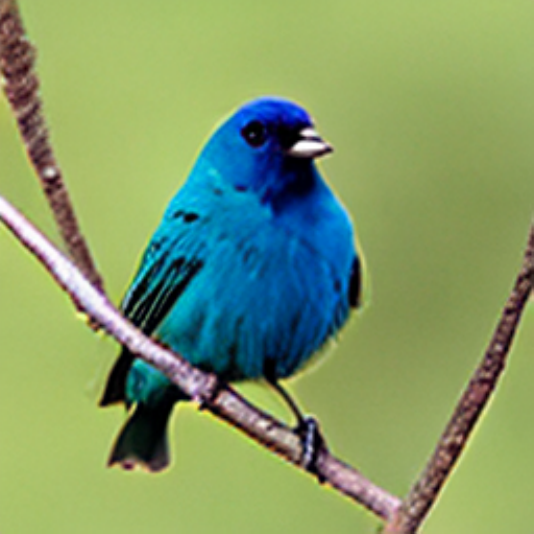}&%
\includegraphics[width=0.333333\textwidth]{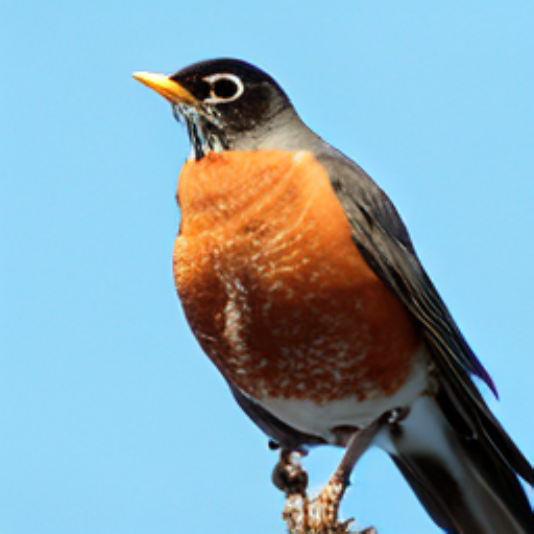}\\[-\lineskip]
\includegraphics[width=0.333333\textwidth]{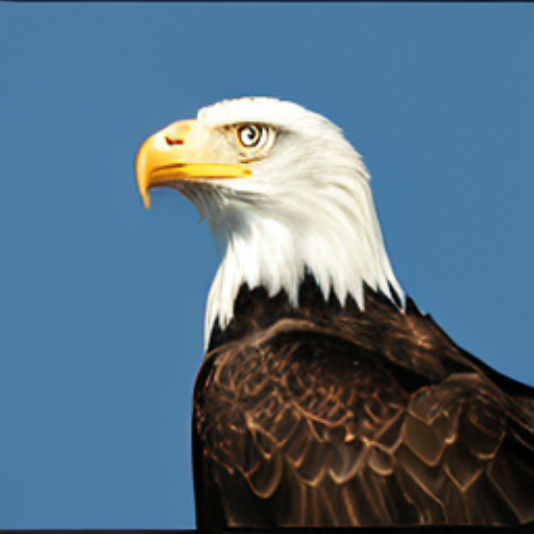}&%
\includegraphics[width=0.333333\textwidth]{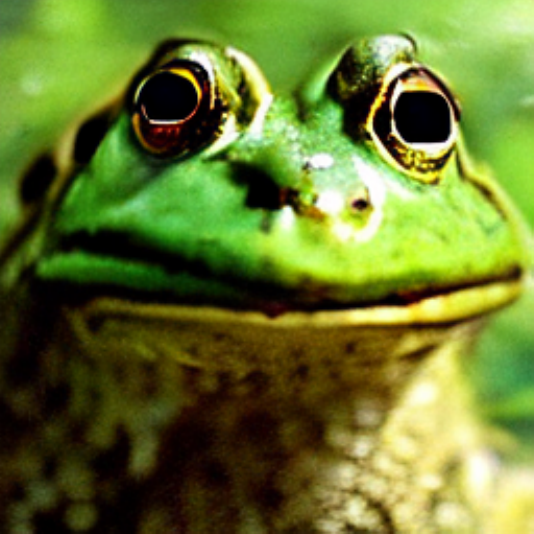}&%
\includegraphics[width=0.333333\textwidth]{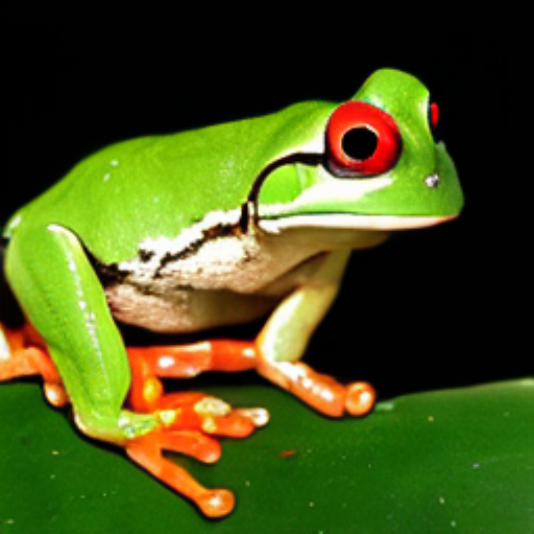}
\end{tabular}
\caption{EnLA generation samples at 256px on ImageNet-1K.}
\label{fig:supp-imagenet-1}
\end{figure*}
\clearpage

\begin{figure*}[p]
\centering
\setlength{\tabcolsep}{0pt}
\renewcommand{\arraystretch}{0}
\begin{tabular}{@{}ccc@{}}
\includegraphics[width=0.333333\textwidth]{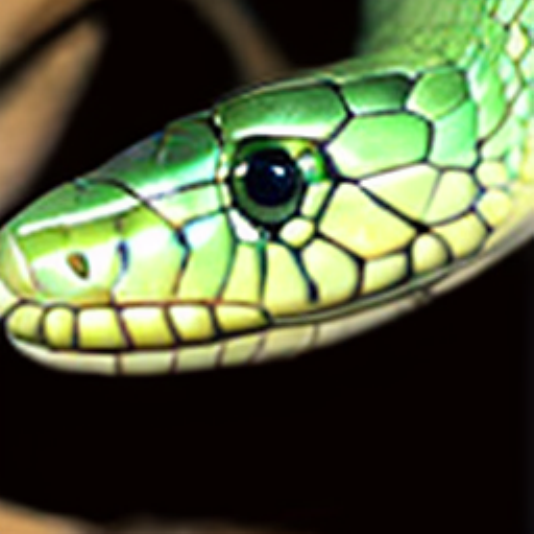}&%
\includegraphics[width=0.333333\textwidth]{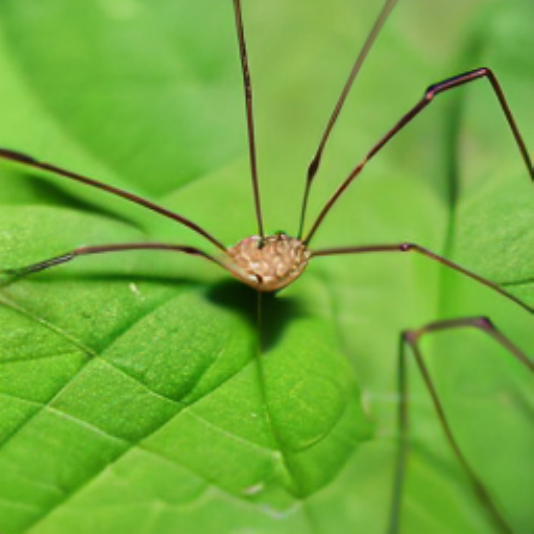}&%
\includegraphics[width=0.333333\textwidth]{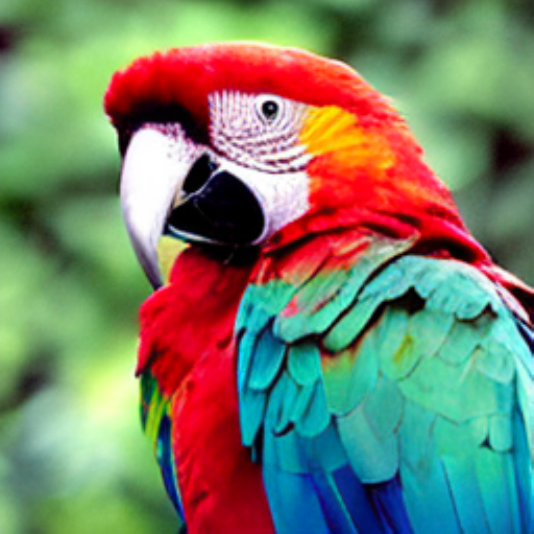}\\[-\lineskip]
\includegraphics[width=0.333333\textwidth]{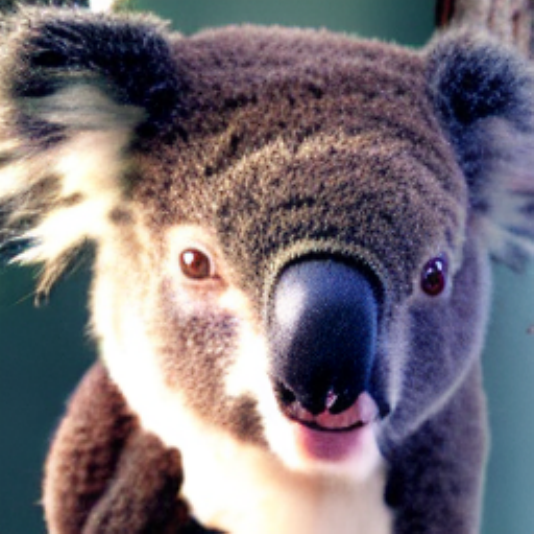}&%
\includegraphics[width=0.333333\textwidth]{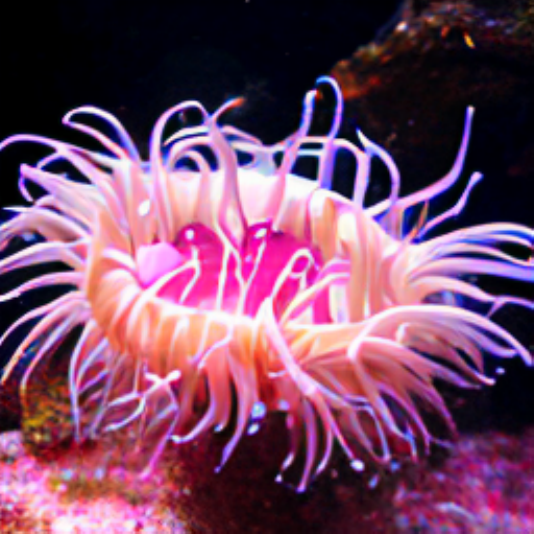}&%
\includegraphics[width=0.333333\textwidth]{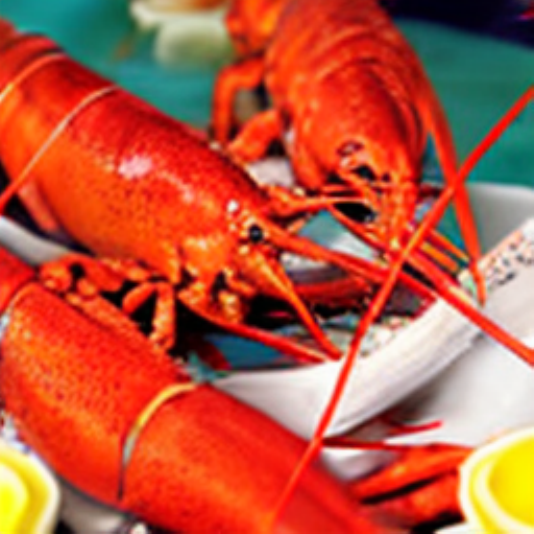}\\[-\lineskip]
\includegraphics[width=0.333333\textwidth]{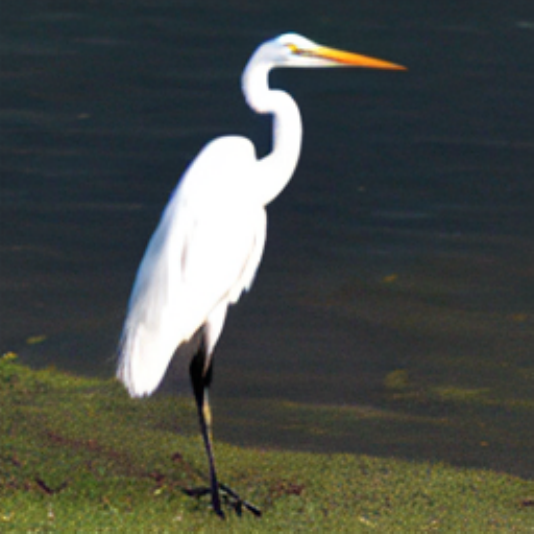}&%
\includegraphics[width=0.333333\textwidth]{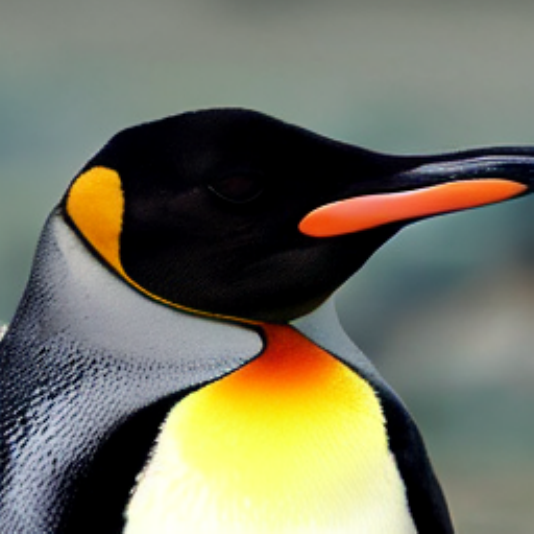}&%
\includegraphics[width=0.333333\textwidth]{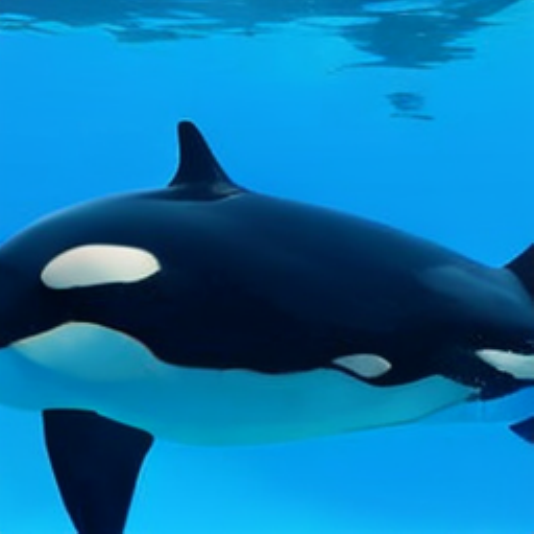}
\end{tabular}
\caption{EnLA generation samples at 256px on ImageNet-1K.}
\label{fig:supp-imagenet-2}
\end{figure*}
\clearpage

\begin{figure*}[p]
\centering
\setlength{\tabcolsep}{0pt}
\renewcommand{\arraystretch}{0}
\begin{tabular}{@{}ccc@{}}
\includegraphics[width=0.333333\textwidth]{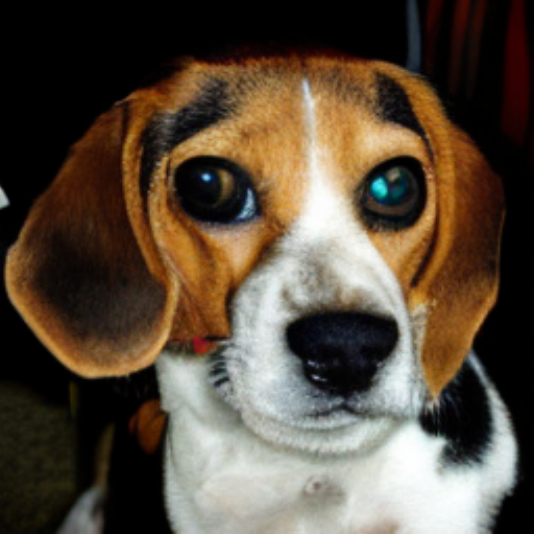}&%
\includegraphics[width=0.333333\textwidth]{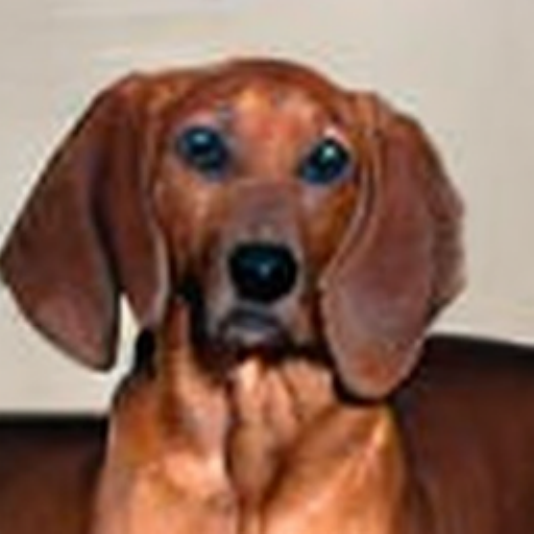}&%
\includegraphics[width=0.333333\textwidth]{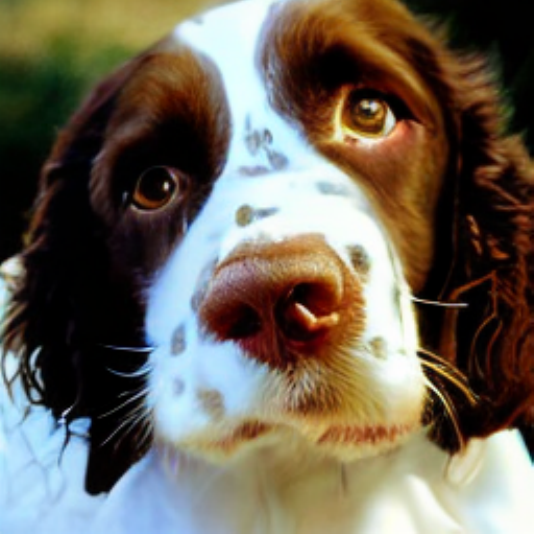}\\[-\lineskip]
\includegraphics[width=0.333333\textwidth]{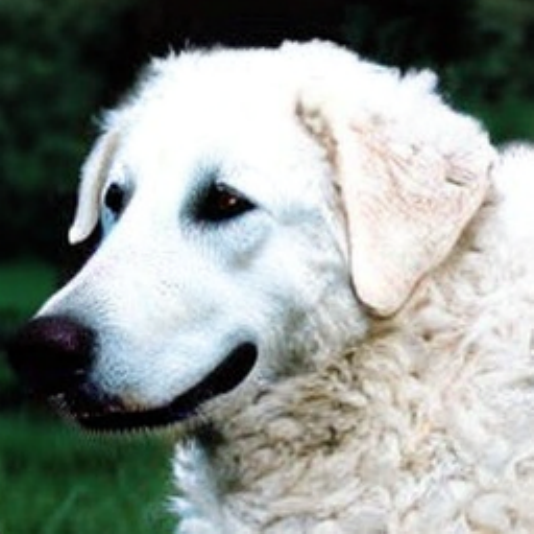}&%
\includegraphics[width=0.333333\textwidth]{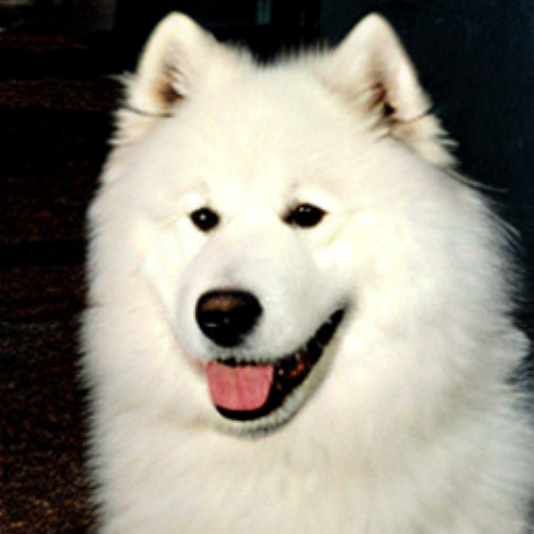}&%
\includegraphics[width=0.333333\textwidth]{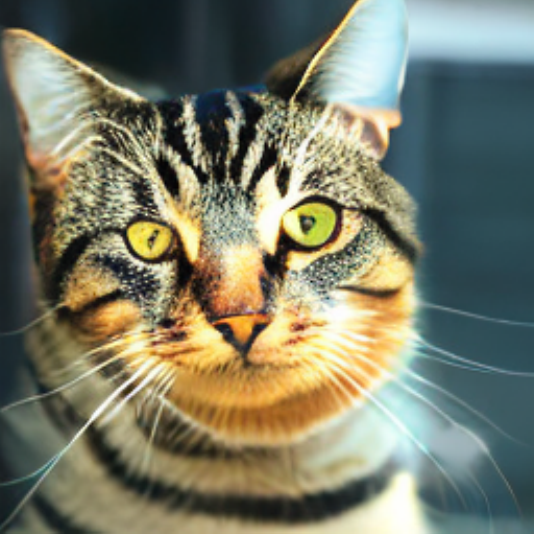}\\[-\lineskip]
\includegraphics[width=0.333333\textwidth]{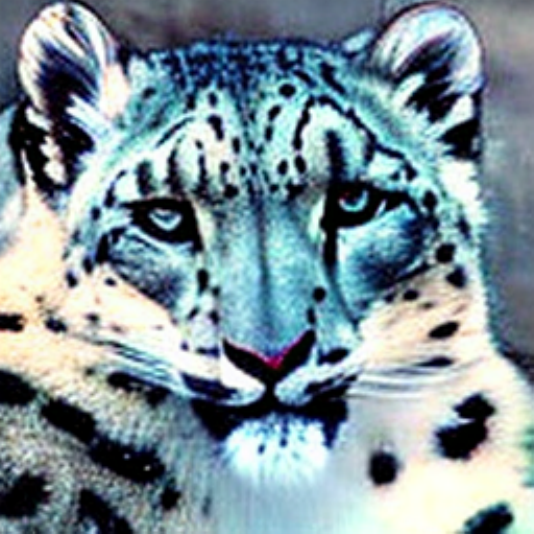}&%
\includegraphics[width=0.333333\textwidth]{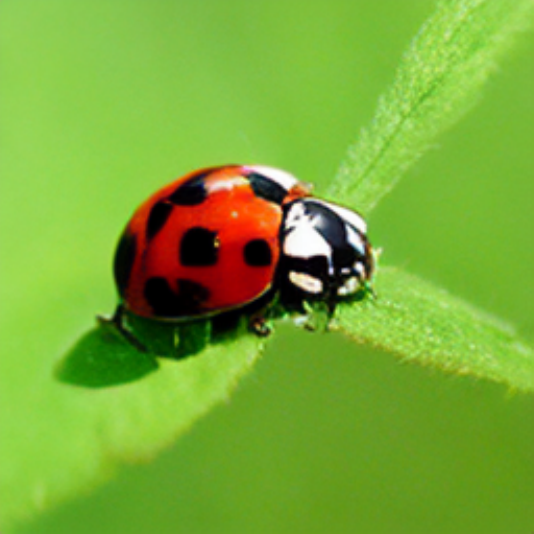}&%
\includegraphics[width=0.333333\textwidth]{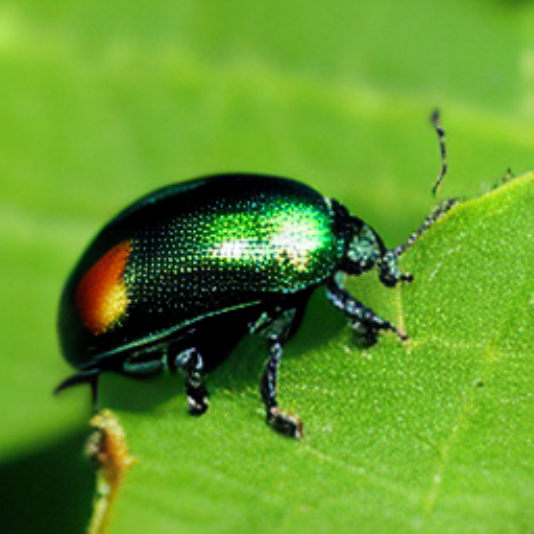}
\end{tabular}
\caption{EnLA generation samples at 256px on ImageNet-1K.}
\label{fig:supp-imagenet-3}
\end{figure*}
\clearpage

\begin{figure*}[p]
\centering
\setlength{\tabcolsep}{0pt}
\renewcommand{\arraystretch}{0}
\begin{tabular}{@{}ccc@{}}
\includegraphics[width=0.31\textwidth]{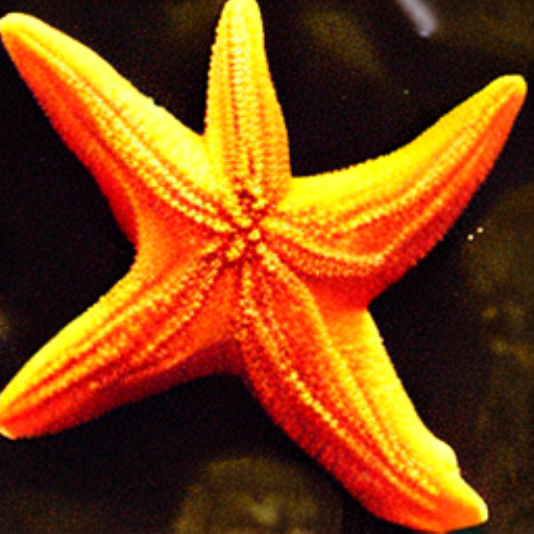}&%
\includegraphics[width=0.31\textwidth]{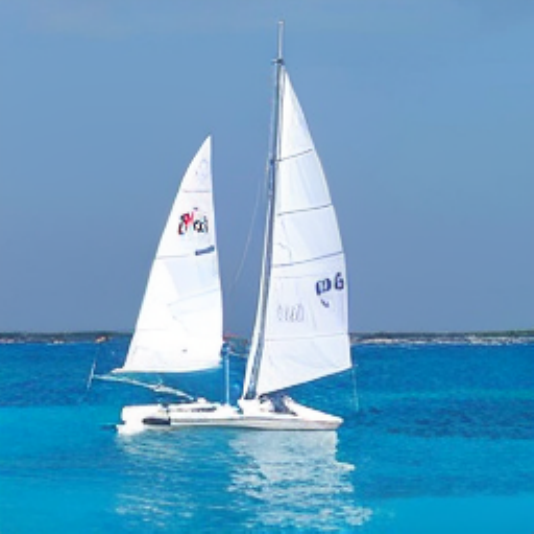}&%
\includegraphics[width=0.31\textwidth]{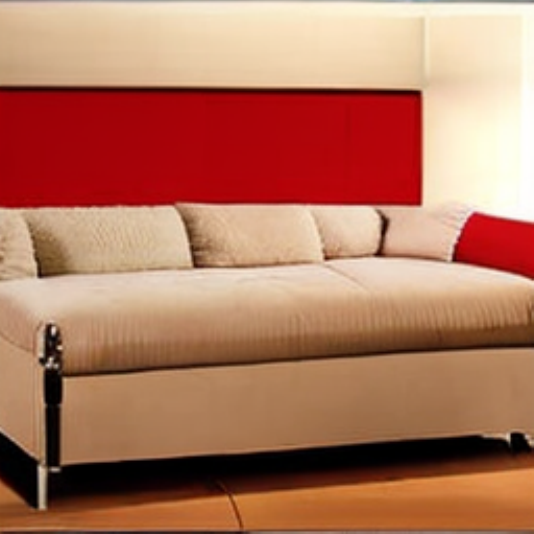}\\[-\lineskip]
\includegraphics[width=0.31\textwidth]{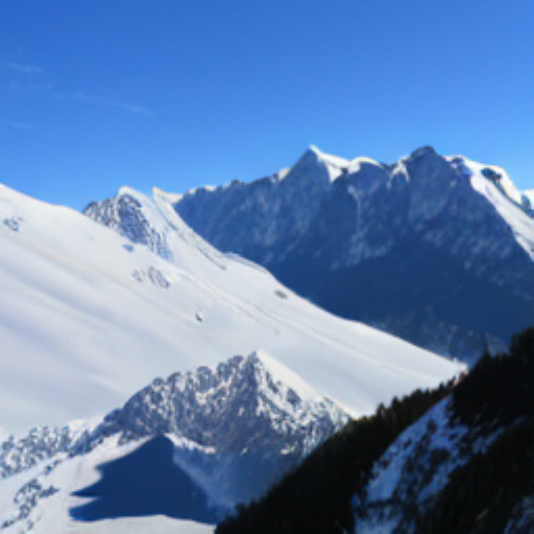}&%
\includegraphics[width=0.31\textwidth]{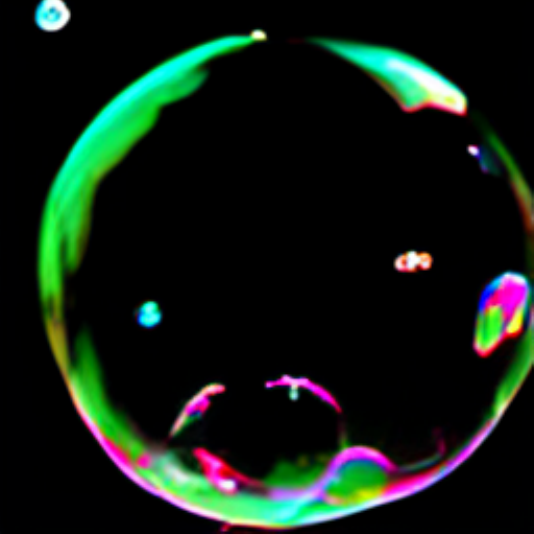}&%
\includegraphics[width=0.31\textwidth]{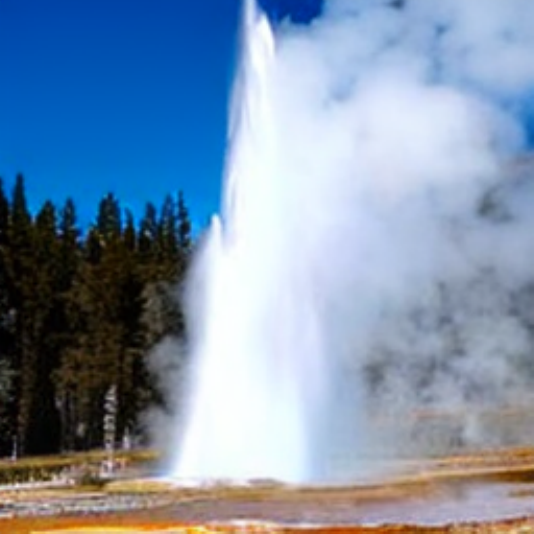}\\[-\lineskip]
\includegraphics[width=0.31\textwidth]{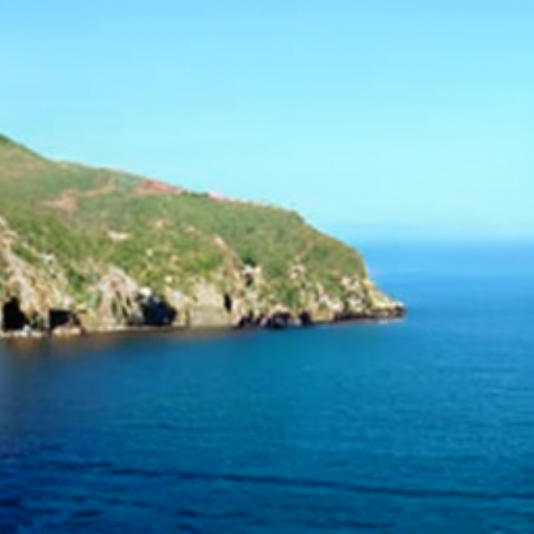}&%
\includegraphics[width=0.31\textwidth]{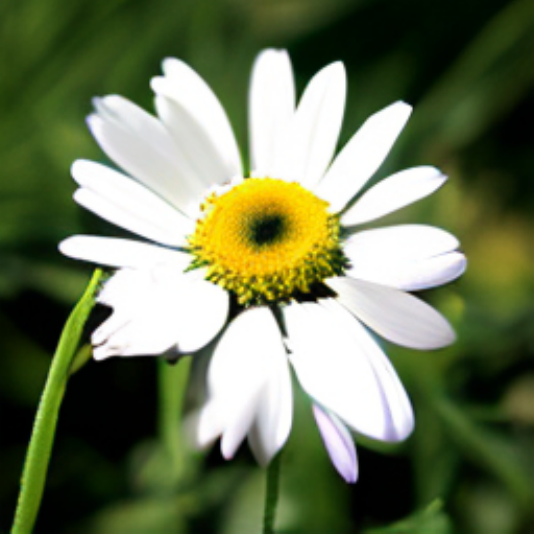}&%
\includegraphics[width=0.31\textwidth]{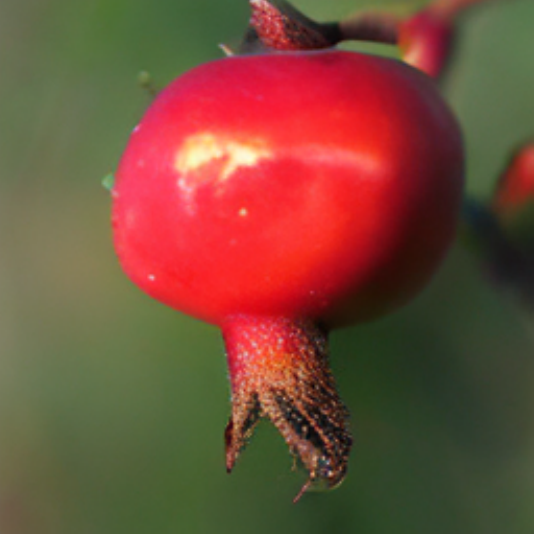}\\[-\lineskip]
\includegraphics[width=0.31\textwidth]{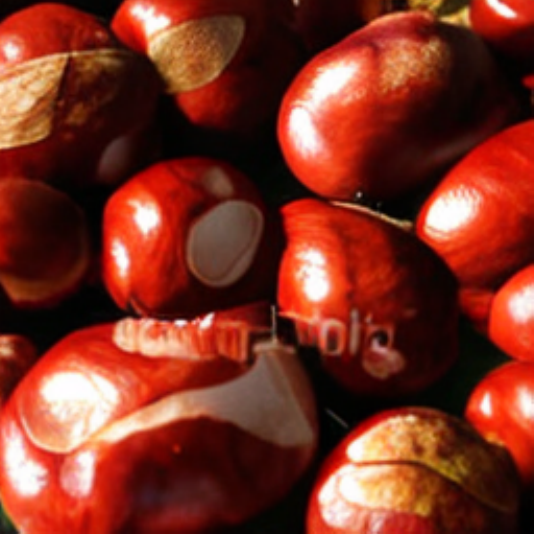}&%
\includegraphics[width=0.31\textwidth]{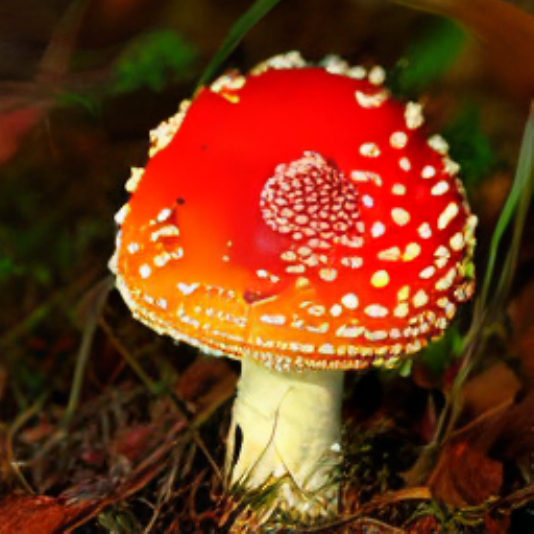}&%
\includegraphics[width=0.31\textwidth]{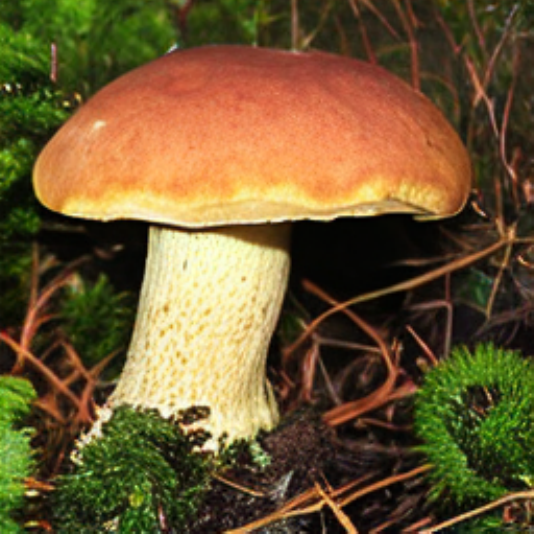}
\end{tabular}
\caption{EnLA generation samples at 256px on ImageNet-1K.}
\label{fig:supp-imagenet-4}
\label{fig:supp-imagenet-5}
\end{figure*}
\clearpage

\clearpage
\endgroup

\end{document}